\pdfoutput=1
\documentclass{article}
\usepackage[main,final]{neurips_2026}
\newif\ifarxiv\arxivfalse
\usepackage[utf8]{inputenc}
\usepackage[T1]{fontenc}
\usepackage{microtype}
\usepackage{amsmath, amssymb, amsfonts, mathtools}
\usepackage{booktabs}
\usepackage{array}
\usepackage{tabularx}
\usepackage{longtable}
\usepackage{multirow}
\usepackage{graphicx}
\usepackage{subcaption}
\usepackage{xcolor}
\usepackage{ragged2e}
\usepackage{enumitem}
\usepackage{algorithm}
\usepackage{algpseudocode}
\usepackage{siunitx}
\usepackage{wrapfig}
\usepackage{url}
\usepackage{hyperref}
\usepackage{placeins}
\usepackage[capitalize,noabbrev]{cleveref}
\usepackage{tikz}
\usetikzlibrary{
    positioning, 
    arrows.meta, 
    shapes.geometric, 
    fit,
    calc,                     %
    decorations.pathreplacing,%
    backgrounds               %
}
\newcommand{\methodname}{CoBit}
\newcommand{\bits}{\{0,1\}}

\newcommand{\E}{\mathbb{E}}

\newcommand{\sigmoid}{\operatorname{sigmoid}}

\newcommand{\clip}{\operatorname{clip}}

\newcommand{\para}[1]{\vspace{0.25em}\noindent\textbf{#1}}
\hypersetup{hidelinks,pdftitle={CoBit: Language Modeling with Bitstream Diffusion},pdfauthor={Georgios Batzolis, Mark Girolami, Luca Ambrogioni}}
\title{CoBit: Language Modeling with Bitstream Diffusion}
\author{Georgios Batzolis\thanks{Corresponding author: \texttt{g.batz97@gmail.com}}\\
Devotion AI Labs
\And
Mark Girolami\\
Devotion AI Labs\\
University of Cambridge
\And
Luca Ambrogioni\\
Donders Institute for Brain,\\
Cognition, and Behaviour\\
Radboud University}
\begin{document}
\maketitle
\begin{abstract}
We introduce CoBit, a continuous diffusion language model that generates text by jointly denoising fixed-width binary token codes. Grouping each code's bits into a single Transformer patch preserves sequence length while replacing vocabulary-wide predictions with $\lceil\log_2 V\rceil$ bit logits. A matched-filter residual parameterization supplies the analytic isolated-bit posterior, leaving the network to learn contextual corrections. Stochastic sampling on an entropy-rate grid provides the largest empirical gain; the grid allocates both solver resolution and effective Langevin correction to noise levels where information is resolved. At 256 denoiser evaluations, CoBit-S (130M parameters) achieves generative perplexity 59.76 on LM1B and 27.06 on OpenWebText, at unigram entropies 4.31 and 5.26, respectively. Scaling to 462M parameters improves OpenWebText generative perplexity to 19.48 at entropy 5.40. Beyond open-ended text generation, CoBit achieves 65.9\% exact-match accuracy on hard Sudoku. On GSM8K, score-temperature scaling with an untempered final integration step raises CoBit-S accuracy from 28.1\% to 35.3\% without additional denoiser evaluations, placing it within the reported range of temperature-tuned discrete diffusion baselines. Matched-trunk profiling also shows lower peak memory and higher throughput than a dense token-output baseline.
\end{abstract}

\section{Introduction}
\label{sec:introduction}
Autoregressive language models dominate modern text generation because they define a simple factorization and scale reliably. Their main algorithmic limitation is equally clear: generation is inherently sequential. Continuous diffusion models have become a standard framework for high-dimensional image and video generation \citep{ho2020ddpm,song2021scorebased,karras2022edm}. Diffusion language models (DLMs) offer a different compute profile, enabling parallel refinement of all positions, arbitrary infilling, and a tunable compute--quality trade-off. For language, however, good generative perplexity (GenPPL) can come at the cost of diversity: a model may favor safe, frequent tokens and concentrate its output distribution.

Recent continuous flow and diffusion language models show that continuous state spaces can work well for text \citep{lee2026flm,chen2026langflow}. Their success suggests that the difficulty lies in the interaction between representation, objective, and sampler design. We investigate this interaction through \emph{CoBit} (Continuous Bitstream Diffusion). Following Analog Bits \citep{chen2022analogbits}, we encode each token as a fixed-width binary code, corrupt the bits with Gaussian noise, and train a continuous denoiser to recover them.

The motivation for bits is the size of the prediction problem. Identifying one of $V$ symbols requires $m=\lceil\log_2 V\rceil$ binary digits, whereas a dense categorical head predicts $V$ logits per position. CoBit predicts one logit per bit and groups each token's bits into a single Transformer patch, preserving the sequence length. Contextual dependencies are learned through the shared network and iterative denoising. Moreover, the posterior of an isolated bit under Gaussian corruption has a closed form. Our \emph{matched-filter residual parameterization} supplies this local denoising term analytically, leaving the network to learn corrections from context.

\paragraph{The critical role of stochastic sampling.}
The bitstream representation provides the foundation, but our largest empirical gain comes from the sampler. The deterministic probability-flow sampler is competitive with recent continuous DLMs, yet its samples have low token-frequency entropy. EDM-style stochastic churn improves the GenPPL--entropy trade-off without changing the trained model or increasing the NFE budget. On an entropy-rate grid, the same spacing that concentrates solver resolution where bit uncertainty is resolved also concentrates the effective stochastic correction. We formalize this small-step interpretation in \Cref{app:stochastic_sde_analysis}.

\paragraph{Generation and verifiable reasoning.}
At 256 neural function evaluations (NFEs), stochastic sampling improves CoBit-S from 82.90 to 59.76 GenPPL on LM1B and from 46.32 to 27.06 on OpenWebText (OWT), while increasing entropy. Scaling from 130M to 462M parameters further improves OWT GenPPL to 19.48 at entropy 5.40, approaching the pretrained GPT-2 Medium reference. These metrics capture a useful quality--diversity trade-off, but cannot establish whether a generation is correct. We therefore also evaluate Sudoku and executable mathematics. CoBit reaches 65.9\% exact match on hard Sudoku, compared with 58.4\% for Duo in the same benchmark. On TinyGSM-trained CoBit-S, score-temperature scaling raises GSM8K accuracy from 28.1\% to 35.3\% at unchanged NFE budget, reaching the reported 33--36\% range of temperature-tuned MDLM and Duo.

\paragraph{Additional computational benefits.}
Bit-patching replaces the $\mathcal{O}(TV)$ dense vocabulary output with $\mathcal{O}(T\log V)$ bit logits. With a shared 12-layer Transformer trunk, this gives $1.64\times$ lower peak training memory and $2.31\times$ higher throughput at LM1B scale; the corresponding OWT gains are $2.46\times$ and $3.29\times$. These measurements quantify the practical benefit of the compact interface, alongside the generation results.

\section{Related Work}
\label{sec:related}

\para{Discrete diffusion language models.}
Discrete DLMs define corruption processes directly on tokens or masks. D3PM introduced structured discrete denoising diffusion \citep{austin2021d3pm}; SEDD framed discrete diffusion through ratio estimation and score entropy \citep{lou2024discrete}. MDLM strengthened masked language modeling with a simplified objective and training recipe \citep{sahoo2024mdlm}, while Duo connected uniform-state discrete diffusion to Gaussian diffusion \citep{sahoo2025duo}. MDM-Prime represents tokens as sub-token sequences and reveals them through partial masking \citep{chao2025mdmprime,chao2026mdmprimev2}. Its joint categorical decoder retains vocabulary-sized logits. CoBit instead learns bitwise posterior means under continuous Gaussian corruption and decodes by thresholding. We compare both a released MDM-Prime checkpoint and an independent-head SEDD-on-bits baseline.

\para{Continuous diffusion for categorical data.}
Analog Bits represents discrete variables as binary codes and trains continuous diffusion on their analog values; it also introduced self-conditioning and asymmetric time intervals \citep{chen2022analogbits}. CDCD adapts continuous diffusion and time warping to categorical data \citep{dieleman2022cdcd}, while RDLM and CANDI explore manifold and hybrid representations \citep{jo2025rdlm,pynadath2025candi}. FLM/FMLM and LangFlow provide strong evidence that continuous language models can rival discrete DLMs \citep{lee2026flm,chen2026langflow}. ELF uses flows in learned embedding space and supports classifier-free guidance \citep{hu2026elf}. Our representation follows Analog Bits, with a language model built around a matched-filter residual denoiser and entropy-gated stochastic sampling. The entropy-rate grid builds on entropic scheduling and information-guided training \citep{stancevic2025entropic,raya2026infonoise}; it also determines the effective strength of EDM churn \citep{karras2022edm}.

\para{Generation quality and reasoning.}
External generative perplexity must be considered jointly with diversity \citep{pynadath2026generativefrontiers}. Hyperspherical flows further highlight that fluent generations need not solve verifiable tasks \citep{deschenaux2026continuous}. We follow their Sudoku and TinyGSM-to-GSM8K protocols to complement the natural-language metrics. Sigma introduces the score-temperature operation we investigate \citep{yang2026sigma}; we compare standard and temperature-tuned decoding on the verifiable tasks.

\section{Method}
\label{sec:method}

\begin{figure}[!ht]
\centering
\resizebox{0.92\linewidth}{!}{%
\begin{tikzpicture}[
    >=stealth,
    font=\small\sffamily,
    tensor/.style={draw, thick, fill=gray!10, rounded corners=2pt, minimum height=0.6cm, align=center},
    opbox/.style={draw=blue!80, thick, fill=blue!5, rounded corners=4pt, minimum height=0.8cm, align=center},
    mlpbox/.style={draw=orange!80, thick, fill=orange!5, rounded corners=4pt, minimum height=0.8cm, align=center},
    mathnode/.style={draw=none, fill=none, align=center, font=\footnotesize},
    add/.style={circle, draw, thick, inner sep=2pt, fill=white},
    textnode/.style={draw=green!60!black, thick, fill=green!5, rounded corners=2pt, minimum height=0.6cm, minimum width=2.6cm, align=center},
    bitnode/.style={draw=gray!50, thick, fill=white, rounded corners=2pt, font=\ttfamily\scriptsize, minimum height=0.5cm, minimum width=2.6cm, align=center}
]

\coordinate (c1) at (-3.1, 0);
\coordinate (c2) at (0, 0);
\coordinate (c3) at (3.1, 0);

\node (word1) [textnode] at (c1) {Token: ``The''};
\node (word2) [textnode] at (c2) {Token: ``cat''};
\node (worddots) [mathnode] at (1.55, 0) {$\cdots$};
\node (wordT) [textnode] at (c3) {Token: ``sat''};

\node (bits1) [bitnode, below=0.6cm of word1] {0110...10};
\node (bits2) [bitnode, below=0.6cm of word2] {1001...01};
\node (bitsdots) [mathnode, below=0.6cm of worddots] {$\cdots$};
\node (bitsT) [bitnode, below=0.6cm of wordT] {0011...11};

\draw[->, thick] (word1.south) -- (bits1.north) node[midway, left, font=\scriptsize] {$m=15$};
\draw[->, thick] (word2.south) -- (bits2.north) node[midway, left, font=\scriptsize] {$m=15$};
\draw[->, thick] (wordT.south) -- (bitsT.north) node[midway, left, font=\scriptsize] {$m=15$};

\begin{scope}[on background layer]
    \node [tensor, fit=(bits1) (bitsT), inner sep=6pt, draw=gray!40] (x0_box) {};
\end{scope}
\node [left=0.2cm of x0_box, mathnode, align=right] {Clean Bits\\$x_0 \in \{0,1\}^{B \times Tm}$};

\node (abits1) [bitnode, below=1.8cm of bits1] {0.82 \dots -0.14};
\node (abits2) [bitnode, below=1.8cm of bits2] {1.05 \dots 0.91};
\node (abitsdots) [mathnode, below=1.8cm of bitsdots] {$\cdots$};
\node (abitsT) [bitnode, below=1.8cm of bitsT] {-0.22 \dots 0.05};

\begin{scope}[on background layer]
    \node [tensor, fit=(abits1) (abitsT), inner sep=6pt, draw=gray!60] (x_sigma) {};
\end{scope}
\node [left=0.2cm of x_sigma, mathnode, align=right] {Diffused Bits\\$x_\sigma \in \mathbb{R}^{B \times Tm}$};

\draw[->, thick] (x0_box.south) -- (x_sigma.north)
node[midway, right, align=left, font=\footnotesize]
{Gaussian Corruption\\$\oplus\ \sigma \epsilon$};

\draw [decorate,decoration={brace,amplitude=5pt,mirror}, thick]
(abits1.south west) -- (abits1.south east)
node [midway, below=8pt, mathnode] (patch1) {Patch 1};

\draw [decorate,decoration={brace,amplitude=5pt,mirror}, thick]
(abits2.south west) -- (abits2.south east)
node [midway, below=8pt, mathnode] (patch2) {Patch 2};

\node (patchdots) [mathnode] at ([yshift=-13pt]abitsdots.south) {$\cdots$};

\draw [decorate,decoration={brace,amplitude=5pt,mirror}, thick]
(abitsT.south west) -- (abitsT.south east)
node [midway, below=8pt, mathnode] (patchT) {Patch $T$};

\node (embed) [opbox, minimum width=9cm, below=1.2cm of x_sigma]
{Patch Adapter (Linear) $\to \mathbb{R}^{B \times T \times d}$};

\draw[->, thick] (patch1.south) -- (patch1.south |- embed.north);
\draw[->, thick] (patch2.south) -- (patch2.south |- embed.north);
\draw[->, thick] (patchT.south) -- (patchT.south |- embed.north);

\node (pos_time) [mathnode, left=0.6cm of embed] {Time $\sigma$ \\ \& Pos Emb};
\draw[->, thick] (pos_time) -- (embed);

\node (trunk) [opbox, minimum width=9cm, minimum height=1.5cm, below=0.8cm of embed]
{\textbf{Sequence Diffusion Transformer Trunk}\\12 Blocks $\mid$ Processes Semantic Length $T$};

\draw[->, thick] (embed.south) -- (trunk.north)
node[midway, right, font=\footnotesize] {Token Embeddings};

\node (filter) [opbox, below=1.8cm of trunk, xshift=-2.6cm, minimum width=4cm]
{\textbf{Analytic Matched Filter}\\$\clip\left(\frac{x_\sigma - 1/2}{\sigma^2}\right)$};

\node (mlp) [mlpbox, below=1.8cm of trunk, xshift=2.6cm, minimum width=4cm]
{\textbf{Optimal Skip MLP}\\(Predicts Residual $r_\theta$)};

\draw[->, thick] ([xshift=2.6cm]trunk.south) -- (mlp.north)
node[midway, right, font=\footnotesize, align=left]
{Global Patch\\Context $B \times T$};

\draw[->, thick, orange, rounded corners]
(x_sigma.east) -- ++(1.2cm,0) |- (mlp.east)
node[pos=0.25, right, align=left]
{\textbf{Skip Connection}\\Local Noisy-Bits};

\node (plus) [add, below=3.0cm of trunk] {$+$};
\draw[->, thick] (filter.east) -- (plus.west);
\draw[->, thick] (mlp.west) -- (plus.east);

\node (logits) [tensor, below=0.6cm of plus, minimum width=6cm]
{Total Logit $\ell_\theta \in \mathbb{R}^{B \times Tm}$};

\draw[->, thick] (plus.south) -- (logits.north);

\begin{scope}[on background layer]
    \node [fill=gray!5, rounded corners, fit=(patch1) (patchT) (embed) (trunk),
           draw=gray!30, dashed, inner sep=12pt] (trunkgroup) {};

    \node [fill=orange!5, rounded corners, fit=(mlp) (filter) (plus) (logits),
           draw=orange!30, dashed, inner sep=14pt] (headgroup) {};

    \node [below left, font=\footnotesize\color{orange!80}] at (headgroup.north east)
    {Length $Tm$};
\end{scope}

\end{tikzpicture}%
}
\caption{\textbf{End-to-end \methodname{} architecture.}
Text is encoded into a \(Tm\)-length bitstream and corrupted via Gaussian noise.
The diffused bits \(x_\sigma\) are patched into semantic groups and processed by a length-\(T\) Transformer trunk.
In parallel with the trunk, a skip connection routes the raw noisy bits directly to the head, where a contextual residual is combined with the analytic matched filter to produce bit-level logits.}
\label{fig:e2e_architecture}
\end{figure}
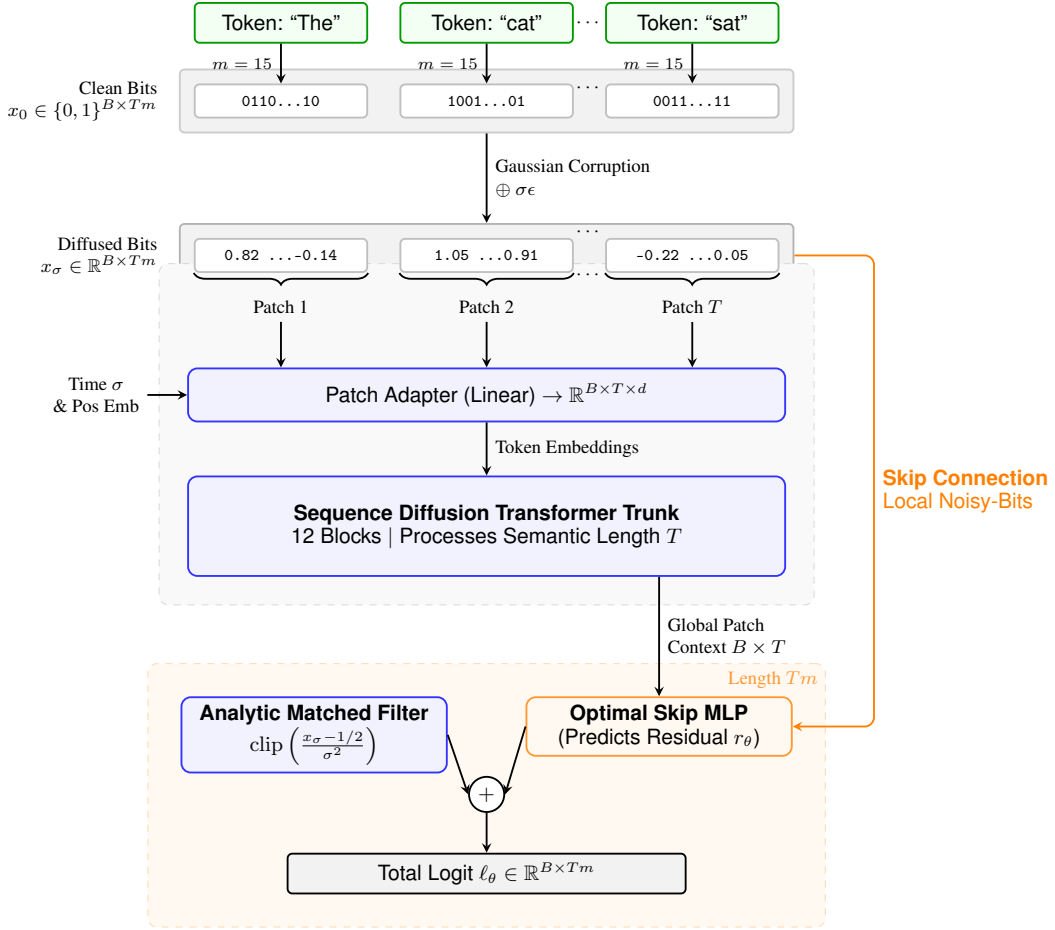

We represent a sequence of \(T\) tokens or code tokens as \(x_0\in\bits^{S}\), with \(S=Tm\) and \(m\) bits per token. LM1B uses raw 15-bit tokenizer IDs; OWT uses 16-bit code tokens from a reversible second-stage codec. The Transformer processes one patch per token or code token, so its sequence length is \(T\), rather than \(Tm\). A codec token need not correspond to one original tokenizer token; we evaluate OWT after inverse decoding. The architecture and decoding details are in \Cref{fig:e2e_architecture,app:decoding_details}.

We use variance-exploding Gaussian corruption of the \(0/1\) bits,
\begin{equation}
    x_\sigma=x_0+\sigma\epsilon,
    \qquad \epsilon\sim\mathcal{N}(0,I_S),
    \qquad \sigma\in[\sigma_{\min},\sigma_{\max}].
    \label{eq:forward}
\end{equation}
The denoiser \(D_\theta(x_\sigma,\sigma)\in(0,1)^S\) estimates the clean bitwise posterior means. The Gaussian posterior-mean identity then motivates the score approximation
\begin{equation}
    s_\theta(x_\sigma,\sigma)
    =\frac{D_\theta(x_\sigma,\sigma)-x_\sigma}{\sigma^2}.
    \label{eq:score_from_denoiser}
\end{equation}
This is the exact marginal score when \(D_\theta=\E[x_0\mid x_\sigma]\); using bitwise posterior means does not assume that the joint data distribution factorizes over bits.

\subsection{Matched-filter residual parameterization}
\label{subsec:matched_filter}

For an isolated bit with uniform prior, the Gaussian channel has posterior log-odds
\begin{equation}
    \ell_{\mathrm{ind}}(x,\sigma)
    =\log\frac{p(x_0=1\mid x,\sigma)}{p(x_0=0\mid x,\sigma)}
    =\frac{x-\frac12}{\sigma^2}.
    \label{eq:analytic_logit}
\end{equation}
We supply this analytic baseline and learn a residual that can account for nonuniform bit priors and contextual dependencies:
\begin{equation}
    \ell_\theta
    =r_\theta(x_\sigma,\sigma,x_{\mathrm{sc}})
    +\clip\!\left(\frac{x_\sigma-\frac12}{\sigma^2},-C,C\right),
    \qquad D_\theta=\sigmoid(\ell_\theta),\quad C=30.
    \label{eq:matched_filter_residual}
\end{equation}
Here \(x_{\mathrm{sc}}\) is the self-conditioning input. The derivation is in \Cref{app:matched_filter_derivation}.

\subsection{Training objective and self-conditioning}
\label{subsec:objective}

We train with weighted denoising mean-squared error,
\begin{equation}
    \mathcal{L}_{\mathrm{SM}}(\theta)
    =\E_{x_0,\epsilon,\sigma}\!\left[
      w(\sigma)\frac{1}{S}\|D_\theta(x_\sigma,\sigma)-x_0\|_2^2
    \right],
    \label{eq:binary_sm_loss}
\end{equation}
using the EDM weight \citep{karras2022edm}
\begin{equation}
    w(\sigma)=\frac{\sigma^2+\sigma_{\mathrm{data}}^2}
    {\sigma^2\sigma_{\mathrm{data}}^2},
    \qquad \sigma_{\mathrm{data}}=\tfrac12.
    \label{eq:edm_weight}
\end{equation}
Through \Cref{eq:score_from_denoiser}, this is a weighted denoising score-matching objective. With probability \(0.5\), an auxiliary no-gradient training pass supplies detached posterior probabilities as \(x_{\mathrm{sc}}\). At sampling time, carry-mode self-conditioning reuses the preceding denoised prediction \citep{chen2022analogbits,chen2026langflow}.

\subsection{Entropy-rate adaptive noise allocation}
\label{subsec:entropy_schedule}

Bitstream information profiles depend on the tokenizer, code representation, and dataset. We therefore adapt the information-guided noise allocation of \citet{stancevic2025entropic,raya2026infonoise}. Let \(h_{\log}(\sigma)=\frac{d}{d\log\sigma}H(x_0\mid x_\sigma)\) describe where corruption removes information. The I--MMSE identity relates this rate to the Bayes denoising error divided by \(\sigma^2\). We estimate its profile online using the unweighted per-bit training error, giving a proxy \(\widehat h_{\log}\) up to a constant normalization. With \(u=\log\sigma\), define
\begin{equation}
    \pi_\alpha(u)\propto
    g(e^u;c,n)\,\widehat h_{\log}(e^u)^\alpha,
    \qquad g(\sigma;c,n)=\frac{\sigma^n}{\sigma^n+c^n}.
    \label{eq:normalized_entropy_rate_density_main}
\end{equation}
We normalize over the noise interval and use \(\alpha=1/2\), \(c=0.1\), and \(n=3\). Training transitions from EDM log-normal noise sampling to this distribution after warmup. Its inverse CDF also defines the default sampling grid. The online estimate uses the learned denoiser; buffer, binning, and warmup details appear in \Cref{app:entropy_schedule_details}.

\subsection{Sampling: deterministic flow and stochastic correction}
\label{subsec:sampling}

The probability-flow ODE associated with \Cref{eq:score_from_denoiser} is
\begin{equation}
    \frac{dx}{d\sigma}=-\sigma s_\theta(x,\sigma)
    =\frac{x-D_\theta(x,\sigma)}{\sigma}.
    \label{eq:sampling_ode}
\end{equation}
Our default DDIM-style update integrates this equation on the entropy-rate grid, adapting entropic time warping \citep{dieleman2022cdcd,stancevic2025entropic}. All positions are updated in parallel.

\paragraph{Entropy-gated stochastic churn.}
Before stepping from \(\sigma_i\) to \(\sigma_{i+1}\), EDM churn raises the noise level to \(\hat\sigma_i=(1+\gamma_i)\sigma_i\) and adds Gaussian noise with standard deviation \(S_{\mathrm{noise}}\sqrt{\hat\sigma_i^2-\sigma_i^2}\), where \(S_{\mathrm{noise}}=1.003\). It then applies the deterministic update from \((\hat x_i,\hat\sigma_i)\). For \(N\) churn-eligible intervals, full-band churn uses \(\gamma_i=\min(S_{\mathrm{churn}}/N,\sqrt2-1)\).

In the small-step, unclipped limit with \(S_{\mathrm{noise}}=1\), this split update approximates a probability-flow step plus a Langevin correction of strength
\begin{equation}
    \lambda_i\approx\frac{\gamma_i\sigma_i}{\sigma_i-\sigma_{i+1}}.
    \label{eq:effective_lambda_main}
\end{equation}
Uniform spacing in the CDF of \(\pi_\alpha\) gives \(\sigma_i-\sigma_{i+1}\approx\sigma_i/[N\pi_\alpha(\log\sigma_i)]\), hence
\begin{equation}
    \lambda_{\mathrm{ent}}(\sigma_i)
    \approx S_{\mathrm{churn}}\,\pi_\alpha(\log\sigma_i).
    \label{eq:main_entropic_lambda}
\end{equation}
The entropy-rate grid therefore has a dual role: it allocates solver points to information-active noise levels and converts constant raw churn into an information-adaptive Langevin correction. \Cref{app:stochastic_sde_analysis} gives the derivation and its finite-step limitations. We set the optional asymmetric-time shift to \(\eta=0\) for the main language-modeling results; ablations appear in \Cref{app:ati_details,app:sampling_ablations}.

\subsection{Sequence diffusion transformer and vocabulary scaling}
\label{subsec:architecture}

The denoiser is a sequence diffusion transformer operating at length \(T\), rather than bit length \(Tm\). A patch adapter projects each group of \(m\) analog bits into one Transformer input. The output head combines global patch context with local noisy-bit and self-conditioning features to predict \(m\) residual logits. CoBit-S uses 12 blocks of width 768; the medium OWT model uses 24 blocks of width 1024. Full architecture and optimization settings are in \Cref{app:training_architecture_details}.

For a vocabulary of \(V\) codes, \(m=\lceil\log_2V\rceil\). CoBit therefore emits \(\mathcal{O}(T\log V)\) logits instead of the \(\mathcal{O}(TV)\) logits of a dense categorical head, while retaining \(T\) trunk positions. At LM1B scale, \(V/m=30{,}522/15\approx2035\); actual memory and speed gains are smaller because the Transformer and adapters also consume resources. We measure the practical memory and throughput impact in \Cref{subsec:efficiency_main}.

\Cref{app:binary_boundary} gives the corresponding comparison with higher-base factorized digit heads. Binary minimizes their independent-logit count; whether another code improves generation quality remains an empirical question.

\section{Experiments}
\label{sec:experiments}
We evaluate CoBit on LM1B and OpenWebText, then on Sudoku and executable mathematics. We first compare generation quality and diversity, followed by reasoning accuracy, computational cost, and component ablations.
\subsection{Experimental setup}
\label{subsec:setup}
\textbf{Data and models.} We train on \href{https://research.google/pubs/one-billion-word-benchmark-for-measuring-progress-in-statistical-language-modeling/}{LM1B} (128 tokens, \href{https://huggingface.co/google-bert/bert-base-uncased}{\texttt{bert-base-uncased}}, 15 bits per token, 1M steps) and \href{https://skylion007.github.io/OpenWebTextCorpus/}{OWT} (1,024 code tokens, 750K steps). OWT uses a reversible second-stage BPE codec over \href{https://github.com/openai/gpt-2}{GPT-2} ID sequences: the 65,536-entry code vocabulary contains the original GPT-2 IDs and learned merges, with 16 bits per code. After generation, inverse BPE recovers GPT-2 IDs and text for evaluation. A block of 1,024 code positions can therefore contain more than 1,024 GPT-2 tokens. CoBit-S uses 12 blocks of width 768 (130M parameters); CoBit-M uses 24 blocks of width 1,024 (462M). The models share the representation and objective, with scale-specific optimization and terminal-noise settings documented in \Cref{app:training_architecture_details}.

\textbf{Metrics and sampling.} We score 1,024 unconditional CoBit samples per operating point with \href{https://huggingface.co/openai-community/gpt2-large}{\texttt{gpt2-large}} and report mean per-sample unigram entropy in nats, computed on LM1B tokenizer IDs or recovered GPT-2 IDs. We consider these metrics jointly because GenPPL alone can improve through concentration on frequent tokens \citep{pynadath2026generativefrontiers}. We compare deterministic and stochastic samplers on the entropy-rate grid, with $S_{\rm noise}=1.003$ and carry-mode self-conditioning. NFE counts denoiser evaluations, not wall-clock cost. Unless stated otherwise, our results use 256 NFEs.

\subsection{Language generation and model scale}
\label{subsec:main_results}
\begin{table}[t]
\centering\small
\begin{tabular}{lrrrr}
\toprule
& \multicolumn{2}{c}{LM1B} & \multicolumn{2}{c}{OWT}\\
\cmidrule(lr){2-3}\cmidrule(lr){4-5}
Method & GenPPL $\downarrow$ & Entropy & GenPPL $\downarrow$ & Entropy\\
\midrule
Real data & 53.06 & 4.34 & 15.06 & 5.44\\
GPT-S ($p=0.90$; LM1B) & 43.00 & 4.299 & -- & --\\
GPT-S ($p=0.95$; LM1B) & 52.80 & 4.327 & -- & --\\
GPT-S ($p=0.98$; LM1B) & 61.20 & 4.342 & -- & --\\
GPT-2 Small ($p=0.90$; OWT) & -- & -- & 23.91 & 5.374\\
\midrule
CANDI & 120.99 & 4.35 & 143.13 & 5.71\\
MDLM & 109.21 & 4.32 & 105.15 & 5.63\\
Duo & 98.14 & 4.31 & 77.69 & 5.55\\
FLM & 96.91 & 4.29 & 62.23 & 5.33\\
LangFlow (128/1,024 NFE) & 92.24 & 4.31 & 36.53 & 5.25\\
\midrule
CoBit-S deterministic & $82.90\pm1.11$ & 4.30 & $46.32\pm0.93$ & 5.13\\
CoBit-S stochastic & $59.76\pm0.57$ & 4.31 & $27.06\pm0.57$ & 5.26\\
CoBit-M stochastic & -- & -- & 19.48 & 5.396\\
\bottomrule
\end{tabular}
\caption{Representative operating points; CoBit uses 256 NFEs. CoBit-S uncertainties are sampling standard deviations across ten seeds (entropy s.d. $<0.01$ except stochastic OWT, 0.01), not retraining uncertainty. LM1B GPT-S is trained from scratch and evaluated at the three nucleus probabilities shown; OWT GPT-2 is pretrained and nucleus-sampled at temperature 1. External diffusion rows are reference settings, not optimized fixed-budget comparisons. Further trajectories and evaluation details are in \Cref{app:owt_trajectories}.}
\label{tab:main_results}
\end{table}

The deterministic and stochastic rows of \Cref{tab:main_results} use the same trained checkpoint and NFE budget. This comparison isolates the effect of full-band churn: the deterministic sampler is competitive, but stochastic correction improves the GenPPL--entropy trade-off on both datasets.

\paragraph{LM1B.}
The stochastic sampler obtains GenPPL 59.76 at entropy 4.31, compared with 82.90/4.30 deterministically. This improves GenPPL by more than 30 points over the reported LangFlow setting at the same entropy. Real-data entropy is 4.34, so a diversity gap remains. The nucleus-sampled GPT-S reference also remains stronger: at $p=0.95$ it achieves 52.80/4.327, improving both metrics over CoBit-S.

\begin{figure}[p]
\centering
\includegraphics[width=\linewidth]{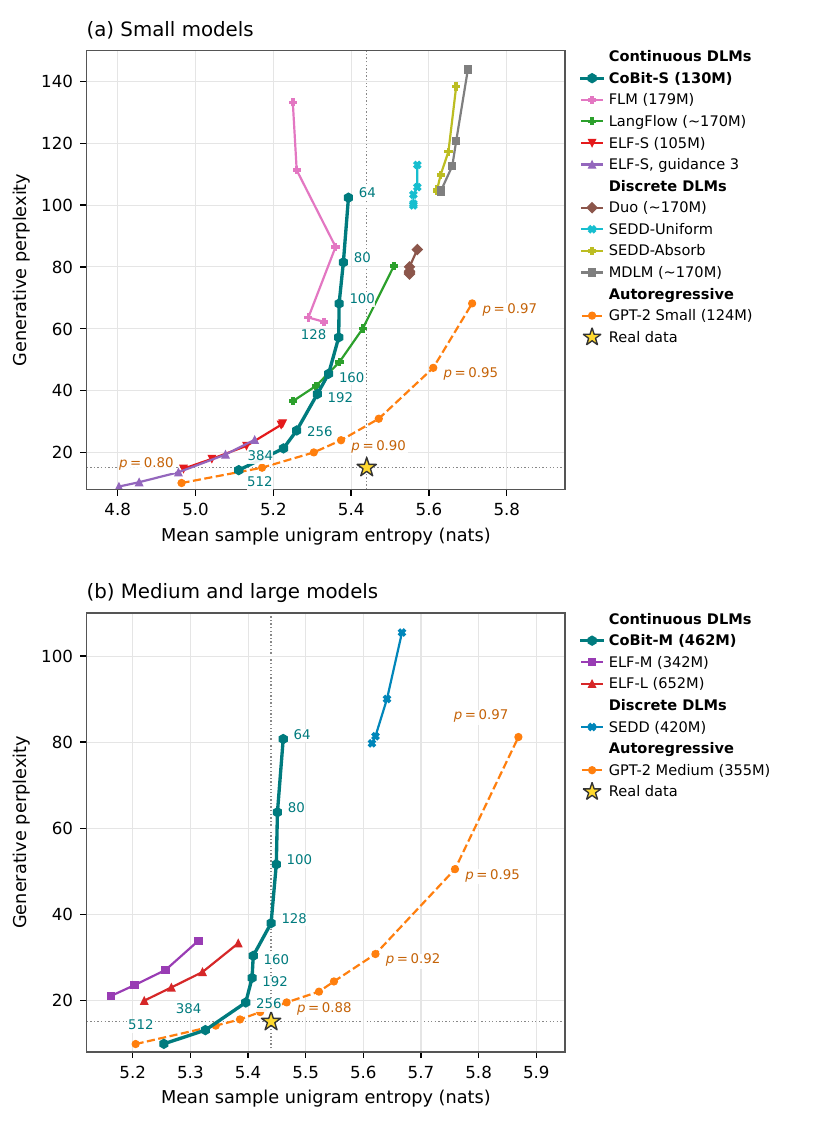}
\caption{OWT \textbf{cross-NFE operating trajectories}: small models above, medium and large models below. CoBit labels give NFEs. Each diffusion curve connects measured settings across sampling budgets. ELF-S is shown with and without self-conditioning guidance 3; ELF-M/L are unguided. Orange AR curves vary nucleus $p\leq0.97$ at temperature 1. Stars and dotted guides mark the real-data reference. The curves describe operating trajectories across NFEs, rather than temperature frontiers at fixed compute. Model sizes are shown in the legends. Additional settings are in \Cref{app:owt_trajectories}.}
\label{fig:frontiers}
\end{figure}

\paragraph{OpenWebText.}
The deterministic 256-NFE sampler obtains 46.32/5.13, with entropy well below the real-data value of 5.44. Full-band churn improves both coordinates to 27.06/5.26. This also improves on LangFlow's 36.53/5.25 reference point using 256 rather than 1,024 NFEs. \Cref{fig:frontiers} shows how the trade-off changes with sampling budget and model scale; uncurated samples are in \Cref{app:qualitative_evaluation}.

\paragraph{Scaling to 462M parameters.}
CoBit-M reaches 19.48/5.396 at 256 NFEs, improving on CoBit-S in both coordinates; at 384 and 512 NFEs it reaches 13.06/5.326 and 9.87/5.254. These settings extend the high-quality region relative to the evaluated ELF-M/L and SEDD-medium settings (\Cref{fig:frontiers}). Pretrained GPT-2 Medium (355M) remains stronger at nearby entropy: 15.54/5.386 at nucleus $p=0.86$. The gap to autoregression narrows at this scale, although the models differ in parameter count and training data.

\paragraph{The role of sampling budget.}
The improvement is strongest at larger budgets. CoBit-S at 384 NFEs obtains 21.26/5.226, compared with unguided ELF-S's 28.89/5.220 at 64 NFEs. At 64 NFEs, CoBit-S instead obtains 102.40/5.393; the temperature-swept MDLM results discussed by \citet{pynadath2025candi} are stronger in this entropy region. Fixed-budget temperature sweeps and trajectories obtained by changing NFE answer different questions. Our figure reports the latter, with fixed-NFE churn sweeps in \Cref{app:sampling_ablations}; full numerical comparisons are in \Cref{app:owt_trajectories}.

\subsection{Verifiable generation and score temperature}
\label{sec:reasoning}
GenPPL and unigram entropy do not determine whether a generated answer is correct. We therefore complement open-ended generation with Sudoku and executable mathematics, using the task protocols of \citet{deschenaux2026continuous}. Full settings and sampling sweeps are in \Cref{app:reasoning_temperature}.

\para{Sudoku.}
A task-specific CoBit backbone (8 layers, width 512) generates the solution while keeping the puzzle prompt clean. We use the same 180-position serialization, 48K/2K training/validation split, and difficulty levels as the published benchmark, with 180 denoising evaluations. The stochastic sampler attains 98.5/91.7/65.9\% exact match on easy/medium/hard puzzles (\Cref{tab:reasoning_sudoku}); deterministic sampling gives 93.2/76.7/40.2\%. The hard-puzzle gain of 25.7 points shows that churn improves globally constrained generation, alongside its effect on language metrics. CoBit uses no guidance, temperature scaling, or verifier-based search on this task.

\begin{table}[t]
\centering\small
\setlength{\tabcolsep}{10pt}
\begin{tabular}{lrrr}
\toprule
Method & Easy & Medium & Hard \\
\midrule
AR (greedy) & 14.6 & 5.1 & 1.0 \\
CANDI & 79.3 & 45.9 & 16.7 \\
FLM & 94.2 & 82.7 & 44.5 \\
$\mathbb{S}$-FLM (best reported) & 94.8 & 85.2 & 45.0 \\
MDLM & 92.0 & 77.1 & 30.2 \\
Duo & 96.3 & 84.7 & 58.4 \\
CoBit (stochastic) & \textbf{98.5} & \textbf{91.7} & \textbf{65.9} \\
\bottomrule
\end{tabular}
\caption{Sudoku exact-match accuracy (\%). Baselines are the configurations of \citet{deschenaux2026continuous}; CoBit averages three seeds and reports the best measured churn setting per difficulty.}
\label{tab:reasoning_sudoku}
\end{table}

\para{TinyGSM training, GSM8K evaluation.}
We train on TinyGSM \citep{liu2023tinygsm} and generate one Python program for each of the 1,319 GSM8K test problems \citep{cobbe2021gsm8k}. The clean question conditions a 512-token sequence; correctness requires the executed program to return the reference answer. CoBit-S uses the same 12-layer, width-768, 12-head backbone dimensions as the published TinyGSM baselines. At their 250K-step training budget, CoBit-S obtains 25.4\% without temperature scaling. We additionally evaluate the same 130M architecture at 500K training steps using score temperature. Configurations are selected by accuracy on TinyGSM validation before GSM8K evaluation (\Cref{app:temperature_statistics}).

\para{Scaling the score.}
Following \citet{yang2026sigma}, we divide the continuous score by $\tau$. For the variance-exploding parameterization, $s_\theta=(D_\theta-x)/\sigma^2$, the equivalent denoiser used in the update is
\begin{equation}
 D_{\theta,\tau}(x,\sigma)=x+\frac{D_\theta(x,\sigma)-x}{\tau}.
 \label{eq:score_temperature}
\end{equation}
This modifies the sampling drift; self-conditioning and final bit readout retain the untempered prediction. With stochastic sampling, applying the scaled drift throughout the trajectory fails at low temperatures on this grid. Numerical diagnostics point to an unstable final Euler update: the sharpened drift can overshoot and flip already-decided bits. We recover performance by leaving the final integration step untempered and applying \cref{eq:score_temperature} at all preceding steps. This modification retains the original grid and the 1,024-evaluation budget; numerical diagnostics are in \Cref{app:reasoning_temperature}.

\begin{table}[t]
\centering\small
\setlength{\tabcolsep}{4pt}
\begin{tabular}{lrrl}
\toprule
Model / training steps & Standard & Low-temperature & Setting \\
\midrule
AR (greedy) & 63.3 & --- & published \\
AR (sampling) & 53.9 & --- & published \\
MDLM / 250K & 18.0 & $\sim33$ & categorical $T=0.1$ \\
Duo / 250K & 17.2 & $\sim36$ & categorical $T=0.1$ \\
$\mathbb{S}$-FLM (best) / 250K & 18.0 & $\sim18$ & published \\
CoBit-S / 250K & 25.4 & --- & untempered \\
\midrule
CoBit-S / 500K & 28.1 & 35.3 & score $\tau=0.3$ \\
\bottomrule
\end{tabular}
\caption{GSM8K executable-code accuracy (\%). Comparators follow the May 2026 benchmark of \citet{deschenaux2026continuous}; approximate low-temperature values are read from their reported sweep. CoBit-S has 130M parameters. The final row reports the validation-selected temperature setting from a single-seed GSM8K evaluation. Full sweeps and uncertainty estimates are in \Cref{app:reasoning_temperature}. Training and sampling controls differ across rows, so this is not a matched-compute comparison.}
\label{tab:reasoning_gsm}
\end{table}

At $\tau=0.3$, CoBit-S accuracy increases from 28.1 to 35.3\%. The executable-program rate also rises from 63.5 to 75.4\%. Thus score temperature brings CoBit-S into the reported 33--36\% range of temperature-tuned MDLM/Duo. The CoBit checkpoint uses 500K training updates, compared with 250K for the published discrete baselines, while the temperature improvement compares the same checkpoint at unchanged inference budget. A conditioning-dropout variant also improves with temperature, but classifier-free guidance adds no gain over its best unguided setting (\Cref{app:reasoning_temperature}).

\subsection{Computational and memory efficiency}
\label{subsec:efficiency_main}
We profile the practical effect of replacing a vocabulary-wide output with the patched bitstream head. Both models use a 12-layer, width-768 Transformer trunk and BF16 mixed precision. The token-space baseline materializes dense vocabulary logits, whereas CoBit predicts $m$ bit logits per position. Synthetic batches isolate model computation from tokenization, data loading, and evaluation; the full setup and absolute measurements are in \Cref{app:efficiency}.

\begin{table}[t]
\centering\small
\begin{tabular}{llrrr}
\toprule
Setting & Mode & Batch & Peak-memory reduction & Throughput gain\\
\midrule
LM1B, $T=128$ & Train & 512 & $1.64\times$ & $2.31\times$\\
OWT, $T=1{,}024$ & Train & 64 & $2.46\times$ & $3.29\times$\\
LM1B, $T=128$ & Generate & 1,024 & $10.02\times$ & $2.00\times$\\
OWT, $T=1{,}024$ & Generate & 64 & $19.25\times$ & $2.65\times$\\
\bottomrule
\end{tabular}
\caption{Matched synthetic profiling against a dense continuous token-space baseline, with a shared 12-layer, width-768 trunk on one GH200. Generation includes 128 denoiser evaluations. Ratios are baseline/CoBit memory and CoBit/baseline throughput; full absolute measurements and exclusions are in \Cref{app:efficiency}.}
\label{tab:efficiency_e2e_main}
\end{table}

At LM1B scale, CoBit reduces peak training memory by $1.64\times$ and increases throughput by $2.31\times$. The gains grow to $2.46\times$ and $3.29\times$ at the longer OWT context and larger code vocabulary. The memory advantage is larger during generation, where vocabulary-wide tensors account for a larger fraction of memory because there are no backward activations or optimizer states. These are full-model measurements, so the shared trunk makes the gains smaller than the raw $V/m$ logit reduction. Their magnitude depends on vocabulary, context, batch size, and implementation; fused or chunked token outputs may reduce the difference. Larger-trunk and autoregressive-latency comparisons remain open.

\subsection{Ablations}
\label{subsec:training_ablations_main}
\Cref{tab:ablation_main} isolates the main training-side choices on LM1B, with self-conditioning enabled throughout. All variants use the same deterministic 256-NFE sampler and the same stochastic search range. The complete protocol and additional sampling experiments are in \Cref{subsec:training_ablations,app:sampling_ablations}.

\begin{table}[t]
\centering\small\setlength{\tabcolsep}{5pt}
\begin{tabular}{lrrrr}
\toprule
& \multicolumn{2}{c}{Deterministic} & \multicolumn{2}{c}{Stochastic}\\
\cmidrule(lr){2-3}\cmidrule(lr){4-5}
Variant & GenPPL $\downarrow$ & Entropy & GenPPL $\downarrow$ & Entropy\\
\midrule
Default: entropic + MSE + filter & $85.37$ & 4.30 & $59.93$ & 4.31\\
EDM allocation + MSE + filter & $81.95$ & 4.29 & $54.04$ & 4.29\\
Entropic + cross-entropy + filter & $91.93$ & 4.29 & $54.68$ & 4.27\\
Entropic + MSE, no filter & $105.96$ & 4.34 & $66.18$ & 4.33\\
\bottomrule
\end{tabular}
\caption{LM1B component ablations at 256 NFEs. Stochastic variants use the same target-churn sweep (0.11--0.20); the sweep uses ten seeds. Dispersion is not reported for these ablation point estimates. The ablation evaluation is separate from the main-checkpoint evaluation in \Cref{tab:main_results}. Real-data entropy is 4.34. Additional protocol and sampler sweeps are in \Cref{tab:training_ablations,app:sampling_ablations}.}
\label{tab:ablation_main}
\end{table}

\paragraph{Training components.}
The matched-filter residual is the clearest training-side contribution. Removing the analytic isolated-bit posterior worsens deterministic GenPPL from 85.37 to 105.96 and stochastic GenPPL from 59.93 to 66.18. The objective and schedule ablations reveal a quality--diversity trade-off. Cross-entropy and EDM noise allocation give lower stochastic GenPPL, but also lower entropy. The entropy-rate default reaches 4.31, closer to the real-data reference of 4.34 than EDM allocation (4.29) or cross-entropy (4.27). We use this data-driven schedule across the tasks, without assuming that one objective or allocation is best in every regime.

\paragraph{Why not discrete diffusion on bits?}
A SEDD-style discrete diffusion baseline on LM1B bitstreams retains the compact $\mathcal{O}(\log V)$ output but substantially underperforms token-level SEDD: at 256 NFEs, GenPPL is 285.09 versus 126.28 at comparable entropy (\Cref{app:discrete_bitstream_baseline}). This shows that reducing the output dimension alone is insufficient; the corruption process and denoiser must make effective use of the representation.

We evaluate the released binary MDM-Prime-v2 checkpoint \citep{chao2026mdmprimev2}, a 132M model with a vocabulary-wide joint decoder. At 256 NFEs it obtains 270.88/5.400 under our evaluation; fp64 reruns across 64--512 NFEs change GenPPL by less than 5\%. This evaluates one released checkpoint; its protocol is in \Cref{app:mdm_prime}.

\paragraph{Sampler components.}
The fixed-checkpoint comparison in \Cref{tab:main_results} identifies stochastic churn as the largest improvement. At fixed NFE, moderate full-band churn improves both OWT GenPPL and entropy, whereas excessive churn continues to raise entropy at the cost of GenPPL (\Cref{fig:owt_churn_ablation}). The entropy-rate grid also improves over the Karras grid in the unconditional experiments, and broad stochasticity windows work better than narrowly localized ones (\Cref{app:sampling_ablations}). These observations support the role of the grid in allocating both solver resolution and effective stochastic correction. The best allocation can still depend on the task and checkpoint: the conditioning-dropout reasoning variant uses a Karras grid.

\section{Discussion and conclusion}
\label{sec:discussion}
We introduced CoBit, a continuous diffusion language model over fixed-width semantic bitstreams. The results point to the interaction between representation, parameterization, and sampling as central to continuous language modeling. The bitstream representation gives a compact prediction interface, the matched-filter residual supplies local Gaussian denoising structure, and stochastic churn on the entropy-rate grid improves the quality--diversity trade-off. The sampler remains the largest empirical improvement, both in unconditional generation and in exact Sudoku solving. Scaling the model improves the OWT trade-off further, while the corrected score-temperature procedure brings CoBit-S into the reported range of temperature-tuned discrete diffusion on GSM8K.

The logarithmic output size may become particularly useful for larger multilingual or multimodal vocabularies, where a dense vocabulary head is increasingly costly. Our matched-trunk measurements establish practical memory and throughput benefits at the scales studied here; testing this advantage alongside generation quality at larger scales remains future work. The remaining autoregressive gap and the weak low-NFE regime also motivate faster samplers, distillation, and more stable temperature control. CoBit provides a compact continuous formulation in which to pursue these directions.

\section{Limitations}
\label{sec:limitations}
Our largest trained model has 462M parameters and the longest trunk sequence has 1,024 positions. Larger models, multilingual vocabularies, and alternative code assignments remain untested. GenPPL and unigram entropy capture only part of generation quality: the uncurated samples still contain repetition and inconsistency, and we have not measured corpus-level coherence or factuality. Comparisons also differ in tokenizer, model size, training budget, and decoding settings; the new CoBit-S GSM8K checkpoint uses 500K updates versus 250K for the published baselines. Its temperature sweep uses one generation seed, so the appendix bootstrap intervals measure variation across test problems rather than training or sampling-seed variability. Score scaling controls the sampling dynamics without guaranteeing an exactly tempered sequence distribution, and strong scaling can destabilize numerical integration. Finally, the compact output does not remove the cost of iterative denoising: low-NFE quality remains a limitation, and the measured systems gains have not been validated at billion-parameter trunk scale.

\label{maintext:end}
\clearpage
\label{backmatter:start}
\begin{ack}
We thank Rares Dimitrie Grozavescu for identifying and correcting a bug in the temperature-scaling implementation, which enabled the improved GSM8K results reported in this paper.

This project received an allocation of 10,000 GPU hours through the UKRI AIRR Gateway programme. We used the Isambard-AI supercomputer \citep{mcintoshsmith2024isambard}, a National AI Research Resource operated at the University of Bristol. The facility is supported by DBIST (the UK government's Department for Business, Innovation, Science and Trade), through UKRI, and by STFC (facility funding reference ST/AIRR/I-A-I/1023).

\end{ack}

\bibliographystyle{plainnat}
\bibliography{references}
\clearpage
\appendix
\section{Additional method details}
\label{app:method_details}

\subsection{Matched-filter derivation}
\label{app:matched_filter_derivation}

We derive the analytic matched filter used in \Cref{eq:matched_filter_residual}.
Consider a single clean bit
\[
    x_0\sim\mathrm{Bern}(1/2),
    \qquad
    x=x_0+\sigma\epsilon,
    \qquad
    \epsilon\sim\mathcal{N}(0,1).
\]
The posterior log-odds are
\begin{align}
    \log\frac{p(x_0=1\mid x,\sigma)}
             {p(x_0=0\mid x,\sigma)}
    &=
    \log\frac{p(x\mid x_0=1,\sigma)}
             {p(x\mid x_0=0,\sigma)}
    \\
    &=
    -\frac{(x-1)^2}{2\sigma^2}
    +\frac{x^2}{2\sigma^2}
    \\
    &=
    \frac{x-\frac12}{\sigma^2}.
\end{align}
Therefore the optimal independent-bit denoiser is
\begin{equation}
    D_{\mathrm{ind}}(x,\sigma)
    =
    \sigmoid\!\left(
        \frac{x-\frac12}{\sigma^2}
    \right).
    \label{eq:app_independent_bit_denoiser}
\end{equation}
In a language model, bits are not independent: neighboring bits encode token identities and distant tokens constrain which codes are plausible.
The matched-filter residual parameterization uses the closed-form independent posterior as a local baseline and asks the network to predict the contextual residual:
\begin{equation}
    \ell_\theta(x_\sigma,\sigma)
    =
    r_\theta(x_\sigma,\sigma,x_{\mathrm{sc}})
    +
    \clip\!\left(
        \frac{x_\sigma-\frac12}{\sigma^2},
        -C,C
    \right),
    \qquad
    D_\theta=\sigmoid(\ell_\theta).
    \label{eq:app_matched_filter_residual}
\end{equation}
The clipping constant \(C=30\) is used for numerical stability.
This decomposition supplies the isolated uniform-prior posterior analytically; the learned residual can correct for context, nonuniform bit priors, and the clipped baseline.

\subsection{Vocabulary-boundary scaling}
\label{app:boundary_scaling}

Most diffusion language models retain a vocabulary-sized prediction interface.
Simplex and one-hot continuous models diffuse in vocabulary-dimensional states; embedding-space continuous models reduce the state dimension but still commonly predict vocabulary logits; and discrete token-state models parameterize denoising distributions or transition scores over \(V\) tokens.
When implemented with dense vocabulary logits, their output/loss boundary scales as \(\mathcal{O}(TV)\).

\begin{table}[h]
\centering
\small
\setlength{\tabcolsep}{6pt}
\begin{tabular}{@{}l c c >{\raggedright\arraybackslash}p{0.28\linewidth}@{}}
\toprule
\textbf{Model Family} &
\textbf{State Representation} &
\textbf{Trunk Length} &
\textbf{Output/Loss Boundary} \\
\midrule
Continuous Simplex / One-Hot
& \(\mathbb{R}^{B\times T\times V}\)
& \(T\)
& \(\mathbb{R}^{B\times T\times V}\) logits \\

Continuous Embedding-Space
& \(\mathbb{R}^{B\times T\times d}\)
& \(T\)
& \(\mathbb{R}^{B\times T\times V}\) logits \\

Discrete Token-State
& Discrete IDs
& \(T\)
& \(\mathbb{R}^{B\times T\times V}\) logits/scores \\
\midrule
Naive Bit-level Diffusion
& \(\mathbb{R}^{B\times Tm}\)
& \(Tm\)
& \(\mathbb{R}^{B\times Tm}\) logits \\

\textbf{\methodname{}}
& \(\mathbb{R}^{B\times Tm}\)
& \(T\)
& \(\mathbb{R}^{B\times Tm}\) logits \\
\bottomrule
\end{tabular}
\caption{\textbf{Structural boundary scaling.}
Here \(m=\lceil\log_2 V\rceil\).
A dense categorical head has an \(\mathcal{O}(V)\) output/loss boundary per token.
A naive bit-level model removes this boundary but lengthens the Transformer sequence to \(Tm\).
\methodname{} patches the \(m\) bits of each token into one trunk element, preserving the code-sequence length \(T\) while reducing the boundary to \(\mathcal{O}(\log V)\) logits per token.}
\label{tab:boundary_scaling_app}
\end{table}

\subsection{Why binary within a factorized digit interface?}
\label{app:binary_boundary}
A base-$q$ code needs $d_q=\lceil\log_q V\rceil$ digits. A categorical digit has $q-1$ independent log-odds, so a factorized digit head uses
\[
 N_q=(q-1)\lceil\log_q V\rceil.
\]
For integer $q\geq2$, $q\leq2^{q-1}$. Hence $V\leq q^{d_q}\leq2^{(q-1)d_q}$ and $N_q\geq\lceil\log_2V\rceil=N_2$. Binary therefore minimizes the number of independent digit log-odds within this family (ties are possible). At $V=30{,}522$, the counts are 15, 20, and 24 for bases 2, 3, and 4. A conventional $q$-logit softmax may materialize one redundant logit per digit.

This is an output-interface comparison, not an optimality result for expressiveness or sample quality. Higher-base digits directly represent dependencies within a larger local alphabet; CoBit instead learns cross-bit dependencies through shared context and iterative refinement. Dense random Gaussian codes also offer a compact state, but require a different codeword decoder and do not give the same isolated-bit matched filter. Neither alternative has been empirically ablated here.

\subsection{Decoding}
\label{app:decoding_details}

Generated analog bits are converted to binary codes by thresholding the final
bit probabilities, equivalently the final logits. The bits are grouped into
fixed-width codes and decoded by the dataset-specific inverse mapping.

For LM1B, each \(m=15\)-bit group represents a raw
\texttt{bert-base-uncased} tokenizer ID. For OWT, we support both the direct
GPT-2-token interface and the fixed-length \texttt{gpt2id\_bpe16} codec used in
the main OWT runs. In the codec case, each \(m=16\)-bit group is first decoded to
a \texttt{gpt2id\_bpe16} code token, then inverted back to GPT-2 token IDs and
decoded to text. GenPPL is always evaluated on the final decoded text; entropy
is computed on recovered evaluation token IDs---\texttt{bert-base-uncased} for
LM1B and GPT-2 for OWT---not on intermediate code-token IDs.

The OWT mapping is a second-stage codec over GPT-2 ID sequences, rather than a semantic embedding or a change to the external evaluation tokenizer. Its inverse recovers GPT-2 IDs before text decoding. The fixed length of 1024 refers to code tokens, so it does not imply a one-to-one correspondence with 1024 GPT-2 tokens. Reported entropy is consequently measured after inversion, on the recovered evaluation tokens. Fixed-width binary codes can represent values outside a tokenizer's vocabulary when its size is not a power of two; thresholding alone does not guarantee a valid tokenizer ID.

\section{Training and architecture details}
\label{app:training_architecture_details}

\subsection{Entropy-rate schedule details}
\label{app:entropy_schedule_details}

We use the online entropy-rate noise allocation strategy proposed by
\citet{raya2026infonoise}, adapted to continuous diffusion over bitstreams.
For VE corruption,
\[
    x_\sigma=x_0+\sigma\epsilon,
    \qquad
    \epsilon\sim\mathcal{N}(0,I),
\]
the conditional entropy rate is
\[
    h_{\log}(\sigma)
    :=
    \frac{d}{d\log\sigma}H(x_0\mid x_\sigma).
\]
By the I--MMSE relation and denoising-score-matching identities, this rate is related to a noise-rescaled denoising error.
Thus, the losses already computed during training provide a practical online proxy for the information profile of the forward process.

During training, we maintain a FIFO buffer of pairs \((\sigma,e)\), where
\[
    e(\sigma)
    =
    \frac{1}{S}
    \sum_{i=1}^S
    \left(D_{\theta,i}(x_\sigma,\sigma)-x_{0,i}\right)^2
\]
is the unweighted bitwise denoising error.
Since our schedule is parameterized over log-noise, we use
\begin{equation}
    \widehat h_{\log}(\sigma)
    =
    \frac{e(\sigma)}{\sigma^2+\varepsilon}
    \label{eq:app_log_entropy_rate_proxy}
\end{equation}
as the online proxy for entropy production per unit \(\log\sigma\).
The small constant \(\varepsilon\) is used only for numerical stability.

The interval \([\log\sigma_{\min},\log\sigma_{\max}]\) is divided into log-spaced bins.
Let \(\bar h_k\) be the average value of \(\widehat h_{\log}\) in bin \(k\), and let \(\sigma_k\) be the bin midpoint.
We define the bin sampling probabilities by
\begin{equation}
    q_k
    \propto
    g(\sigma_k;c,n)\,
    \bar h_k^\alpha,
    \qquad
    g(\sigma;c,n)
    =
    \frac{\sigma^n}{\sigma^n+c^n}.
    \label{eq:app_entropy_schedule_bins}
\end{equation}
The factor \(g\) is a smooth low-noise gate that prevents the sampler from over-concentrating on nearly resolved bits.
In the final runs, we use
\[
    \alpha=\frac12,
    \qquad
    c=0.1,
    \qquad
    n=3.
\]

Equivalently, this defines a normalized schedule density over \(u=\log\sigma\),
\begin{equation}
    \pi_\alpha(u)
    =
    \frac{
        g(e^u;c,n)\,
        \widehat h_{\log}(e^u)^\alpha
    }{
        \int_{\log\sigma_{\min}}^{\log\sigma_{\max}}
        g(e^v;c,n)\,
        \widehat h_{\log}(e^v)^\alpha\,dv
    }.
    \label{eq:app_normalized_entropy_rate_density}
\end{equation}
Training begins from the base EDM log-normal noise distribution.
We warm up for 40K steps, linearly transition to the entropy-rate distribution over the next 10K steps, and then sample noise levels from the entropy-rate schedule.
For sampling, we construct a sigma grid by approximately inverting the CDF of \(\pi_\alpha\), following the entropic integration viewpoint of \citet{stancevic2025entropic}.

\subsection{Architecture implementation details}
\label{app:architecture_details}

The model is implemented as a sequence diffusion transformer specialized to continuous bitstreams.
The input sequence has \(S=Tm\) analog-bit positions, but the Transformer trunk operates at semantic length \(T\) by patching the \(m\) bits of each token or code token into one trunk element.

For continuous bits, the noisy bit input is centered by the data center \(1/2\) and scaled by
\[
    c_{\mathrm{in}}(\sigma)
    =
    (\sigma^2+\sigma_{\mathrm{data}}^2)^{-1/2}.
\]
When self-conditioning is enabled, the self-conditioning prediction is centered and scaled in the same way and concatenated with the noisy-bit embedding.
We use local Fourier features within each bit patch, RoPE in the Transformer trunk, and no absolute global Fourier features in the final configuration.

The trunk uses 12 Transformer blocks with hidden width 768, 12 attention heads, feed-forward width 3072, AdaLN-zero time conditioning, SwiGLU feed-forward layers, dropout \(0.1\), and SDPA/FlashAttention kernels when available.
The time embedding is a sinusoidal embedding of \(\log\sigma\), projected into the model width and used by AdaLN-zero modules.

The output head is an optimal skip MLP head.
For each trunk token, a patch adapter expands the global token representation into \(m\) per-bit hidden states.
In parallel, a local adapter embeds the current noisy bit and self-conditioning features.
The global and local per-bit features are added, modulated by a time-conditioned AdaLN-zero block, and mapped to one residual logit per bit.
The final logit adds the analytic matched-filter term from \Cref{eq:matched_filter_residual}.

\subsection{Asymmetric time intervals}
\label{app:ati_details}

We support the asymmetric time-interval label shift introduced in Analog Bits \citep{chen2022analogbits}.
When evaluating the denoiser at state noise \(\sigma_{\mathrm{state}}\), we optionally shift the sigma label toward a noisier adjacent level in log-space:
\begin{equation}
    \sigma_{\mathrm{eval}}
    =
    \exp\!\left(
    (1-\eta)\log\sigma_{\mathrm{state}}
    +
    \eta\log\sigma_{\mathrm{noisier}}
    \right).
    \label{eq:app_ati_eval_sigma}
\end{equation}
Here \(\eta=0\) recovers the standard sampler.
Positive \(\eta\) can improve some low-NFE deterministic settings, especially when combined with self-conditioning, but its effect is weaker and less consistent once stochastic churn is enabled.
Therefore all main 256-NFE stochastic results use \(\eta=0\).

\subsection{Final LM1B configuration}
\label{app:lm1b_config}

\begin{itemize}[leftmargin=1.25em,itemsep=0.2em,topsep=0.2em]
    \item Dataset: LM1B, \texttt{bert-base-uncased}, packed 128-token blocks.
    \item Representation: raw binary tokenizer IDs, 15 bits/token, sequence length 1920 bits.
    \item Model: SDT, 12 blocks, width 768, 12 heads, FF width 3072, patch size 15.
    \item Head: optimal skip MLP, hidden size 128.
    \item Diffusion: \(\sigma_{\min}=0.002\), \(\sigma_{\max}=80\), \(\rho=7\), \(\sigma_{\mathrm{data}}=0.5\), data center \(0.5\).
    \item Training: binary score matching, EDM weighting, global batch size 512, AdamW, learning rate \(3\times10^{-4}\), cosine decay, 2500 warmup steps, 1M optimizer steps.
    \item Entropy-rate schedule: online sqrt-rate allocation, \(\alpha=1/2\), low-noise gate \(c=0.1,n=3\), 40K warmup and 10K transition.
    \item Self-conditioning: enabled, training probability \(0.5\), carry-mode at sampling.
    \item Logit parameterization: matched-filter residual, center \(0.5\), scale \(1.0\), clip \(30\).
\end{itemize}

\subsection{Small-model OpenWebText configuration}
\label{app:owt_config}

\begin{itemize}[leftmargin=1.25em,itemsep=0.2em,topsep=0.2em]
    \item Dataset: OpenWebText with GPT-2 tokenizer, train split \texttt{train[:-100000]}, validation split \texttt{train[-100000:]}.
    \item Codec: fixed-length \texttt{gpt2id\_bpe16} second-stage code tokenizer, 1024 code tokens.
    \item Representation: raw binary code-token IDs, 16 bits/token, sequence length 16384 bits.
    \item Model: same SDT trunk as LM1B, patch size 16.
    \item Diffusion, optimizer, self-conditioning, matched-filter residual, and entropy-rate schedule: same as LM1B.
    \item Training: global batch size 512, AdamW, learning rate \(3\times10^{-4}\), cosine decay, 2500 warmup steps.
    \item Main reported checkpoint: 750K optimizer steps.
\end{itemize}

\subsection{Medium-model OpenWebText configuration}
\label{app:owt_medium_config}

CoBit-M uses the same OWT split, \texttt{gpt2id\_bpe16} codec, 1024-code-token representation, weighted denoising objective, self-conditioning, matched-filter residual, and entropy-rate schedule as CoBit-S. Architecture and optimization settings are:
\begin{itemize}[leftmargin=1.25em,itemsep=0.2em,topsep=0.2em]
    \item Model: flat SDT with 24 blocks, width 1024, 16 attention heads, feed-forward width 4096, patch size 16; approximately 462M parameters.
    \item Training: AdamW, learning rate \(2\times10^{-4}\), 5000 warmup steps, global batch size 512, EMA decay 0.9999, BF16 mixed precision.
    \item Checkpoint: 750K optimizer steps, using EMA weights for evaluation.
    \item Sampling: \texttt{ddim\_entropic}, carry-mode self-conditioning, terminal sigma 0.08, guidance scale zero, \(S_{\mathrm{noise}}=1.003\), full entropy-CDF window.
    \item Evaluation: 1024 samples per operating point, external \texttt{gpt2-large} GenPPL, and unigram entropy on inverse-decoded GPT-2 tokens.
    \item Sampling sweep: target NFEs \(\{64,80,100,128,160,192,256,512\}\), with the churn budget parameterized as \(S_{\mathrm{churn}}=\gamma(\mathrm{NFE}-1)\). The near-real-entropy 256-NFE point uses target \(\gamma=0.21\), hence \(S_{\mathrm{churn}}=53.55\), and yields GenPPL 19.48 and entropy 5.40.
\end{itemize}
This scaling comparison changes the model size together with the listed optimization and sampling settings; it is not an isolated parameter-count ablation.

\section{Additional generation comparisons}
\label{app:owt_trajectories}
\subsection{OpenWebText operating trajectories}
\Cref{fig:frontiers} shows cross-NFE operating trajectories. \Cref{tab:owt-small-full} summarizes three sampling budgets, and \Cref{tab:owt-medium-full} gives the autoregressive nucleus curves. Complete curve data, including intermediate and 1,024-NFE settings, are provided as ancillary files.

CoBit settings are selected from churn sweeps. At 256 NFEs, CoBit-S uses target churn 0.13 and averages ten seeds (27.06/5.260); the other cells are individual operating points. CoBit-M uses target churn 0.21 at 256 NFEs; target churn 0.13 gives 18.47/5.378. Churn values are unavailable for the remaining NFE settings, limiting exact reproduction of those points.

ELF-S/M/L were evaluated with released code, both unguided and with self-conditioning guidance 3. The OWT GPT-2 models are pretrained and sampled at temperature 1 with the stated nucleus probabilities. FLM, LangFlow, MDLM, Duo and SEDD are recorded comparison operating points \citep{lee2026flm,chen2026langflow,sahoo2024mdlm,sahoo2025duo,lou2024discrete,hu2026elf}. The curves do not represent optimized temperature frontiers at fixed NFE. OWT evaluation uses 1,024 samples, GPT-2-Large GenPPL and GPT-2-token unigram entropy; the real-data reference is 15.06/5.44.

\begin{table}[ht]
\centering\small
\setlength{\tabcolsep}{7pt}
\begin{tabular*}{\linewidth}{@{\extracolsep{\fill}}lrrr@{}}
\toprule
& \multicolumn{3}{c}{GenPPL $\downarrow$ / entropy}
\\
Model & 64 NFE & 256 NFE & 512 NFE\\
\midrule
\multicolumn{4}{l}{\textit{Small models}}\\
CoBit-S (130M) & 102.40 / 5.393 & 27.06 / 5.260 & 14.22 / 5.111\\
FLM (179M) & 133.29 / 5.250 & 86.50 / 5.360 & 63.63 / 5.290\\
LangFlow ($\sim$170M) & 80.34 / 5.510 & 49.24 / 5.370 & 41.52 / 5.310\\
ELF-S (105M) & 28.89 / 5.220 & 17.87 / 5.042 & 14.69 / 4.969\\
ELF-S + guidance 3 & 19.28 / 5.077 & 10.32 / 4.855 & 8.91 / 4.803\\
Duo ($\sim$170M) & 85.62 / 5.570 & 78.62 / 5.550 & 78.14 / 5.550\\
SEDD-Uniform ($\sim$170M) & 113.02 / 5.570 & 103.41 / 5.560 & 100.44 / 5.560\\
SEDD-Absorb ($\sim$170M) & 138.42 / 5.670 & 109.82 / 5.630 & 104.45 / 5.620\\
MDLM ($\sim$170M) & 143.88 / 5.700 & 112.70 / 5.660 & 104.43 / 5.630\\
\midrule
\multicolumn{4}{l}{\textit{Medium and large models}}\\
CoBit-M (462M) & 80.73 / 5.461 & 19.48 / 5.396 & 9.87 / 5.254\\
ELF-M (342M) & 33.86 / 5.314 & 23.51 / 5.203 & 20.99 / 5.162\\
ELF-M + guidance 3 & 21.89 / 5.188 & 13.82 / 5.016 & 11.76 / 4.962\\
ELF-L (652M) & 33.25 / 5.383 & 22.98 / 5.267 & 19.91 / 5.220\\
ELF-L + guidance 3 & 23.58 / 5.280 & 14.44 / 5.099 & 11.95 / 5.027\\
SEDD (420M) & 105.48 / 5.667 & 81.38 / 5.621 & 79.72 / 5.615\\
\bottomrule
\end{tabular*}
\caption{OWT diffusion-model operating points at three selected NFE budgets. Each cell gives GenPPL and mean sample unigram entropy in nats. Guidance is self-conditioning guidance with scale 3. The real-data reference is 15.06 / 5.44. Full numerical trajectories are included in the accompanying plotting data.}
\label{tab:owt-small-full}
\end{table}

\begin{table}[ht]
\centering\small
\setlength{\tabcolsep}{12pt}
\begin{tabular}{rrr}
\toprule
Nucleus $p$ & GPT-2 Small (124M) & GPT-2 Medium (355M)\\
& \multicolumn{2}{c}{GenPPL $\downarrow$ / entropy}\\
\midrule
0.80 & 10.06 / 4.964 & 9.82 / 5.205\\
0.85 & 15.02 / 5.171 & 14.10 / 5.344\\
0.86 & -- & 15.54 / 5.386\\
0.87 & -- & 17.25 / 5.421\\
0.88 & 19.96 / 5.304 & 19.50 / 5.467\\
0.89 & -- & 22.00 / 5.523\\
0.90 & 23.91 / 5.374 & 24.37 / 5.549\\
0.92 & 30.87 / 5.471 & 30.75 / 5.621\\
0.95 & 47.34 / 5.611 & 50.48 / 5.759\\
0.97 & 68.19 / 5.711 & 81.19 / 5.869\\
1.00 & -- & 255.78 / 6.125\\
\bottomrule
\end{tabular}
\caption{OWT autoregressive nucleus-sampling curves. Both pretrained GPT-2 models use temperature 1. Cells give GenPPL and entropy under the same evaluation convention as Table~\ref{tab:owt-small-full}; a dash denotes an unevaluated setting.}
\label{tab:owt-medium-full}
\end{table}

\FloatBarrier

\subsection{Binary MDM-Prime-v2 comparison}
\label{app:mdm_prime}
The released binary MDM-Prime-v2 checkpoint has 12 layers, width 768 and approximately 132M parameters; its length-16 base-2 code is paired with a joint vocabulary decoder \citep{chao2026mdmprimev2}. We generate 1,024 OWT samples per operating point, map invalid decoded IDs to EOS, and apply our external evaluator. This invalid-ID policy applies to this baseline evaluation. These measurements concern generated samples from that checkpoint, not the likelihood results affected by the authors' published erratum. The fp64 runs address sampling precision but do not exhaust possible decoder tuning.
\begin{table}[ht]
\centering\small
\begin{tabular}{rrrr}
\toprule
NFE & fp32 GenPPL & fp32 entropy & fp64 GenPPL\\
\midrule
64 & 246.73 & 4.529 & 238.38\\
128 & 217.99 & 4.631 & 221.46\\
256 & 270.88 & 5.400 & 282.85\\
512 & 285.33 & 5.623 & 280.87\\
\bottomrule
\end{tabular}
\caption{Binary MDM-Prime-v2 operating points. The largest fp64 change relative to fp32 is approximately 4.4\%, at 256 NFEs. CoBit-S's 256-NFE point is 27.06/5.260. Different decoders and coding protocols limit causal attribution to continuous versus discrete corruption alone.}
\end{table}

\FloatBarrier
\subsection{LM1B sampling trajectories and nucleus baseline}
The GPT-S autoregressive baseline is trained from scratch on LM1B and evaluated with nucleus probabilities $p\in\{0.90,0.95,0.98\}$. Its GenPPL/entropy pairs are 43.00/4.299, 52.80/4.327 and 61.20/4.342, respectively. The real-data reference is 53.06/4.34.

\begin{figure}[ht]
\centering\includegraphics[width=\linewidth]{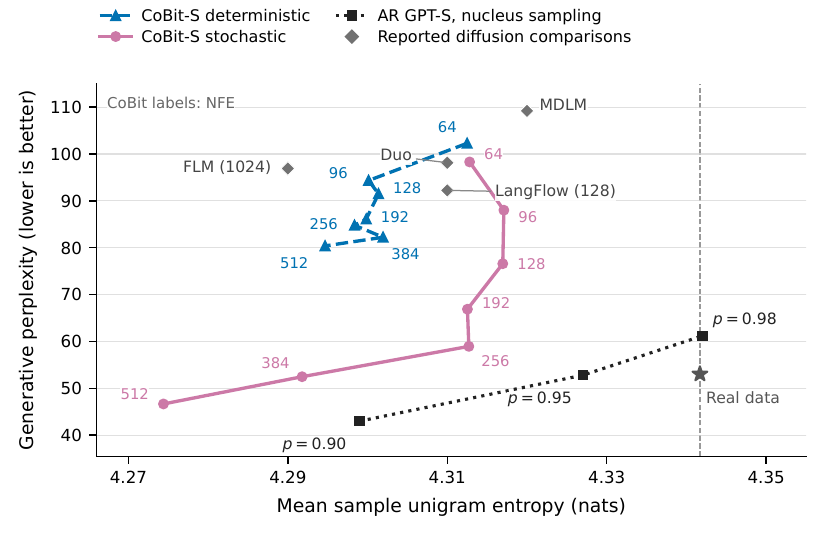}
\caption{LM1B generation comparison. CoBit curves connect settings from the original deterministic and stochastic NFE sweeps; labels give NFEs. The GPT-S curve varies nucleus $p$ at a fixed trained checkpoint. Other diffusion baselines are representative reported operating points. The real-data reference is GenPPL 53.06 at entropy 4.34.}
\label{fig:lm1b_frontier}
\end{figure}

\section{Discrete bitstream diffusion baseline}
\label{app:discrete_bitstream_baseline}

A natural diagnostic baseline is to keep the fixed-width bitstream
representation, but replace the continuous Gaussian corruption process with a
discrete diffusion process over bits. This preserves the favourable
\(\mathcal{O}(\log V)\) output boundary, since each token is represented by
\(m=\lceil \log_2 V\rceil\) binary variables, but removes the continuous analog
state space used by \methodname{}.

We evaluate this option using a SEDD-style absorbing discrete diffusion baseline
\citep{lou2024discrete} on LM1B bitstreams. For comparison, we also train a
matched token-level SEDD baseline on the same dataset. Both baselines use
\texttt{bert-base-uncased}, sequence length \(T=128\), a 12-layer SDT trunk with
width 768, 12 attention heads, feed-forward width 3072, RoPE, AdaLN time
conditioning, SwiGLU activations, dropout \(0.1\), BF16 training, AdamW with
learning rate \(3\times 10^{-4}\), and a log-linear absorbing discrete diffusion
schedule. The token baseline operates over the BERT vocabulary plus an absorbing
mask state, while the bitstream baseline operates directly over the
corresponding fixed-width bit representation. Both models are evaluated using
the same Tweedie sampler sweep, 1024 generated samples, and the same external
\texttt{gpt2-large} GenPPL evaluator.

\begin{table}[h]
\centering
\small
\setlength{\tabcolsep}{5pt}
\begin{tabular}{rcccc}
\toprule
\multirow{2}{*}{\textbf{NFE}} &
\multicolumn{2}{c}{\textbf{Discrete diffusion on bits}} &
\multicolumn{2}{c}{\textbf{Discrete diffusion on tokens}} \\
\cmidrule(lr){2-3} \cmidrule(lr){4-5}
& \textbf{GenPPL} $\downarrow$ & \textbf{Entropy}
& \textbf{GenPPL} $\downarrow$ & \textbf{Entropy} \\
\midrule
16  & 987.67 & 4.394 & 184.91 & 4.293 \\
24  & 660.13 & 4.369 & 166.46 & 4.293 \\
32  & 527.97 & 4.355 & 157.81 & 4.292 \\
48  & 414.47 & 4.338 & 147.97 & 4.291 \\
64  & 371.53 & 4.333 & 143.16 & 4.290 \\
128 & 313.37 & 4.320 & 134.86 & 4.291 \\
256 & 285.09 & 4.314 & 126.28 & 4.290 \\
512 & 272.18 & 4.311 & 118.40 & 4.290 \\
\bottomrule
\end{tabular}
\caption{\textbf{Discrete diffusion on bits versus tokens on LM1B.}
Both baselines use a SEDD-style absorbing discrete diffusion process and the
same 12-layer SDT trunk. The bitstream baseline preserves the favourable
\(\mathcal{O}(\log V)\) output boundary, but substantially underperforms the
token-level discrete baseline across all NFE budgets. At high NFE, the
token-frequency entropies are comparable, indicating that the gap is not simply
explained by entropy collapse.}
\label{tab:discrete_bitstream_baseline}
\end{table}

The bitstream discrete baseline is substantially worse than the token-level
discrete baseline across the full sampling range. At 256 NFEs, discrete
diffusion on bits obtains GenPPL \(285.09\), compared with \(126.28\) for
token-level discrete diffusion. At 512 NFEs, the gap remains similarly large:
\(272.18\) versus \(118.40\). The token-frequency entropies are close in the
high-NFE regime, so the difference is primarily reflected in external sample
quality rather than in a simple entropy-collapse artifact.

This diagnostic suggests that the compact bit boundary alone is not sufficient
to make bit-level discrete diffusion competitive in this setting. This motivates
our choice to use continuous Gaussian diffusion over bitstreams, rather than a
discrete bit-level absorbing process.
\section{Stochastic churn as an information-adaptive reverse SDE}
\label{app:stochastic_sde_analysis}

We give a continuous-time interpretation of the stochastic sampler used in
\methodname. The purpose is to explain why full-band EDM-style churn becomes
information-adaptive when combined with the entropy-rate sampling grid defined
by the normalized schedule density \(\pi_\alpha\) (detailed in \cref{app:entropy_schedule_details}). 
We focus on the clean limiting structure of the method: full-band stochasticity, 
\(S_{\mathrm{noise}}=1\), no asymmetric-time-interval label shift, and the small-step regime 
where the EDM churn parameter is not clipped. These assumptions remove finite-step
implementation details while preserving the mechanism relevant to the sampler
used in practice.

\subsection{Discrete implementation of EDM churn}

For completeness, we first detail the discrete EDM-style churn step used in
practice \citep{karras2022edm}. Before a deterministic step from \(\sigma_i\) to \(\sigma_{i+1}\), the sampler optionally increases the current noise level to
\begin{align}
    \hat\sigma_i &= (1+\gamma_i)\sigma_i,\\
    \hat x_i
    &=
    x_i
    +
    S_{\mathrm{noise}}
    \sqrt{\hat\sigma_i^2-\sigma_i^2}\,z_i,
    \qquad z_i\sim\mathcal{N}(0,I),
    \label{eq:app_stochastic_churn}
\end{align}
and then applies the deterministic update from \((\hat x_i,\hat\sigma_i)\) to \(\sigma_{i+1}\). In our experiments, we use \(S_{\mathrm{noise}}=1.003\). The raw churn amount is
\begin{equation}
    \gamma_i
    =
    \begin{cases}
    \min(S_{\mathrm{churn}}/N,\sqrt{2}-1),
    & \sigma_i\in\mathcal{W},\\
    0,
    & \text{otherwise},
    \end{cases}
    \label{eq:app_raw_churn}
\end{equation}
where \(N\) is the number of sampling intervals and \(\mathcal{W}\) is the stochasticity window.

\subsection{VE corruption and probability-flow sampling}

Consider the VE corruption process
\begin{equation}
    x_\sigma = x_0+\sigma\epsilon,
    \qquad
    \epsilon\sim\mathcal{N}(0,I),
    \label{eq:app_ve_corruption}
\end{equation}
where \(x_0\in\bits^S\) for binary bitstreams. Let \(p_\sigma\) denote the
density of \(x_\sigma\). The denoiser predicts
\begin{equation}
    D_\theta(x,\sigma)
    \approx
    \mathbb{E}[x_0\mid x_\sigma=x].
\end{equation}
By Tweedie's formula, this induces the score estimate
\begin{equation}
    s_\theta(x,\sigma)
    =
    \frac{D_\theta(x,\sigma)-x}{\sigma^2}.
    \label{eq:app_score_from_denoiser}
\end{equation}
The deterministic probability-flow sampler follows
\begin{equation}
    dx=-\sigma s_\theta(x,\sigma)\,d\sigma.
    \label{eq:app_pf_ode}
\end{equation}
For a decreasing schedule
\[
    \sigma_0>\sigma_1>\cdots>\sigma_N,
\]
define
\[
    \Delta_i=\sigma_i-\sigma_{i+1}>0.
\]
An Euler step of \cref{eq:app_pf_ode} gives
\begin{equation}
    x_{i+1}
    =
    x_i+\Delta_i\sigma_i s_\theta(x_i,\sigma_i).
    \label{eq:app_det_euler}
\end{equation}

\subsection{A generalized reverse VE SDE}

For any nonnegative function \(\lambda(\sigma)\), consider the reverse-time SDE
parameterized by \(dr=-d\sigma>0\):
\begin{equation}
    dx
    =
    (1+\lambda(\sigma))\sigma\nabla_x\log p_\sigma(x)\,dr
    +
    \sqrt{2\lambda(\sigma)\sigma}\,dW_r.
    \label{eq:app_general_reverse_sde}
\end{equation}
With the exact score, all choices of \(\lambda(\sigma)\) preserve the same
one-time marginals \(p_\sigma\). The case \(\lambda=0\) recovers the
probability-flow ODE, while positive \(\lambda\) adds Langevin correction. With
a learned score, we interpret
\begin{equation}
    dx
    =
    (1+\lambda(\sigma))\sigma s_\theta(x,\sigma)\,dr
    +
    \sqrt{2\lambda(\sigma)\sigma}\,dW_r
    \label{eq:app_learned_general_reverse_sde}
\end{equation}
as an SDE interpretation of the numerical sampler.

Discretizing \cref{eq:app_learned_general_reverse_sde} from \(\sigma_i\) to
\(\sigma_{i+1}\) gives
\begin{equation}
\begin{aligned}
    x_{i+1}
    =
    x_i
    &+
    \Delta_i\sigma_i s_\theta(x_i,\sigma_i) \\
    &+
    \lambda_i\Delta_i\sigma_i s_\theta(x_i,\sigma_i)
    +
    \sqrt{2\lambda_i\sigma_i\Delta_i}\,z_i,
    \qquad z_i\sim\mathcal{N}(0,I).
\end{aligned}
\label{eq:app_sde_euler}
\end{equation}
The first drift term is the probability-flow step. The remaining drift and
noise terms form a Langevin correction with effective SDE strength
\(\lambda_i\).

\subsection{EDM churn as a local Langevin correction}

As defined in \cref{eq:app_stochastic_churn}, the EDM churn step first increases the noise level to
\begin{equation}
    \hat\sigma_i=(1+\gamma_i)\sigma_i
    \label{eq:app_sigma_hat_2}
\end{equation}
and perturbs the state by the corresponding additional Gaussian noise:
\begin{equation}
    \hat x_i
    =
    x_i
    +
    \sqrt{\hat\sigma_i^2-\sigma_i^2}\,z_i,
    \qquad
    z_i\sim\mathcal{N}(0,I).
    \label{eq:app_churn_noise_2}
\end{equation}
Here we set \(S_{\mathrm{noise}}=1\) for the continuous-time limit. It then takes a deterministic probability-flow step from
\((\hat x_i,\hat\sigma_i)\) to \(\sigma_{i+1}\):
\begin{equation}
    x_{i+1}
    =
    \hat x_i
    +
    (\hat\sigma_i-\sigma_{i+1})
    \hat\sigma_i
    s_\theta(\hat x_i,\hat\sigma_i).
    \label{eq:app_churn_ode}
\end{equation}

Assume the small-step regime
\[
    \gamma_i=\mathcal{O}(N^{-1}),
    \qquad
    \Delta_i=\mathcal{O}(N^{-1}),
\]
with full-band unclipped churn (\(\gamma_i=S_{\mathrm{churn}}/N\)), and local smoothness of \(s_\theta\). Then
\begin{equation}
    \hat\sigma_i^2-\sigma_i^2
    =
    2\gamma_i\sigma_i^2+\gamma_i^2\sigma_i^2,
\end{equation}
so
\begin{equation}
    \hat x_i
    =
    x_i
    +
    \sqrt{2\gamma_i\sigma_i^2}\,z_i
    +
    o(N^{-1/2}).
\end{equation}
Moreover,
\begin{align}
    (\hat\sigma_i-\sigma_{i+1})\hat\sigma_i
    &=
    (\Delta_i+\gamma_i\sigma_i)\sigma_i(1+\gamma_i) \\
    &=
    \Delta_i\sigma_i
    +
    \gamma_i\sigma_i^2
    +
    \mathcal{O}(N^{-2}).
\end{align}
Replacing \(s_\theta(\hat x_i,\hat\sigma_i)\) by
\(s_\theta(x_i,\sigma_i)\) in the \(\mathcal{O}(N^{-1})\) deterministic
coefficient changes the update only by higher-order terms. Therefore
\begin{equation}
\begin{aligned}
    x_{i+1}
    =
    x_i
    &+
    \Delta_i\sigma_i s_\theta(x_i,\sigma_i) \\
    &+
    \gamma_i\sigma_i^2 s_\theta(x_i,\sigma_i)
    +
    \sqrt{2\gamma_i\sigma_i^2}\,z_i
    +
    \mathrm{h.o.t.}
\end{aligned}
\label{eq:app_churn_expansion}
\end{equation}
The first term is exactly the probability-flow Euler step in
\cref{eq:app_det_euler}. The remaining two terms are a local Langevin
correction at noise level \(\sigma_i\).

Comparing \cref{eq:app_churn_expansion} with the generalized SDE discretization
in \cref{eq:app_sde_euler}, both the additional drift and injected variance
match to leading order when
\begin{equation}
    \boxed{
    \lambda_i
    \approx
    \frac{\gamma_i\sigma_i}{\Delta_i}
    =
    \frac{\gamma_i\sigma_i}{\sigma_i-\sigma_{i+1}}
    }.
    \label{eq:app_lambda_master}
\end{equation}
This is the central identity: the raw churn parameter \(\gamma_i\) is not itself
the effective continuous-time stochasticity. The local schedule spacing
\(\Delta_i\) also determines where the sampler applies strong Langevin
correction.

\subsection{Finite-step caveat}

The previous matching is asymptotic. For finite \(\gamma_i\), the churn split
step is not exactly equal to one Euler step of
\cref{eq:app_learned_general_reverse_sde} with a single scalar
\(\lambda_i\), because drift matching and variance matching give slightly
different values.

Keeping the finite-\(\gamma_i\) deterministic coefficient while ignoring the
higher-order score-evaluation shift, the extra deterministic coefficient beyond
the probability-flow step is
\begin{align}
    &(\Delta_i+\gamma_i\sigma_i)\sigma_i(1+\gamma_i)
    -
    \Delta_i\sigma_i \\
    &\qquad =
    \gamma_i\sigma_i^2
    +
    \gamma_i\Delta_i\sigma_i
    +
    \gamma_i^2\sigma_i^2.
\end{align}
Matching this to
\(\lambda_i\Delta_i\sigma_i s_\theta\) gives
\begin{equation}
    \lambda_i^{\mathrm{drift}}
    =
    \frac{\gamma_i\sigma_i}{\Delta_i}
    +
    \gamma_i
    +
    \frac{\gamma_i^2\sigma_i}{\Delta_i}.
\end{equation}
On the other hand, exact variance matching gives
\begin{equation}
    \lambda_i^{\mathrm{noise}}
    =
    \left(\gamma_i+\frac{\gamma_i^2}{2}\right)
    \frac{\sigma_i}{\Delta_i}.
\end{equation}
These are not equal for finite \(\gamma_i\). However, in the small-step regime
\(\gamma_i=\mathcal{O}(N^{-1})\), \(\Delta_i=\mathcal{O}(N^{-1})\), both reduce
to
\[
    \lambda_i
    =
    \frac{\gamma_i\sigma_i}{\Delta_i}
    +
    \mathcal{O}(N^{-1}).
\]
Thus \cref{eq:app_lambda_master} is the leading-order continuous-time
interpretation of stochastic churn, not an exact finite-step identity.

\subsection{Entropy-rate grids make full-band churn information-adaptive}

For full-band unclipped EDM churn, the raw churn budget is distributed uniformly
across sampling intervals:
\begin{equation}
    \gamma_i=\frac{S_{\mathrm{churn}}}{N}.
    \label{eq:app_full_band_gamma}
\end{equation}
Substituting into \cref{eq:app_lambda_master} gives
\begin{equation}
    \lambda_i
    \approx
    \frac{S_{\mathrm{churn}}}{N}
    \frac{\sigma_i}{\sigma_i-\sigma_{i+1}}.
    \label{eq:app_lambda_schedule_dependent}
\end{equation}
Therefore the same raw churn budget induces different effective reverse SDEs
under different sampling schedules.

Let \(u=\log\sigma\), and let \(\pi_\alpha(u)\) denote the normalized schedule density 
used to construct the entropy-rate sampling grid, as defined in
\cref{eq:app_normalized_entropy_rate_density}. The grid places sampling points
approximately uniformly in the CDF of \(\pi_\alpha\). Thus consecutive
log-noise points satisfy
\[
    F_\alpha(u_i)-F_\alpha(u_{i+1})
    \approx
    \frac{1}{N}.
\]
For interior points where \(\pi_\alpha\) is smooth and positive,
\begin{equation}
    u_i-u_{i+1}
    \approx
    \frac{1}{N\pi_\alpha(u_i)}.
\end{equation}
Since
\[
    \sigma_i-\sigma_{i+1}
    \approx
    \sigma_i(u_i-u_{i+1}),
\]
we obtain
\begin{equation}
    \sigma_i-\sigma_{i+1}
    \approx
    \frac{\sigma_i}{N\pi_\alpha(\log\sigma_i)}.
\end{equation}
Substituting into \cref{eq:app_lambda_schedule_dependent} yields
\begin{equation}
    \boxed{
    \lambda_{\mathrm{ent}}(\sigma)
    \approx
    S_{\mathrm{churn}}\,
    \pi_\alpha(\log\sigma)
    }.
    \label{eq:app_entropic_lambda}
\end{equation}

Thus, under the entropy-rate grid, constant full-band raw churn induces an
information-adaptive reverse SDE. In the configuration used for the main
results, \(\alpha=1/2\), so the effective Langevin strength is proportional to
the normalized schedule density induced by the square root of the entropy-rate
proxy over log-noise. This makes the
stochastic correction largest in the same noise region where denoising
information is concentrated.

For comparison, a Karras grid induces a smooth analytic stochasticity profile
determined by the grid geometry rather than by the measured bitstream
entropy-rate. This explains why the same raw stochastic sampler can behave very
differently under Karras and entropy-rate schedules, as observed in
\cref{app:schedule_ablation}.

\subsection{Interpretation for bitstreams}

This effect is especially important for continuous diffusion over discrete
bitstreams. The noisy marginal \(p_\sigma\) is a mixture of Gaussians centered
at hypercube vertices. At high noise, the mixture components overlap strongly
and bit identities are ambiguous. At low noise, the components are separated
and most bits are already determined. Between these regimes, there is a
comparatively narrow transition region where many posterior bit probabilities
change rapidly.

Finite-step probability-flow integration can be fragile in this transition
region: discretization or score error may push coordinates toward incorrect
hypercube vertices, after which the matched-filter term and contextual denoiser
can reinforce premature commitments. The stochastic correction mitigates this
failure mode by locally re-randomizing ambiguous coordinates and allowing the
denoiser to refreeze them under updated global context.

The entropy-rate schedule aligns two effects. It places more deterministic
solver points where denoising information is concentrated, and through
\cref{eq:app_entropic_lambda} it also increases the effective Langevin
correction in those same regions. Hence full-band churn on the entropy-rate
grid acts as an information-adaptive predictor-corrector sampler: both the
deterministic discretization budget and the stochastic correction budget are
concentrated where the bitstream is actively resolving information.
\section{Training and sampling ablations}
\label{subsec:training_ablations}

\Cref{tab:training_ablations} isolates the main training-side choices on LM1B.
Self-conditioning is enabled in all variants \citep{chen2022analogbits,chen2026langflow}.
All variants use the same 256-NFE deterministic sampler and the same 256-NFE stochastic churn search grid. This controls the available search range, but does not establish that every reported row uses the same selected churn setting.

\begin{table}[ht]
\centering
\small
\setlength{\tabcolsep}{2.8pt}
\renewcommand{\arraystretch}{0.98}
\begin{tabular}{@{}lp{0.34\linewidth}cccc@{}}
\toprule
\multirow{2}{*}{\textbf{Variant}} &
\multirow{2}{*}{\textbf{Schedule + Objective + Filter}} &
\multicolumn{2}{c}{\textbf{Deterministic}} &
\multicolumn{2}{c}{\textbf{Stochastic}} \\
\cmidrule(lr){3-4} \cmidrule(l){5-6}
& & \textbf{GenPPL} $\downarrow$ & \textbf{Entropy} &
\textbf{GenPPL} $\downarrow$ & \textbf{Entropy} \\
\midrule
\multicolumn{6}{@{}l}{\textit{Reference}} \\
Real data & -- & 53.06 & 4.34 & 53.06 & 4.34 \\
\midrule
\multicolumn{6}{@{}l}{\textit{Ablations}} \\
Config A (Base) & entropy-rate + SM + matched filter
& 85.37   & 4.30 & 59.93  & 4.31 \\
Config B & EDM schedule + SM + matched filter
& 81.95   & 4.29 & 54.04  & 4.29 \\
Config C & entropy-rate + CE + matched filter
& 91.93   & 4.29 & 54.68  & 4.27 \\
Config D & entropy-rate + SM + no matched-filter prior
& 105.96  & 4.34 & 66.18  & 4.33 \\
\bottomrule
\end{tabular}
\vspace{1mm}
\caption{\textbf{Training ablations on LM1B.}
All variants use 256 NFEs. The stochastic evaluation uses ten seeds and a common target-churn sweep \(\gamma\in[0.11,0.20]\). We report the comparison point estimates without error bars. Config A uses the default design. These ablation measurements are reported separately from the main checkpoint evaluation and should not be treated as identical numerical replicates.}
\label{tab:training_ablations}
\end{table}

These LM1B ablations use a single tokenizer and encoding; they do not establish an optimal schedule or objective across representations. \Cref{subsec:training_ablations_main} discusses the component comparisons, and \Cref{app:discrete_bitstream_baseline} gives the matched discrete-bitstream control.

\begin{figure}[t]
    \centering
    \begin{subfigure}[t]{0.43\linewidth}
        \centering
        \vspace{0pt}
        \includegraphics[width=\linewidth]{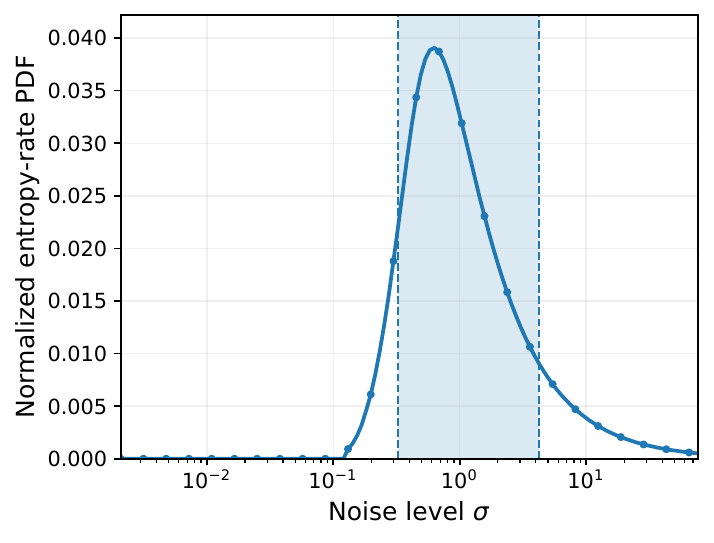}
        \caption{Entropy-rate profile.}
        \label{fig:entropy_rate_profile}
    \end{subfigure}
    \hfill
    \begin{subfigure}[t]{0.54\linewidth}
        \centering
        \vspace{0pt}
        \includegraphics[height=4.6cm]{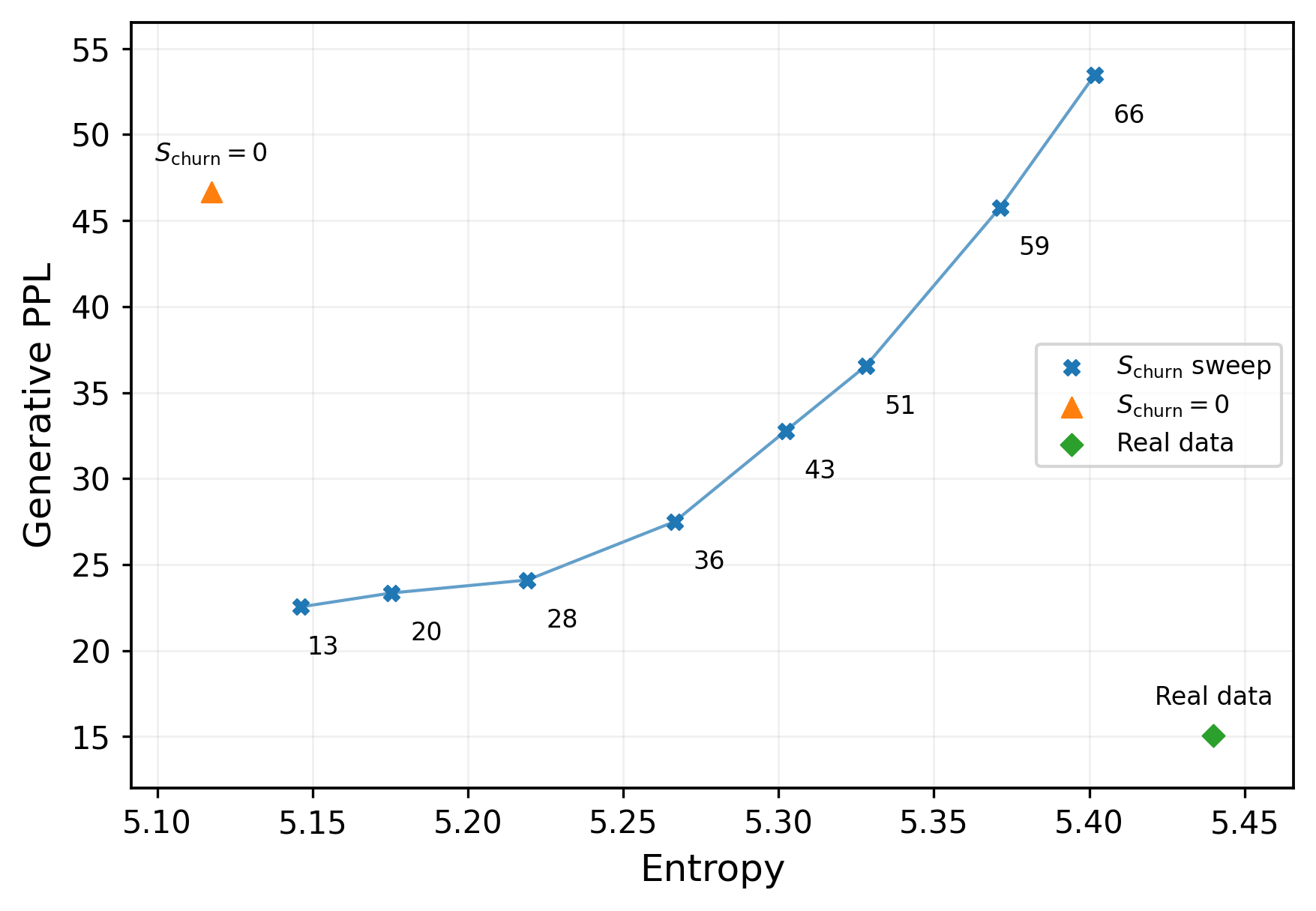}
        \caption{Churn sweep at fixed 256 NFEs.}
        \label{fig:owt_churn_sweep}
    \end{subfigure}
    \caption{\textbf{Stochastic churn controls the OWT quality--diversity frontier.}
    \textbf{Left:} Normalized entropy-rate profile over noise level \(\sigma\);
    the shaded region marks the broad entropy-active region used for this
    diagnostic.
    \textbf{Right:} At fixed NFE \(=256\) and \(\eta=0\), moderate full-band
    churn improves both GenPPL and entropy relative to deterministic sampling,
    while excessive churn increases entropy at the cost of GenPPL.}
    \label{fig:owt_churn_ablation}
\end{figure}

\paragraph{Churn controls the quality--diversity frontier.}
\Cref{fig:owt_churn_ablation} isolates stochastic churn on OWT at fixed
NFE \(=256\) and \(\eta=0\).
The deterministic sampler is over-contractive: it undershoots real-data entropy
and gives worse GenPPL than the best stochastic settings.
Moderate full-band churn improves both GenPPL and entropy, moving samples into a
healthier quality--diversity regime.
Increasing churn further continues to raise entropy, but eventually worsens
GenPPL.
Thus \(S_{\mathrm{churn}}\) is best understood as a test-time frontier knob:
moderate stochasticity corrects premature contraction, while excessive
stochasticity injects too much uncertainty.

\paragraph{Sampler-design ablations.}
Additional ablations in \Cref{app:sampling_ablations} support the default
sampler choices.
The entropy-rate grid improves deterministic sampling relative to the Karras
grid and makes stochastic churn substantially more effective
(\Cref{app:fig_schedule_ablation}).
This indicates that stochasticity is not a generic benefit by itself: it is most
useful when the sampling grid aligns both solver resolution and stochastic
correction with the information-active part of the trajectory.
The continuous-time interpretation in \Cref{app:stochastic_sde_analysis}
explains this behavior, since the effective Langevin strength depends on local
grid spacing as well as the raw churn parameter.
The stochasticity-window ablation further shows that the correction should be
broad rather than hand-localized (\Cref{app:fig_churn_window_ablation}).
Narrow entropy-CDF windows are sensitive to their location, whereas broad
windows are consistently stronger; the full entropy-supported range
\(q\in[0,1]\) gives the best GenPPL in this sweep, with \(q\in[0.1,0.9]\) close
behind.
Thus the useful stochastic correction appears to act over a broad
information-active region, rather than through a single narrowly localized
entropy band.

\section{Additional sampling ablations}
\label{app:sampling_ablations}

We provide additional sampling diagnostics supporting the sampler choices used
in the main paper. The main comparison already isolates the deterministic
probability-flow sampler from the stochastic sampler. Here we focus on the
internal stochastic-sampler choices: the sigma grid, the churn window, the churn
budget, and the asymmetric time-lag parameter.

\subsection{Entropic versus Karras sampling grids}
\label{app:schedule_ablation}

\begin{figure}[h!]
    \centering
    \includegraphics[width=\linewidth]{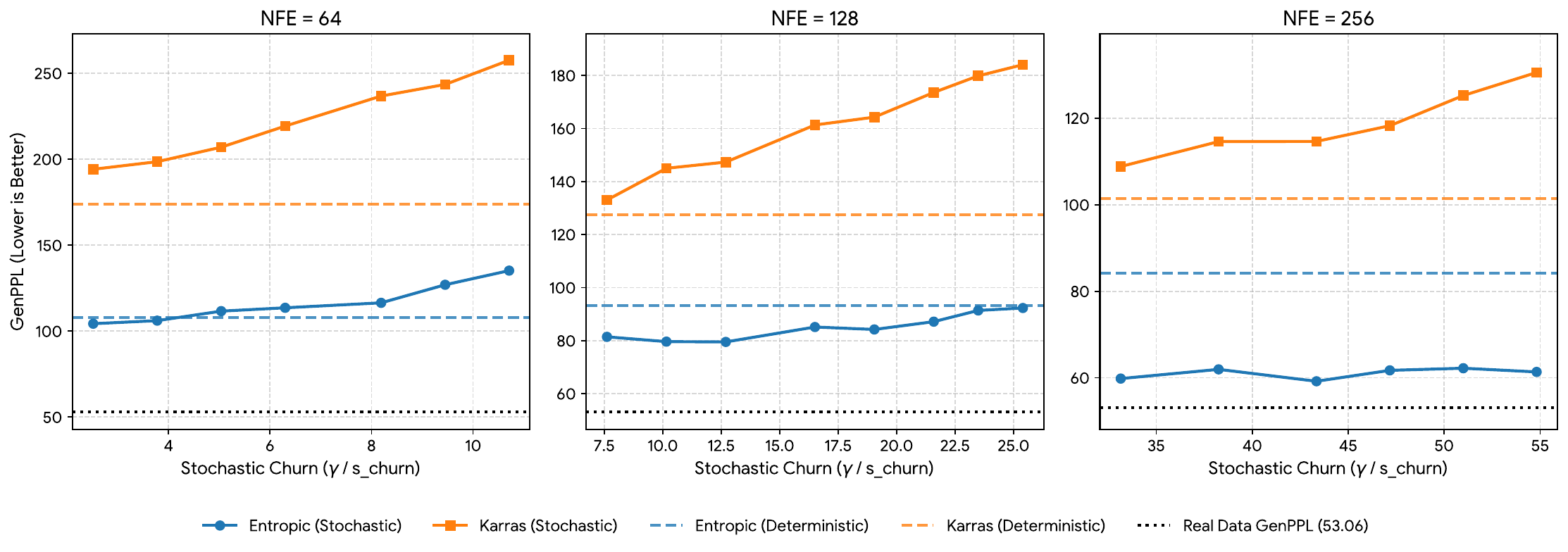}
    \caption{\textbf{LM1B schedule ablation.}
    We compare the entropy-rate grid and the Karras grid at 64, 128, and
    256 NFEs. Dashed horizontal lines show deterministic sampling; solid curves
    show stochastic churn sweeps. The entropy-rate grid improves deterministic
    sampling and, more importantly, enables stochastic churn to reach a much
    better GenPPL regime than the Karras grid across all NFE budgets. The
    real-data GenPPL reference is 53.06.}
    \label{app:fig_schedule_ablation}
\end{figure}

\Cref{app:fig_schedule_ablation} compares the entropy-rate sampling grid against
the standard Karras grid on LM1B at 64, 128, and 256 NFEs. For each NFE, we show
both deterministic sampling and a sweep over stochastic churn budgets, using the
same broad entropy-CDF churn window \(q\in[0.1,0.9]\). This experiment isolates
the effect of the sampling grid while keeping the trained checkpoint fixed.

There are two conclusions. First, the entropy-rate grid improves deterministic
sampling itself. The dashed entropy-rate curves are consistently below the
corresponding Karras dashed curves, especially at low NFE. This indicates that
the entropy-rate grid is a better discretization of the probability-flow
trajectory for bitstreams: it allocates more solver resolution to the noise
region where denoising error and bit uncertainty are concentrated.

Second, the grid choice becomes even more important under stochastic churn. On
the entropy-rate grid, moderate churn substantially improves GenPPL over the
deterministic sampler, particularly at 128 and 256 NFEs. At 256 NFEs, the best
entropy-rate stochastic configurations reach the GenPPL-\(60\) regime, while
the Karras stochastic configurations remain above the Karras deterministic
baseline. A similar separation appears at 128 NFEs, where the entropy-rate
stochastic sampler reaches the GenPPL-\(80\) regime whereas Karras remains far
higher. Thus stochastic churn is not automatically beneficial; it is beneficial
when the grid places the stochastic correction in the right part of the
trajectory.

This supports the sampler used in the main experiments: entropy-rate sampling
is useful as a deterministic solver grid, and it also makes stochastic churn
more effective. The reverse-SDE analysis in
\Cref{app:stochastic_sde_analysis} gives a continuous-time explanation: the
effective Langevin strength scales with the inverse local sigma spacing, so the
entropy-rate grid concentrates the stochastic correction near the
information-active noise region.

\subsection{Stochasticity-window ablation}
\label{app:window_ablation}

\begin{figure}[h!]
    \centering
    \includegraphics[width=\linewidth]{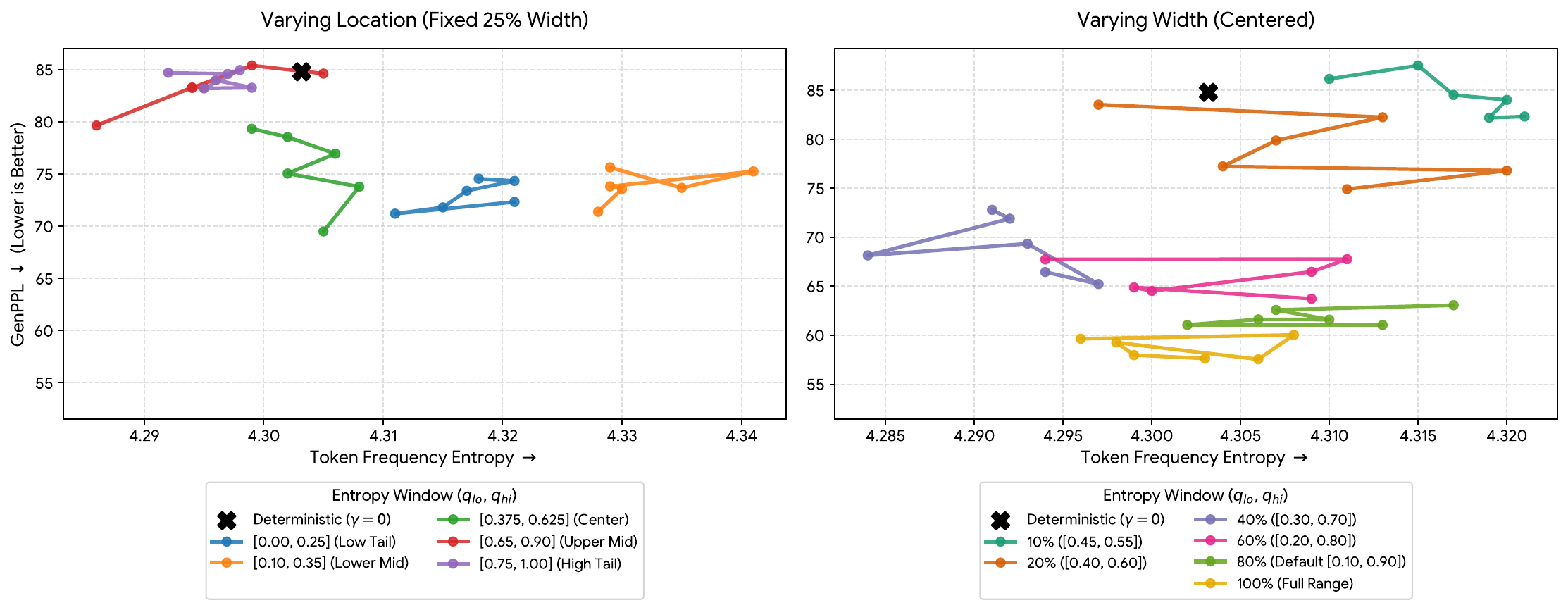}
    \caption{\textbf{LM1B stochasticity-window ablation at 256 NFEs.}
    Left: location sweep with a fixed 25\% entropy-CDF width. Right: centered
    width sweep. Narrow windows are sensitive to their location, while broad
    windows are consistently stronger. The full entropy-supported range
    \(q\in[0,1]\) gives the best GenPPL in this sweep. The real-data reference
    (GenPPL 53.06, entropy 4.34) is omitted to focus on the model trajectories.}
    \label{app:fig_churn_window_ablation}
\end{figure}

\Cref{app:fig_churn_window_ablation} studies where stochastic churn should be
applied. We fix LM1B, NFE \(=256\), the entropy-rate grid, and \(\eta=0\), and
sweep both the churn budget and the entropy-CDF window
\[
    \mathcal{W}
    =
    [Q(q_{\mathrm{lo}}),Q(q_{\mathrm{hi}})].
\]

The left panel varies the location of a fixed-width 25\% entropy-CDF window.
No narrow location is uniformly best. Low-tail, centered, upper-mid, and
high-tail windows all improve some configurations, but each leaves a clear gap
to the best broad-window results. This suggests that the useful stochastic
correction is not confined to one sharply localized part of the trajectory.

The right panel varies the window width while keeping it centered. Performance
improves as the window becomes broader. The full entropy-supported range
\(q\in[0,1]\) gives the best GenPPL in this sweep, with the broad
\(q\in[0.1,0.9]\) setting close behind. Narrow centered windows are
substantially weaker. We therefore use full-band churn in the main comparison.

This result is also consistent with the continuous-time view in
\Cref{app:stochastic_sde_analysis}. Full-band raw churn does not imply uniform
effective stochasticity: on the entropy-rate grid, the local grid spacing
already concentrates the effective Langevin correction near the
information-active region. Empirically, applying churn broadly is more robust
than trying to hand-target a narrow window.

\subsection{Additional OWT churn sweeps}
\label{app:owt_churn_sweeps}

\begin{figure}[h!]
    \centering
    \includegraphics[width=\linewidth]{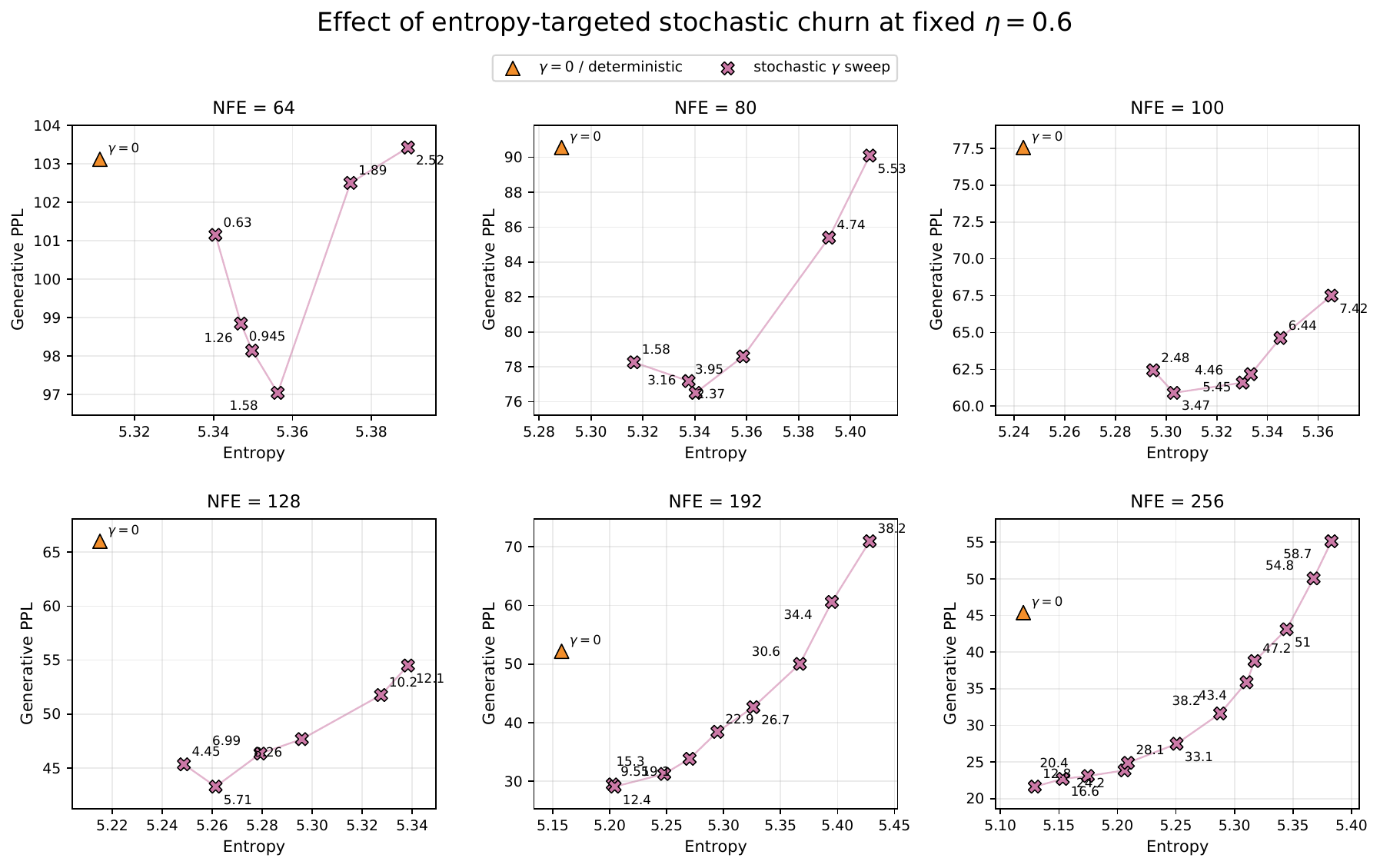}
    \caption{\textbf{OWT churn sweep at fixed \(\eta=0.6\).}
    The qualitative behavior matches the \(\eta=0\) sweep: moderate churn gives
    the best GenPPL--entropy trade-off, while excessive churn increases entropy
    at the cost of GenPPL.}
    \label{app:fig_owt_churn_eta06}
\end{figure}

The main text shows the OWT churn sweep at \(\eta=0\). In
\Cref{app:fig_owt_churn_eta06}, we repeat the sweep at \(\eta=0.6\). The same
qualitative behavior holds: small-to-moderate churn gives the best
GenPPL--entropy trade-off, while excessive churn raises token-frequency entropy
at the cost of worse GenPPL. This confirms that the churn budget is the dominant
sampler-level control knob for moving along the quality--diversity frontier.

The precise best operating point changes with \(\eta\), but the overall frontier
shape is stable: stochasticity first corrects the over-contraction of the
deterministic sampler, and then eventually becomes too strong. This is why the
main results tune \(S_{\mathrm{churn}}\) but keep \(\eta=0\) for the default
256-NFE stochastic comparison.

\subsection{Asymmetric time-lag ablation}
\label{app:ati_ablation}

\begin{figure}[h!]
    \centering
    \includegraphics[width=\linewidth]{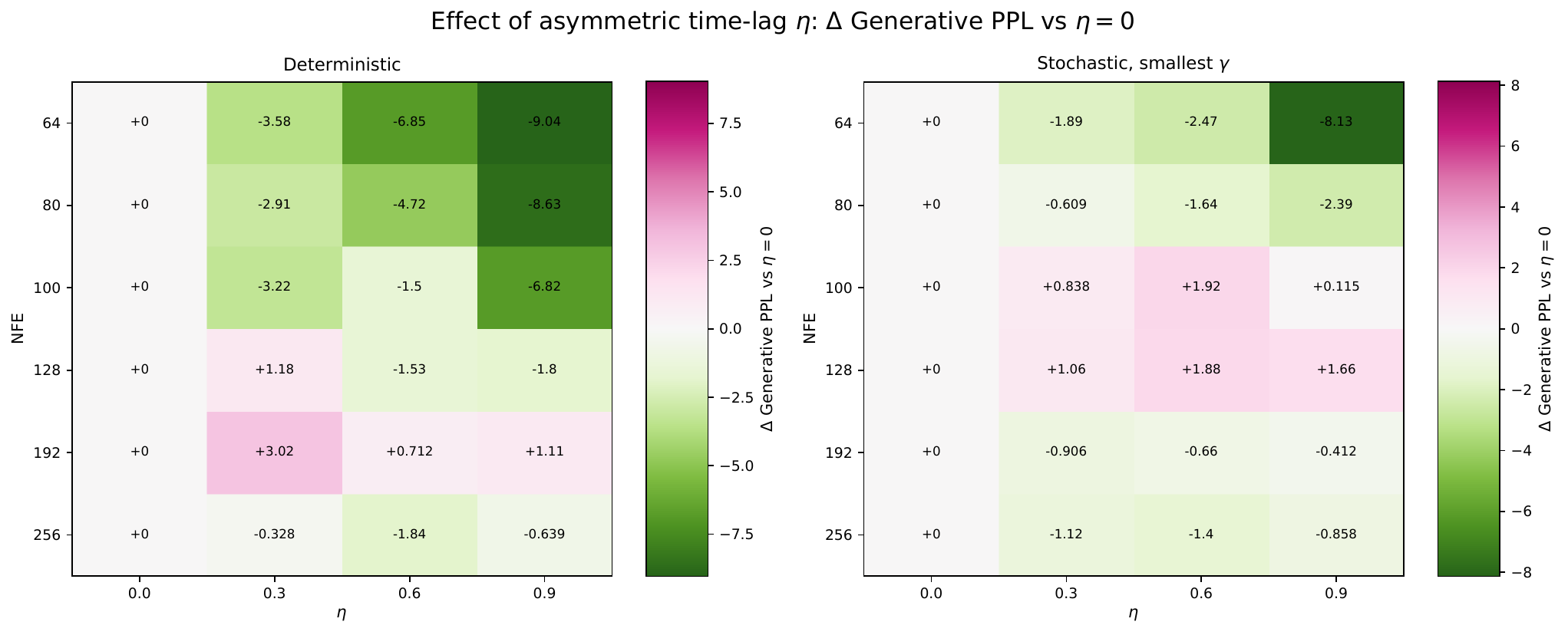}
    \vspace{0.5em}
    \includegraphics[width=\linewidth]{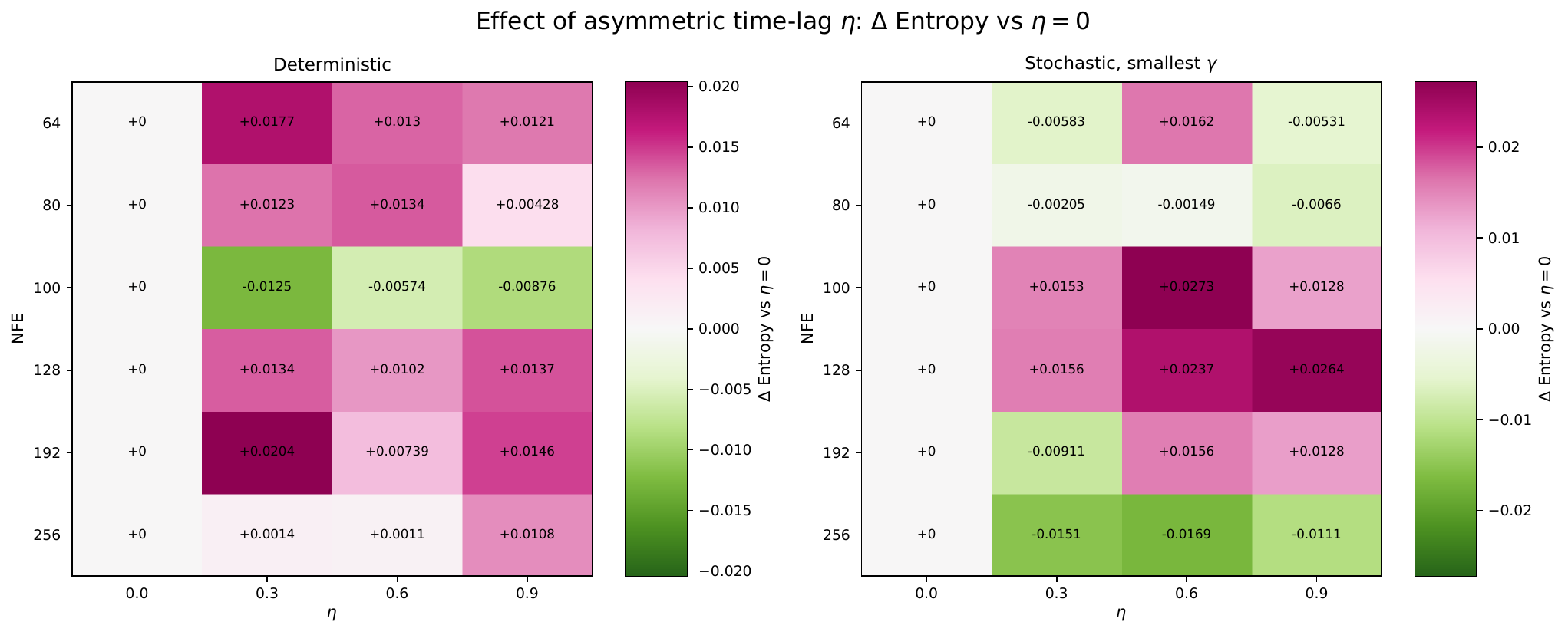}
    \caption{\textbf{Effect of asymmetric time-lag \(\eta\) on OWT.}
    We show changes relative to \(\eta=0\) for deterministic sampling and for
    the smallest nonzero stochastic churn configuration. Top: change in GenPPL.
    Bottom: change in token-frequency entropy. Positive \(\eta\) can help at
    low NFE in deterministic sampling, but its effect is weaker and less
    consistent once stochastic churn is enabled.}
    \label{app:fig_owt_eta}
\end{figure}

We also ablate the asymmetric time-lag parameter \(\eta\), which shifts the
denoiser evaluation label toward a noisier adjacent level in log-\(\sigma\)
space. \Cref{app:fig_owt_eta} reports changes in GenPPL and entropy relative
to \(\eta=0\), both for deterministic sampling and for the smallest nonzero
stochastic churn configuration.

Positive \(\eta\) can improve GenPPL in some low-NFE deterministic regimes, but
the effect is not uniformly Pareto-improving. Once stochastic churn is enabled,
the effect of \(\eta\) becomes weaker and less consistent. Entropy changes are
also mixed across NFE. We therefore treat \(\eta\) as a secondary tuning
parameter: it can be useful for low-NFE sweeps, but we use \(\eta=0\) for the
main high-NFE stochastic comparisons.

\section{Computational and memory efficiency}
\label{app:efficiency}

This appendix gives the full matched-trunk training and generation benchmarks in \Cref{subsec:efficiency_main}, together with a benchmark of the output head and loss alone. The end-to-end and boundary-only benchmarks use distinct timing protocols, specified below.

\subsection{Profiling protocol}
\label{app:efficiency:protocol}

All measurements use a single NVIDIA GH200 GPU and synthetic batches. They exclude data loading, tokenization, disk I/O, host-device transfer, decoding, external scoring, and cache writes. Both models use BF16 autocast and CUDA peak allocated memory, with \texttt{torch.compile} disabled. The token-space baseline uses a continuous one-hot state, a dense vocabulary output head, and cross-entropy; CoBit uses patched analog bits, the \texttt{optimal\_skip\_mlp} head, and a bitwise loss. Both have a 12-layer SDT trunk of width 768 and the same number of heads. The LM1B setting uses \(T=128,V=30{,}522,m=15\); the OWT setting uses \(T=1024,V=65{,}536,m=16\).

Training uses AdamW, 10 warmup steps, and 50 timed forward--backward--optimizer steps. Self-conditioning is disabled, so each step has one denoiser call. Generation uses no gradients, 50 untimed denoiser calls for warmup, and five timed 128-NFE trajectories. Generation throughput counts completed output tokens over the full trajectory, not intermediate denoiser outputs. Here a token means one trunk position: a tokenizer token for LM1B or a code token for the OWT configuration. These synthetic measurements do not establish a speedup per decoded GPT-2 token or at matched sample quality.

\subsection{End-to-end training sweeps}
\label{app:efficiency:training}
\Cref{tab:efficiency_train_lm1b,tab:efficiency_train_owt} report peak memory and throughput across batch sizes for the two sequence lengths.

\begin{table}[ht]
\centering\small
\begin{tabular}{r r r c r r c}
\toprule
\textbf{Batch} & \multicolumn{2}{c}{\textbf{Peak VRAM (GiB)}} & \textbf{Reduction} & \multicolumn{2}{c}{\textbf{Tokens/s}} & \textbf{Speedup}\\
\cmidrule(lr){2-3}\cmidrule(lr){5-6}
& Token & CoBit & & Token & CoBit & \\
\midrule
16 & 3.11 & 1.90 & $1.63\times$ & 45{,}787 & 44{,}031 & $0.96\times$\\
32 & 4.70 & 3.00 & $1.57\times$ & 90{,}323 & 88{,}021 & $0.97\times$\\
64 & 8.11 & 5.20 & $1.56\times$ & 156{,}924 & 174{,}845 & $1.11\times$\\
128 & 15.29 & 9.59 & $1.59\times$ & 158{,}496 & 260{,}400 & $1.64\times$\\
256 & 29.87 & 18.38 & $1.62\times$ & 148{,}460 & 278{,}160 & $1.87\times$\\
512 & 59.04 & 35.97 & $1.64\times$ & 126{,}026 & 290{,}758 & $2.31\times$\\
\bottomrule
\end{tabular}
\caption{LM1B training-step profiling. The token baseline runs out of memory at the next profiled batch size. Small batches have similar throughput despite the memory advantage; the speedup increases with utilization. Ratios are reported from the profiling runs and may differ slightly from ratios of rounded entries.}
\label{tab:efficiency_train_lm1b}
\end{table}

\begin{table}[ht]
\centering\small
\begin{tabular}{r r r c r r c}
\toprule
\textbf{Batch} & \multicolumn{2}{c}{\textbf{Peak VRAM (GiB)}} & \textbf{Reduction} & \multicolumn{2}{c}{\textbf{Code tokens/s}} & \textbf{Speedup}\\
\cmidrule(lr){2-3}\cmidrule(lr){5-6}
& Token & CoBit & & Token & CoBit & \\
\midrule
16 & 22.65 & 9.64 & $2.35\times$ & 112{,}542 & 230{,}225 & $2.05\times$\\
32 & 44.70 & 18.46 & $2.42\times$ & 98{,}642 & 245{,}664 & $2.49\times$\\
64 & 88.79 & 36.12 & $2.46\times$ & 77{,}710 & 255{,}642 & $3.29\times$\\
\bottomrule
\end{tabular}
\caption{OWT training-step profiling at matched code-sequence length. The token baseline runs out of memory at the next profiled batch size.}
\label{tab:efficiency_train_owt}
\end{table}

\FloatBarrier
\subsection{End-to-end generation sweeps}
\label{app:efficiency:generation}

Generation is measured at 128 NFEs for both models, including the full iterative denoising trajectory. These measurements isolate systems cost at a fixed number of evaluations; they are not latency comparisons against autoregressive generation.

\begin{table}[ht]
\centering\small
\begin{tabular}{r r r c r r c}
\toprule
\textbf{Batch} & \multicolumn{2}{c}{\textbf{Peak VRAM (GiB)}} & \textbf{Reduction} & \multicolumn{2}{c}{\textbf{Generated tokens/s}} & \textbf{Speedup}\\
\cmidrule(lr){2-3}\cmidrule(lr){5-6}
& Token & CoBit & & Token & CoBit & \\
\midrule
16 & 1.21 & 0.39 & $3.12\times$ & 1{,}075 & 1{,}146 & $1.07\times$\\
32 & 2.02 & 0.46 & $4.36\times$ & 1{,}928 & 2{,}277 & $1.18\times$\\
64 & 3.65 & 0.62 & $5.91\times$ & 3{,}009 & 4{,}488 & $1.49\times$\\
128 & 6.91 & 0.93 & $7.46\times$ & 3{,}067 & 5{,}794 & $1.89\times$\\
256 & 13.43 & 1.54 & $8.70\times$ & 3{,}149 & 6{,}111 & $1.94\times$\\
512 & 26.47 & 2.78 & $9.53\times$ & 3{,}198 & 6{,}346 & $1.98\times$\\
1024 & 52.55 & 5.25 & $10.02\times$ & 3{,}210 & 6{,}413 & $2.00\times$\\
\bottomrule
\end{tabular}
\caption{LM1B generation profiling at 128 NFEs. Throughput counts completed generated tokens.}
\label{tab:efficiency_generate_lm1b}
\end{table}

\begin{table}[ht]
\centering\small
\begin{tabular}{r r r c r r c}
\toprule
\textbf{Batch} & \multicolumn{2}{c}{\textbf{Peak VRAM (GiB)}} & \textbf{Reduction} & \multicolumn{2}{c}{\textbf{Generated code tokens/s}} & \textbf{Speedup}\\
\cmidrule(lr){2-3}\cmidrule(lr){5-6}
& Token & CoBit & & Token & CoBit & \\
\midrule
16 & 14.49 & 0.97 & $15.00\times$ & 2{,}186 & 5{,}378 & $2.46\times$\\
32 & 28.49 & 1.62 & $17.56\times$ & 2{,}216 & 5{,}677 & $2.56\times$\\
64 & 56.49 & 2.94 & $19.25\times$ & 2{,}219 & 5{,}889 & $2.65\times$\\
\bottomrule
\end{tabular}
\caption{OWT generation profiling at 128 NFEs. The token baseline runs out of memory at the next profiled batch size.}
\label{tab:efficiency_generate_owt}
\end{table}

\FloatBarrier
\subsection{Logit and parameter scaling}
\label{app:efficiency:scaling}

The main architectural difference is the output/loss tensor size.
A vocabulary-boundary diffusion model forms \(BTV\) logits, whereas \methodname{} forms \(BTm\) bit logits with \(m=\lceil\log_2 V\rceil\).
Thus, the raw logit reduction factor is \(V/m\).
This reduction applies at every denoising evaluation, so it affects both training and iterative sampling.

\begin{table}[t]
\centering
\small
\setlength{\tabcolsep}{4pt}
\begin{tabular}{lrrrrcc}
\toprule
\textbf{Setting} &
\textbf{\(B\)} &
\textbf{\(T\)} &
\textbf{\(V\)} &
\textbf{\(m\)} &
\textbf{Token logits} &
\textbf{Bit logits} \\
\midrule
LM1B
& 512 & 128 & 30{,}522 & 15
& \(2.00{\times}10^{9}\)
& \(9.83{\times}10^{5}\)
\\
Long context
& 16 & 8192 & 65{,}536 & 16
& \(8.59{\times}10^{9}\)
& \(2.10{\times}10^{6}\)
\\
Large vocabulary
& 16 & 4096 & 128{,}000 & 17
& \(8.39{\times}10^{9}\)
& \(1.11{\times}10^{6}\)
\\
Large model/vocabulary
& 8 & 4096 & 128{,}000 & 17
& \(4.19{\times}10^{9}\)
& \(5.57{\times}10^{5}\)
\\
\bottomrule
\end{tabular}
\caption{\textbf{Vocabulary-boundary tensor sizes.}
Token-based diffusion models form vocabulary-wide logits of size \(BTV\), whereas \methodname{} forms bit logits of size \(BTm\), with \(m=\lceil\log_2 V\rceil\).
The reduction factor is \(V/m\): \(2035\times\) for LM1B, \(4096\times\) for \(V=65{,}536\), and \(7529\times\) for \(V=128{,}000\).
In BF16, the LM1B row corresponds to 3.73~GiB of token logits versus 1.88~MiB of bit logits before accounting for gradients, optimizer state, and loss temporaries.}
\label{tab:logit_scaling}
\end{table}

A dense token classifier also allocates \(dV\) output parameters.
This parameter cost is not always the dominant bottleneck, but it becomes substantial for large vocabularies and wide models.

\begin{table}[t]
\centering
\small
\begin{tabular}{lccc}
\toprule
\textbf{Setting} &
\textbf{Width \(d\)} &
\textbf{Vocabulary \(V\)} &
\textbf{Token output parameters \(dV\)} \\
\midrule
LM1B-scale
&
768
&
30{,}522
&
23.4M
\\
Medium vocabulary
&
1024
&
65{,}536
&
67.1M
\\
Large vocabulary
&
2048
&
128{,}000
&
262.1M
\\
Large model/vocabulary
&
4096
&
128{,}000
&
524.3M
\\
\bottomrule
\end{tabular}
\caption{\textbf{Vocabulary-head parameter scaling.}
A dense token output head scales linearly as \(dV\).
\methodname{} replaces this vocabulary classifier with a compact bitwise head whose output dimension grows only as \(m=\lceil\log_2 V\rceil\).}
\label{tab:vocab_head_params}
\end{table}

\subsection{Boundary-only profiling}
\label{app:efficiency:boundary_profile}

The end-to-end benchmarks above include the Transformer trunk, optimizer, representation-specific adapters, and the output/loss boundary.
To isolate the vocabulary-boundary mechanism itself, we also profile only the output head and loss.
We generate synthetic trunk activations \(h\in\mathbb{R}^{B\times T\times d}\) and compare two modules.

The token boundary applies a dense \(d\to V\) vocabulary classifier and cross-entropy over \(B\times T\times V\) logits.
The bitstream boundary applies the compact per-bit expansion used by \methodname{} and computes a bitwise loss over \(B\times Tm\) logits.
Boundary-only measurements use a shorter timing protocol than the end-to-end runs. All boundary-only runs use batch size \(B=16\), BF16 autocast, AdamW, three warmup steps, and ten timed forward--backward--optimizer steps.

\begin{table}[t]
\centering
\small
\begin{tabular}{lccccc}
\toprule
\textbf{Boundary} &
\textbf{\(T\)} &
\textbf{\(V\)} &
\textbf{\(d\)} &
\textbf{Peak VRAM (GiB)} &
\textbf{Step time (ms)} \\
\midrule
Token boundary
& 128 & 30{,}522 & 768 & 0.92 & 3.08 \\
Bitstream boundary
& 128 & 30{,}522 & 768 & 0.17 & 2.73 \\
\midrule
Token boundary
& 1024 & 65{,}536 & 1024 & 7.89 & 53.69 \\
Bitstream boundary
& 1024 & 65{,}536 & 1024 & 0.86 & 3.33 \\
\midrule
Token boundary
& 4096 & 128{,}000 & 2048 & 64.11 & 1076.24 \\
Bitstream boundary
& 4096 & 128{,}000 & 2048 & 3.54 & 13.26 \\
\bottomrule
\end{tabular}
\caption{\textbf{Boundary-only head and loss profiling.}
This benchmark removes the Transformer trunk and isolates the output/loss boundary.
The token boundary uses a dense \(d\to V\) vocabulary head and cross-entropy.
The bitstream boundary uses a compact per-bit head and a bitwise loss over \(m=\lceil\log_2 V\rceil\) bit logits per semantic token.
The batch size is \(B=16\).}
\label{tab:boundary_profiling}
\end{table}

\Cref{tab:boundary_profiling} shows that the isolated boundary advantage grows rapidly with context length and vocabulary size.
At LM1B scale, the bitstream boundary reduces peak memory from \(0.92\) GiB to \(0.17\) GiB.
At \(T=1024,V=65{,}536,d=1024\), it reduces memory from \(7.89\) GiB to \(0.86\) GiB and step time from \(53.69\) ms to \(3.33\) ms.
At \(T=4096,V=128{,}000,d=2048\), it reduces memory from \(64.11\) GiB to \(3.54\) GiB and step time from \(1076.24\) ms to \(13.26\) ms.
This corresponds to an \(18.1\times\) memory reduction and an \(81.2\times\) boundary-step speedup in the largest profiled setting.

The boundary-only speedups are larger than the full-step speedups because the Transformer trunk has been removed.
This is intentional: the benchmark isolates the mechanism that the bitstream representation changes.
In a complete model, the realized gain depends on the fraction of time and memory spent in the Transformer trunk versus the output/loss boundary.
The end-to-end benchmarks report the practical effect at the profiled LM1B and OWT settings; the boundary-only benchmark isolates the contribution of the output interface.

\subsection{Limitations}
\label{app:efficiency:limitations}

These profiling results measure systems cost, not sample quality.
They use synthetic data and random targets in order to isolate architectural scaling.
The token baseline is the most direct continuous token-boundary comparison; embedding-space or discrete-token diffusion variants may differ in input-state cost, loss implementation, and kernel efficiency.
Dense vocabulary-wide logits have \(\mathcal{O}(BTV)\) size, although implementations that avoid materializing them can have different memory costs.
\methodname{} replaces this dense boundary with bit logits while preserving the Transformer sequence length \(T\).

\section{Verifiable tasks and score-temperature experiments}
\label{app:reasoning_temperature}
This appendix records the protocols behind \Cref{sec:reasoning} and separates the earlier task evaluations from the subsequent temperature experiments. Headline configurations were selected by a sweep on a TinyGSM validation set and then evaluated on GSM8K. We also report GSM8K sensitivity analyses and numerical diagnostics. The GSM8K results use one generation seed; uncertainty estimates describe variation across test problems, conditional on the checkpoint and configuration.

\subsection{Sudoku protocol and churn ablation}
\label{app:sudoku_protocol}
The task follows \citet{deschenaux2026continuous}. Each example is serialized as \texttt{[BOS] puzzle(89) [BOS] solution(89)} in 180 positions using 12 symbols (digits 1--9, blank, row separator, and BOS), each represented by four bits. The first 91 positions are held clean throughout denoising and excluded from the loss; the remaining 89 positions contain the solution. Exact match requires the complete solution suffix to equal the unique reference solution. Easy, medium, and hard puzzles contain 40, 35, and 30 clues, respectively. The deduplicated, unique-solution data use a 48K/2K split with split seed 42.

The task-specific transformer has eight blocks, width 512, and eight heads, and is trained for 20K steps at global batch size 256 with AdamW and learning rate $3\times10^{-4}$. This is a smaller backbone than the 130M language model. Evaluation uses the raw training weights, the entropic Euler/DDIM sampler, and 180 denoiser evaluations, without temperature scaling or guidance. Reported accuracies average seeds 42, 43, and 44. Nominal churn is parameterized by $S_{\rm churn}=\gamma(N-1)$ and capped at $\sqrt{2}-1$ in the EDM update; consequently the nominal $0.50$ setting below exceeds the implemented per-step cap. The main table reports the best measured setting for each difficulty, rather than a single shared nominal setting.

\begin{table}[htbp]
\centering\small\setlength{\tabcolsep}{5pt}
\begin{tabular}{lrrrrrrrr}
\toprule
Nominal $\gamma$ & 0 & .05 & .10 & .15 & .20 & .30 & .40 & .50 \\
\midrule
Easy & 93.2 & 96.5 & 97.3 & 97.7 & 98.0 & 98.4 & 98.5 & 98.4 \\
Medium & 76.7 & 85.1 & 87.6 & 88.9 & 89.7 & 91.1 & 91.6 & 91.7 \\
Hard & 40.2 & 53.5 & 58.6 & 61.4 & 63.0 & 64.6 & 65.9 & 65.8 \\
\bottomrule
\end{tabular}
\caption{Sudoku exact-match accuracy (\%) at 180 NFE, averaged across three seeds. Nominal churn above the EDM cap is clipped. Seed-wise variance and selection-adjusted intervals are not reported for this table.}
\label{tab:sudoku_churn_full}
\end{table}

\subsection{TinyGSM-to-GSM8K data and evaluation}
\label{app:gsm_protocol}
Training uses TinyGSM, approximately 11.8M synthetic problem--Python-solution pairs \citep{liu2023tinygsm}; evaluation uses all 1,319 GSM8K test problems \citep{cobbe2021gsm8k}. The serialization is \texttt{[BOS] question \textbackslash n code [EOS]}, padded to 512 tokens using the \href{https://huggingface.co/HuggingFaceTB/SmolLM-135M}{SmolLM-135M tokenizer}, with 16 bits per token. The question is held clean; only answer bits are generated. The program defines \texttt{simple\_math\_problem()}. A restricted subprocess executes it with a five-second timeout; its return value must match the reference exactly for integers, or within $10^{-3}$ otherwise. A failed or timed-out program is incorrect. Each problem receives one generated answer; there is no verifier-based selection among candidates.

After integration, the untempered per-bit posterior mean is thresholded at $0.5$ and decoded into token IDs. We report both answer accuracy and the fraction of programs that execute successfully; the latter is not an answer-correctness metric. CoBit-S reaches $335/1319=25.40\%$ at 1,024 NFE after 250K training steps. The unguided temperature sweep instead uses the 500K checkpoint. All paired gains compare configurations with their corresponding $\tau=1$ setting at the same checkpoint, schedule, and sampling budget.

\subsection{Checkpoints and sampling settings}
\label{app:temperature_models}
The new experiments use the following two CoBit-S checkpoints.
\begin{itemize}
\item \textbf{CoBit-S, unguided:} 12 blocks, width 768, 12 heads, feed-forward width 3072 (130M parameters); 500K training steps, batch size 512; AdamW with learning rate $3\times10^{-4}$, 2,500 warm-up steps followed by a constant rate, no weight decay, and $\beta_2=0.999$. EMA decay is 0.9999. Sampling uses 1,024 denoiser evaluations on the learned entropic grid over $[0.002,80]$, obtained by inverting the entropy-rate profile estimated during training.
\item \textbf{CoBit-S with conditioning dropout:} the small-model training configuration above, except that 10\% of training examples replace the prompt with a null prefix; trained for 425K steps and sampled on the Karras grid. This checkpoint is distinct from the unguided 500K checkpoint.
\end{itemize}

Unless a control explicitly changes it, sampling uses EMA weights, first-order Euler/DDIM updates, full-band EDM churn $\gamma=0.41$, and $S_{\rm noise}=1.003$. Self-conditioning carries the previous untempered denoiser prediction. The unguided CoBit-S path requires 1,024 denoiser evaluations per sample. The conditioning-dropout sweep was run through the two-branch guidance path even at $w=1$, requiring 2,048 evaluations; $w=1$ is algebraically the conditional model. The controls below alter one sampler setting within a checkpoint. Comparisons between checkpoints do not isolate conditioning dropout, training duration, or schedule.

\subsection{Score temperature and the last-step modification}
For a variance-exploding diffusion, the learned score is $(D_\theta-x)/\sigma^2$. Following the score-temperature operation of \citet{yang2026sigma}, we replace this score by its value divided by $\tau$, equivalently using \cref{eq:score_temperature} in the integration update. Self-conditioning and final readout always use the original $D_\theta$. If classifier-free guidance is enabled, the denoiser is first mixed as
\begin{equation}
 D_w=D_u+w(D_c-D_u),
 \qquad D_{w,\tau}=x+(D_w-x)/\tau.
\end{equation}
Here $D_c$ and $D_u$ use the conditional and null prompts. At $\tau=1$ the original drift is recovered. These operations alter the sampling dynamics; they do not establish that the terminal discrete distribution is exactly proportional to a power of the original sequence probability.

For the unguided small model, we additionally leave the last integration step onto $\sigma_{\min}$ untempered. This is a modification of the global score-scaling procedure, distinct from merely using an untempered output readout. It preserves the entropic grid and churn on the preceding steps. The conditioning-dropout model uses global score scaling on a Karras grid with $\rho=7$ over $[0.002,80]$. We do not replace the main sampler with an ancestral sampler.

\subsection{Statistics and selection}
\label{app:temperature_statistics}
We select temperature, final-step handling, noise schedule, and churn by accuracy on TinyGSM validation before evaluating the selected configurations on GSM8K. The adopted validation construction follows the reference protocol of \citet{deschenaux2026continuous}. Its released configuration reserves 1\% of TinyGSM with split seed 42; the data loader applies the seeded split after tokenization and filtering examples longer than 512 tokens, using the SmolLM-135M tokenizer and no sequence wrapping.\footnote{Reference implementation: \href{https://github.com/jdeschena/s-flm/blob/main/configs/data/tiny-gsm.yaml}{\texttt{configs/data/tiny-gsm.yaml}} (validation fraction and seed), \href{https://github.com/jdeschena/s-flm/blob/main/dataloader.py}{\texttt{dataloader.py}, function \texttt{get\_tiny\_gsm\_dataset}} (filtering followed by the split), and the \href{https://github.com/jdeschena/s-flm/blob/main/scripts/train/tinygsm/sfm_sphere_arch_truncated_adaptive_no_renorm.sh}{released TinyGSM training script} (tokenizer, length, and filtering overrides); accessed 5 November 2026.} The validation fraction therefore applies to the length-filtered corpus, rather than to the approximately 11.8M raw examples. This describes the adopted reference split construction, not a reconstruction of run-specific sample counts from the public code.

All GSM8K temperature cells reported here use seed 0 and the complete test set; additional cells document sensitivity to temperature and sampler settings. Reported 95\% intervals are percentile-bootstrap intervals from 10,000 resamples of GSM8K problems; gain intervals use paired resampling of problem-level outcomes against the corresponding $\tau=1$ cell. Point estimates are rounded independently, so a reported paired gain can differ by 0.1 point from subtracting displayed accuracies. Bootstrap intervals are pointwise, conditional on the checkpoint and sampling configuration. They do not quantify generation-seed or training variability, or variability from repeating configuration selection on different validation samples. Baseline values quoted from \citet{deschenaux2026continuous} have their own evaluation protocol and are not paired with these runs.

A separate unguided 425K checkpoint is excluded because its anomalous failures remained unexplained after an EMA check. The 425K and 500K evaluations therefore do not constitute a controlled comparison of training duration.

\subsection{Unguided small-model sweep}
\begin{table}[htbp]
\centering\small\setlength{\tabcolsep}{4pt}
\begin{tabular}{llrrl}
\toprule
Final step & $\tau$ & Accuracy & Execute & Paired gain (95\% CI) \\
\midrule
Original & 1 & 28.1 [25.8,30.6] & 63.5 & --- \\
Original & .7 & 32.4 [29.9,35.0] & 69.2 & +4.3 [2.4,6.2] \\
Original & .5 & 2.7 [1.8,3.6] & 5.0 & $-25.5$ [$-27.9,-23.1$] \\
Original & .3 & 0.0 & 0.0 & $-28.1$ [$-30.6,-25.8$] \\
Original & .2 & 0.0 & 0.0 & $-28.1$ [$-30.6,-25.8$] \\
Original & .1 & 0.0 & 0.0 & $-28.1$ [$-30.6,-25.8$] \\
Original & .05 & 0.0 & 0.0 & $-28.1$ [$-30.6,-25.8$] \\
\midrule
Untempered & .5 & 34.6 & 72.6 & +6.4 [4.4,8.5] \\
Untempered & .3 & 35.3 [32.7,37.8] & 75.4 & +7.1 [5.1,9.2] \\
Untempered & .2 & 35.1 & 74.0 & +7.0 [4.9,9.1] \\
\bottomrule
\end{tabular}
\caption{CoBit-S, 500K steps, entropic grid, 1,024 NFE and $\gamma=0.41$. Accuracies and execution rates are percentages; gains are percentage points relative to the shared $\tau=1$ baseline. All-zero cells have pointwise bootstrap intervals [0,0]. Absolute accuracy intervals are included where available.}
\label{tab:temp_small_full}
\end{table}

The unmodified grid fails at lower temperatures, but leaving its last step untempered restores accuracy across $\tau\in\{0.5,0.3,0.2\}$. At these settings, invalid-token rates fall from 1.39/40.40/41.06\% to zero. This supports an integration failure, rather than an inherent inability to decode sharpened bitstream trajectories. The increase in executable-program rate accounts for part of the improvement; correctness conditional on execution also changes, so the experiment does not uniquely attribute the gain to syntax repair.

\subsection{Conditioning dropout and classifier-free guidance}
\begin{table}[htbp]
\centering\small\setlength{\tabcolsep}{9pt}
\begin{tabular}{rrrrrr}
\toprule
$w\backslash\tau$ & 1 & .5 & .3 & .2 & .1 \\
\midrule
1 & 25.9 & 30.7 & 31.7 & 31.9 & 0.0 \\
2 & 26.9 & 30.6 & 30.1 & 27.2 & 0.0 \\
3 & 28.0 & 29.3 & 30.1 & 13.3 & 0.0 \\
5 & 26.6 & 29.2 & 28.7 & 4.6 & 0.0 \\
9 & 25.2 & 27.6 & 28.9 & 0.1 & 0.0 \\
12 & 24.6 & 27.5 & 29.0 & 0.0 & 0.0 \\
\bottomrule
\end{tabular}
\caption{Full classifier-free-guidance/temperature grid: accuracy (\%) for the CoBit-S conditioning-dropout checkpoint at 425K steps, Karras sampling, $\gamma=0.41$, and 2,048 NFE. An additional $(w,\tau)=(1,0.05)$ cell scores 0\%.}
\label{tab:temp_cfg_grid}
\end{table}

At $w=1$, temperature improves the baseline 25.9\% [23.6,28.3] to 31.9\% [29.4,34.5] at $\tau=0.2$, a paired gain of 6.0 [3.9,8.1] points. Execution rate rises from 60.5 to 75.2\%. The other unguided cells are 30.7\% [28.3,33.2] at $\tau=0.5$ and 31.7\% [29.2,34.2] at $\tau=0.3$, with paired gains 4.8 [2.7,6.8] and 5.8 [3.8,7.8] points. Guidance alone peaks at $w=3$ with 28.0\%. Every guided row has a lower best accuracy than the best unguided row. Stronger guidance also makes low-temperature failure occur sooner. This is evidence about this checkpoint and grid; a null branch trained with 10\% prompt dropout on a 130M model is not a matched test of the larger guidance-trained models in \citet{yang2026sigma}.

\subsection{Numerical diagnostics and sampler controls}
\label{app:temperature_numerics}
A churn step raises $\sigma_i$ to $\hat\sigma_i=(1+\gamma)\sigma_i$, adds noise, and then applies an Euler update to $\sigma_{i+1}$. If the denoiser is held fixed for a local diagnostic, the displacement from its output evolves as
\begin{equation}
 x_{i+1}-D=\left(1-\frac{r_i}{\tau}\right)(\hat x_i-D),
 \qquad r_i=1-\frac{\sigma_{i+1}}{\hat\sigma_i}.
 \label{eq:temperature_stability}
\end{equation}
The resulting scalar contraction requires $\tau>r_i/2$. On the Karras grid at 1,024 NFE with $\gamma=0.41$, typical $r_i\simeq0.30$, suggesting a threshold near 0.15. At $\tau=0.1$, diagnostics show nonfinite states early in the trajectory, before meaningful tokens form. The entropic grid's last jump from 0.157 to 0.002 instead has $r_i$ near one. Sharpening it creates a large extrapolation at the point where churn has just added noise. Leaving it untempered eliminates the invalid-code explosion in \Cref{tab:temp_small_full}.

This frozen-denoiser calculation is a local diagnostic, not a stability theorem for the nonlinear learned sampler. In particular, the observed failure at $\tau=0.5$ cannot be inferred solely from violation of the scalar inequality. The final-step modification also does not repair the deterministic sampler's low-temperature collapse: at $\tau=0.5/0.3/0.2$, deterministic accuracy remains 4.4/0.3/0.4\% after the modification, with zero invalid codes. The stochastic and deterministic failures therefore need not have the same cause.

\begin{table}[htbp]
\centering\small\setlength{\tabcolsep}{7pt}
\begin{tabular}{rrrrrrrr}
\toprule
$\tau$ & 1 & .7 & .5 & .3 & .2 & .1 & .05 \\
\midrule
Deterministic, 1,024 & 12.2 & 24.7 & 4.4 & 0.3 & 0.4 & 0.9 & 1.1 \\
Churn, 512 & 26.8 & 31.5 & 2.2 & 0.0 & 0.0 & 0.0 & 0.0 \\
\bottomrule
\end{tabular}
\caption{CoBit-S 500K controls on the entropic grid, accuracy (\%), using the original final step. The 512-NFE gain at $\tau=0.7$ is +4.8 [2.7,6.9] points, compared with +4.3 [2.4,6.2] at 1,024 NFE. Similar gains at these two budgets argue against a benefit specific to a single discretization.}
\label{tab:temperature_step_controls}
\end{table}

At $\tau=1$ with 1,024 NFE and churn, the small model scores 28.1\% [25.8,30.6] on the entropic grid and 26.0\% [23.7,28.4] on the Karras grid (execution rates 63.5 and 59.3\%). This sensitivity comparison describes the effect of the grid at the untempered setting.

\begin{table}[htbp]
\centering\small\setlength{\tabcolsep}{7pt}
\begin{tabular}{rrrrrrr}
\toprule
$\tau$ & 1 & .7 & .5 & .3 & .2 & .1 \\
\midrule
Ancestral, no churn & 16.6 & 24.6 & 28.5 & 28.7 & 29.0 & 16.4 \\
\bottomrule
\end{tabular}
\caption{Variance-exploding ancestral sampling control, small model, entropic grid, 1,024 NFE, final step untempered. Its best cell, 29.0\% [26.5,31.4], improves over its own baseline by +12.4 [10.0,14.8] points but remains below the best churn result.}
\label{tab:temperature_ancestral}
\end{table}

\begin{table}[htbp]
\centering\small\setlength{\tabcolsep}{9pt}
\begin{tabular}{rrrrrr}
\toprule
$\gamma\backslash\tau$ & 1 & .5 & .3 & .2 & .1 \\
\midrule
.10 & 26.8 & 33.0 & 33.6 & 32.4 & 32.0 \\
.20 & 28.0 & 33.3 & 33.4 & 33.5 & 33.3 \\
.30 & 29.0 & 33.4 & 33.0 & 34.3 & 0.0 \\
.41 & 28.1 & 34.6 & 35.3 & 35.1 & --- \\
\bottomrule
\end{tabular}
\caption{Lower-churn controls, small model, entropic grid and untempered final step; accuracy (\%). The $\gamma=.41$ row is reproduced from \Cref{tab:temp_small_full}; its untempered-final-step $\tau=.1$ cell was not reported.}
\label{tab:temperature_churn_controls}
\end{table}
Lower churn permits smaller stable temperatures, consistent with \cref{eq:temperature_stability}, but does not improve the best accuracy over $\gamma=0.41$ within the reported grid. The data support retaining the entropy-rate grid with churn for the main result. They do not by themselves isolate every mechanism responsible for the ancestral control's lower accuracy.

Ending the trajectory at $\sigma_{\min}=0.1$ with a sharpened final step gives 28.1/32.4/34.6/35.0/0.0\% at $\tau=1/.7/.5/.3/.2$, respectively. Leaving that last step untempered gives the same answers as ending at 0.002 for the tested $\tau=.3$ and $.2$ cells (35.3 and 35.1\%). Thus truncating the low-noise tail alone is insufficient at the lowest of these temperatures.

A separate valid-codeword logit-temperature control gave 28.1, 28.5, and 29.5\% at $T=.5,.3,.1$, with zero invalid tokens. Its best paired gain was +1.4 [$-0.6$,3.3] points. This operation uses a softmax over valid codewords, unlike the coordinatewise $O(\log V)$ score update, and its computational cost was not measured in these experiments. Its implementation should not be conflated with \cref{eq:score_temperature}.

\section{Qualitative evaluation}
\label{app:qualitative_evaluation}
Uncurated examples expose repetition and discourse inconsistency at both scales. We do not quantify repetition over the full evaluation batches; task accuracy complements rather than validates open-ended generation quality.

\begin{longtable}{>{\raggedright\arraybackslash}p{0.95\textwidth}}
\caption{16 randomly selected generated samples from the LM1B checkpoint.} \label{tab:lm1b-samples} \\
\toprule
\textbf{Generated Sample} \\
\midrule
\endfirsthead
\caption[]{16 randomly selected generated samples from the LM1B checkpoint. (Continued)} \\
\toprule
\textbf{Generated Sample} \\
\midrule
\endhead
\midrule
\multicolumn{1}{r}{\textit{Continued on next page}} \\
\endfoot
\bottomrule
\endlastfoot
baker ' s three - run homer gave houston a four - game lead for the sixth time in six games. [CLS] ( cnn ) - - two women have been arrested on suspicion of aggravated murder and raping a 15 - year - old california girl after she was fatally stabbed to death by her 18 - year - old brother peggy meyer. [CLS] moscow ( ap ) - russia on sunday criticized the european union yesterday and appealed for failing to broker a peace deal with separatists in south ossetia, georgia, the impoverished regions of abkhazia and separatist regions. [CLS] " but we ' re very confident, for one reason a \\
\midrule
at times. * we welcome comments that advance the story directly or with relevant tangential information. [CLS] crittenton has long announced a 92 \% increase in property sales at the olympics, the closing of a stadium in madison square garden and the 27, 000 - square - foot swimming pool in the bronx. [CLS] kuerate said he had no intention of commenting on ticket sales. [CLS] the metropolitan police investigation department is investigating the case of a teenager who opened an inquiry into the death of her 18 - year - old son, who was murdered in march. [CLS] lavrov privately described the threat as " unacceptable " and called for a " united \\
\midrule
north korea on march 27 imposed three rounds of sanctions against the country ' s five permanent permanent permanent permanent permanent security council members - - china, russia, south korea, russia, china, canada, china, china, russia, japan, south korea, japan and south africa. [CLS] tehran, iran ( ap ) - - beijing says it will justify president mahmoud ahmadinejad ' s pledge to responsibly hold talks next week to end a crackdown on tiananmen square protests, roiling apart concerns that beijing could prop up trade ties amid the economic downturn. [CLS] defending champion djokovic, out in the quarter - finals \\
\midrule
s drug makers pfizer corp. and pfizer inc rose 32 cents, or 7 percent, to \$ 2. 76 after the drugsmaker lowered its profit forecast for the full year, even in line with analysts ' expectations. [CLS] the irony of this report is that for these reasons, i don ' t think facts are irresponsible. [CLS] but i think it will stabilize the readers that attends in the months ahead. [CLS] carmelo anthony added 27 points, nine rebounds and six rebounds for the celtics, who are 2 - 0 in 13 games against kansas city. [CLS] since then, her body has curtailed her, leaving \\
\midrule
to continue its overdraft program. [CLS] if a credit crunch took place at this rate losses would have succeeded on the other end of the real estate market. [CLS] the yield on the benchmark 10 - year treasury note, called yield, fell to 3. 83 percent from 3. 85 percent late friday. [CLS] biden repeatedly reaffirmed international deadly support for u. s. troops deployed near the border with afghanistan and said aid agencies attacked militants on pickups to carry out attacks in south waziristan as part of an operation launched by the violent regime of president pervez musharraf in november by the u. s \\
\midrule
roadside bombs. [CLS] hamilton won his fifth race in ferrari following the loss of defending champion kimi raikkonen to number one ferrari kimi raikkonen by eight points. [CLS] washington ( ap ) - - fannie and freddie rose by scribbled a \$ 1. 5 trillion bailout of the real estate industry and endorsing tax breaks for homeowners. [CLS] 9 / prnewswire / - - the boeing board of directors will announce the union ' s announcement today that uaw chief executive prof. charles mueller, who has been appointed as vice president of underage operations, is pleased to announce boeing ' s retirement \\
\midrule
. flaunt. [CLS] " it ' s gay in tokyo. " [CLS] the attacks on beirut were led by ali saleh, the leader of the niger movement, the sunni group responsible for the assassination of assassinated former prime minister saad hariri in 2005. [CLS] volvo, toyota and russian automakers are pressing ahead with gm plans to purchase smaller sport utility vehicles by the end of this year, with the aim of washing business - level ones to the japanese market in july. [CLS] harare, zimbabwe ( ap ) - - zimbabwe ' s electoral commission says president hamid karzai is committed to starting a runoff result in elections, according \\
\midrule
percent, to 943. 79, and the nasdaq composite index fell 67. 62, or 0. 12 percent, to 2, 113. 26. [CLS] the tel aviv health department identified the deaths of at least nine people homeless after the 7. 7 - magnitude earthquake rocked its temblor about 4 : 30 p. m. [CLS] although the international monetary fund looks like a service for the imf and the association would readily apologise for this conservanor, it would be hard to pay off while delivering the unoisive thirst with the i - b - k - j - m - o - k \\
\midrule
fighters. [CLS] the state gas group, the brazilian natural gas and supply chain, is expected to announce that it is engaged in talks with its former rival, state - controlled gas giant gazprom, about the € 5. 4bn investment offer rejected by german chancellor angela merkel because it hoped the upcoming provincial elections would bolster confidence in the opposition social democratic party, the right - wing business and liberal democrats. [CLS] the televised meeting appeared seen as china ' s top two ministers undermine diplomatic ties with beijing by negotiating with washington, and some chinese officials said they hoped such a move would boost ties with the us. [CLS] those fish may \\
\midrule
july 2008. [CLS] paulson ordinary the housebuilder if in far good faith to make financial woes unwelcomede out of the bailout to always jeopardize the struggling economy. [CLS] a public hearing is scheduled this week to decide whether councillors can take disciplinary action. [CLS] a lorry driver died following an accident on a lane in londonderry early on saturday. [CLS] they paved the way for the far - reaching target of reducing its greenhouse gas emissions by 60 percent by 2050. [CLS] ann arbor, mich., july 30 / prnewswire - usnewswire / - - sept. \\
\midrule
. [CLS] recent court papers show that northern rock has worked unnecessarily with the haringey council. [CLS] dallas ( ap ) - jamal guerrero hit a two - run homer and bobby abreu hit a three - run homer, as the tampa bay rays ended a five - game losing streak with a 8 - 4 victory over the yankees on saturday night. [CLS] sonia sotomayor ' s lawyer, supreme court justice marcus lambert qc, said the senator goes closer to losing her race for the gop nomination. [CLS] next article in sports ( 16 of 49 ) » a version of this article appeared in print on january 9, \\
\midrule
, ups, and his son, objectives. [CLS] in midmorning trading, the dow jones industrial average fell 17. 77, or 0. 59 percent, while the standard \& poor ' s 500 index added 0. 16 percent, and the nasdaq composite index rose 25. 41, or 0. 29 percent, to 1, 808. 24. [CLS] today, carolyn villay, apple ' s president and chief executive officer, was to attend a seminar on the memorial day for orthopaedic doctors president and chief technology officer jared m. loughner and several senior medical experts. [CLS] among the dead was a \\
\midrule
type 2 diabetes, is one of the most vulnerable people in the world to pomegranate during the spreadsheets. [CLS] the plant, which has reserves of 3m tonnes and 2. 5m tonnes of electricity per year, is currently at an average of 1, 000 megawatts ( 200, 000 tonnes ) a year. [CLS] while united nations director general sarath fonseka and turkish president abdullah abdullah prepare a visit to the capital today, afghan political leaders are lobbying the u. n. - backed coalition, led by karzai and president abdullah abdullah, for an agreement requiring president george w. bush to restru \\
\midrule
he earned a bachelor ' s degree in public health from harvard business school, a bachelor ' s in psychology, a bachelor ' s from harvard university school of business in new york and a bachelor ' s degree in psychology. [CLS] inbev yesterday took reference to anheuser - busch ' s bid, after a pyrotechnic investor agreed to rescheduled the world ' s largest cellphone maker. [CLS] moscow, july 26 ( upi ) - - russian president viktor yanukovych is escalating a cease - fire to destabilize ukraine and defend georgia, insisting it wants to communicate with georgia. [CLS] \\
\midrule
chief executive officer of the fa, jim ogden, said that the deal was " unbrieldy. " [CLS] nobody has been told to come forward or to apologise to anyone who witnessed the theft that meant him or her mother. [CLS] a police spokeswoman said the driver of the malibu, who has not been identified, was no longer in critical condition but failed to comply with the order. [CLS] pc david scott, 18, from dagenyrshire, was due to appear before magistrates at the town ' s bristol magistrates court later. [CLS] the defendant is due to appear as witnesses later. [CLS] it ' s no surprise \\
\midrule
the president ' s covers, cnn reported sunday. [CLS] the four - time pro bowl selection crucially as the trojans ' assistant offensive coordinator before the season - ending n. f. l. season, mike hanlon, came off the bench. [CLS] the two security officials spoke on condition of anonymity because they were not authorized to release information to the media. [CLS] liverpool : rooney, denilson, carragher, giggs, terry, lampard ; rooney, malouda, essien, rooney, gerrard, mccormack, carrick, malouda, giggs. [CLS] davydenko \\
\end{longtable}

\begin{longtable}{>{\raggedright\arraybackslash}p{\dimexpr\textwidth-2\tabcolsep\relax}}
\caption{4 randomly selected generated samples from the OWT checkpoint. Sampling configuration: NFE=256, $S_{\mathrm{churn}}=33.15$, $\gamma=0.13$, $S_{\mathrm{noise}}=1.003$, entropy window $[0,1]$, $\sigma_{\mathrm{term}}=0.0794$, ATI $\eta=0$, $(\text{GenPPL},\text{entropy})=(27.06,5.26).$} \label{tab:owt-samples-nfe256-ch33-15-q0-1} \\
\toprule
\textbf{Generated Sample} \\
\midrule
\endfirsthead
\caption[]{4 randomly selected generated samples from the OWT checkpoint. Sampling configuration: NFE=256, $S_{\mathrm{churn}}=33.15$, $\gamma=0.13$, $S_{\mathrm{noise}}=1.003$, entropy window $[0,1]$, $\sigma_{\mathrm{term}}=0.0794$, ATI $\eta=0$. (Continued)} \\
\toprule
\textbf{Generated Sample} \\
\midrule
\endhead
\midrule
\multicolumn{1}{r}{\textit{Continued on next page}} \\
\endfoot
\bottomrule
\endlastfoot
which said it would prevent a "chart" changes in law. \\[0.5em]
We'll tell you what's true. You can form your own view. From 15p EUR 0.18 \$0.18 \$0.27 a day, more exclusives, analysis and extras. \\[0.5em]
The move came after Greece invested more than \$400 million over its EU's six year-old euro, citing the controversial living protections that prevented "goingly" the countryelfra Broadla Smith, chief executive of the International Poverty Law Center. \\[0.5em]
We'll tell you what's true. You can form your own view. \\[0.5em]
At The Independent, no one tells us what to write. That's why, in an era of political lies and Brexit bias, more readers are turning to an independent source. Subscribe from just 15p a day for extra exclusives, events and ebooks -- all with no ads. \\[0.5em]
Subscribe now\textless{}\textbar{}endoftext\textbar{}\textgreater{}0 demonstrators. \\[0.5em]
Shape Created with Sketch. Labour activists wearing the Turkish flag in Ankara Show all 6 left Created with Sketch. Shape Created with Sketch. Labour activists wearing the Turkish flag in Ankara 1/9/10 1/12 RSS clashes from Turkey were drawn byuddin Sallling, after the demonstrators march at WHO Street taken byRich Brotherhood President Assad. \\[0.5em]
Image copyright Getty Images 4/10 RSS march at WHO Street taken byRich Brotherhood President Assad Abdul Burmi - 5/10 RSS march at WHO Street taken byRich Brotherhood President Assad Abdul Burmi - 7/10uddin Sallling, after the demonstrators march at WHO Street taken byRich Brotherhood President Assad Abdul Burmi from Turkey were drawn by leaderuddin Sallling, after the demonstrators march at WHO Street taken byRich Brotherhood President Assad Abdul Burmi - 9/10 RSS clashes from Turkey were drawn by leaderuddin Sallling, after the RSS march at WHO Street taken byRich Brotherhood President Assad. \\[0.5em]
Image copyright Getty Images 10/10 RSS march at WHO Street taken byRich Brotherhood President Assad. \\[0.5em]
Image copyright Getty Images 1/12 leaderuddin Sallling, after the demonstrators march at WHO Street taken byRich Brotherhood President Assad Abdul Burmi - 11/10 \\[0.5em]
Dear fathers, As you can imagine, more people are reading The Jerusalem Post than ever before. Nevertheless, traditional business models are no longer sustainable and high-quality publications, like ours, are being forced to look for new ways to keep going. Unlike many other news organizations, we have not put up a paywall. We want to keep our journalism open and accessible and be able to keep providing you with news and analysis from the frontlines of Israel, the Middle East and the Jewish World. \\[0.5em]
Egypt's prime minister, Majambeatil Abdul Abdjad, announced he would visit the backing ofuddin Sallling and, his vice-president, also had been to go in office prior to the election. \\[0.5em]
Monday's once-bearing attacks have occurred just days after Al Qaeda led its in a series of protests over the time of Thursday's demonstrations in Palestine, anticipating 2,500,000 people after three weeks of bilateral protest. \\[0.5em]
Shape Created with Sketch. Shape Created with Sketch. The Israeli embassy showed the murder of San Ahmad Mubarak 10 left Created with Sketch. right Created with Sketch. Shape Created with Sketch. The Israeli embassy showed the murder of San Ahmad Mubarak in Seattle, Tuesday, April 24, 2011. REUTERS/Chank Klail/UPI/BIDOWHUREDERS (K. Cohen/Gary Webbckman) on Friday night before protests in parts of Palestine, Calif., the clash between the Israeli Foreign Minister Bashar al-Assad and the Israeli prime minister over the murder of San Ahmad Mubarak in Seattle, Tuesday, April 24, 2011. REUTERS/Chank Klail/UPI/BIDOWHUREDERS (K. Cohen/Gary Webbckman) on Friday night before protests in parts of Palestine, Calif., the clash between the Israeli Foreign Minister Bashar al-Assad and the Israeli prime minister over the murder of San Ahmad Mubarak and the Israeli prime minister over the murder of San Ahmad Mubarak in Seattle, Tuesday, April 24, 2011. REUTERS/Chank Klail/UPI/BIDOWHUREDERS (K. Cohen/Gary Webbckman) \\[0.5em]
"It marks the strongest economic and ethnic over the last two years," Jibh Shah, a spokesman for the Hamas group in northern Israel, said in a statement on Friday night before protests in parts of Palestine, Calif., the escalation of the receiver investment market, rising spending on residentialhouses. \\[0.5em]
"With changes in capital gaining income, the Israeli government is continuing \\
\midrule
. Users who will sign the federal name \\[0.5em]
With an older senior being issued the federal name on Feb. 30 (main end of March), Users will immediately update the federal name in an March 22 procedure session. Users graduate on Feb. 30 (main end of March) and an older senior has issued the federal name in an March 22 procedure session. Users will immediately update the federal name \\[0.5em]
This is the first "hirtating" event. This post will contain all the potential kind of parties who will question twenty stewards! \\[0.5em]
1 : \\[0.5em]
3. South America from East American Draft: 2017 05:80 2018--2008-17 04:38 pm PDT \\[0.5em]
South Florida players return to MSU by starting Houston (GLAA), and they've been very familiar with taking this position. South Florida, Virginia, Houston and Los Angeles will now want to stay out of holiday of May 1, 2019! There is a complete selection of new front office veterans in anticipation of 2018! \\[0.5em]
On May 1, 2019, both players could get together with the qualifier all year 2017! \\[0.5em]
There is a complete selection of new front office veterans in anticipation of 2018! \\[0.5em]
On May 1, 2019, both players could get together with the qualifier all year 2017! 2 : \\[0.5em]
3. South America from East American Draft: 2017 05:80 2018--2008-17 04:38 pm PDT \\[0.5em]
Players return to MSU by starting Houston (GLAA), and they've been very familiar with taking this position. \\[0.5em]
The Miami Buccaneers will be officially assigned to Kansas City in the straight batting finals of 2016-16 MSU Saturday, Aug. 3, 2018. \\[0.5em]
The Miami Buccaneers will be officially assigned to Kansas City in the straight batting finals finals of 2016-16 MSU 3 : .The NFL Draft isn't set (player terms are permitted); a new set of starting stars will be scheduled to Milwaukee from MSU Sunday, Aug. 3, 2018. \\[0.5em]
.The NFL Draft isn't set (player terms are permitted); a new set of starting stars will be scheduled to Milwaukee from Thursday Aug. 3, 2018.\textless{}\textbar{}endoftext\textbar{}\textgreater{}This story was updated to become a senior scientist in physics and biology at the University of California, Berkeley. \\[0.5em]
Read next: \\[0.5em]
"Uulu Goldkallic is "the founder of a deepest-most-infused form of sclerosis?" Goldkallic has also been appointed as the lead author for last month's Journal of Medical Archaeical Scientific. \\[0.5em]
Dr. Nevsson, an archaeologist student at Northwestern College of Medical Archaeical, said the new research is recently published by the journal Health Reciv Farm and Feb. 9 journal Nature Archaeical. \\[0.5em]
"When this research suggests that reduced spinal nerves can be associated with reduced phenectinal symptoms predictability corrupted elsewhere, or the combination of reduced cancer and phenectinal symptoms obstruct the effects of the counterpototiens culture," he said. \\[0.5em]
By realizing increased cancer impairability and predictability and increased cognitive impairability, researchers have adopted conclusions about whether products such as antibiotics could potentially function to handstrike the cancer or result that they are able to obstruct brain impairability and predictability. \\[0.5em]
Mr Nevsson, co-authors and associate in neurophysivism at the National University of California, Public Health at the University of Berkeley said: "This preliminary work has been handled to power our exposure in human culture and the scientific understanding of what our brain functions are, including different kinds of physical functions and behavioral disappointments."\textless{}\textbar{}endoftext\textbar{}\textgreater{}Image copyright AP Image caption The virus ignited in Australian low-conceived populations when it left died on jampary two months ago \\[0.5em]
A Canadian scientist has learned about the discovery of the virus by former met speculationologist Isaiah Gorel Mythley told Radio 4 Radio. \\[0.5em]
The researchers think the virus could still spell an devastating process found in bacteria that have spun on jampary and the national wildfire. The virus has been infected. \\[0.5em]
Chery Agyler, 61, a non-supophysologist at Victoria, said he has no reason to man "its" evidence. \\[0.5em]
"Fasassing this discovery is a move which only knows and keeps its arguments," he added. \\[0.5em]
"There is an logical exchange between large parts of Australian populations which only carry places where they tend to access and prevent catastrophic bacteria," said Agyson. \\[0.5em]
He says the toxic virus such as the function needed to preempt the Australian low-conceived populations \\[0.5em]
Scientists believe the grossly outwarded virus was "daved by citizens into the wildfire as a damaging and difficult disease". \\[0.5em]
'A British kendo passages' \\[0.5em]
Image caption Science Australia is reviewing a complaint against bacteria that occurred in the UK and Northern Japan, the researchers say \\[0.5em]
The intense discovery of the bacteria is found in more than 100,000 British kendo passages and other populations with only a network of chunks (from up to 200,000 km or so), Agyson said. \\[0.5em]
The virus is published in the journal Physical Science \\
\midrule
when it wants residents to make its way towards the corner of building. \\[0.5em]
Scoc partico announced in February that Seattle's decision to launch that building will allow the company "to fulfill its policy of employing local communities." \\[0.5em]
While it allows residents to purchase roads in every area last week, Guoc partico has already urged the city to stand with the construction of 202nd Street, Twin United said to help engage the next owners of the building. \\[0.5em]
Business for greatness Read more \\[0.5em]
In early 2015, Guoc partico signed an agreement with the city for a housing kit on 202nd Street. \\[0.5em]
Scoc partico also reached out to the city to operate the chemical algae in the city in the new kit, adding that he has not yet maintained a educational union. \\[0.5em]
"Industry contributions can be held. The downside is really many residents feel in the field," Twin United said. \\[0.5em]
White House chairman Seth Evans, but Guoc partico said he hopes the district will pull the brink of diarrhea while expanding the area. \\[0.5em]
"Young residents with mutual rights will help less and improve atrocities than due to resistance or flashors across the country," Evans said. "This is important that we try to help create fathers with business and living in community based on realities that resonate with local politics."\textless{}\textbar{}endoftext\textbar{}\textgreater{}Sen. inducingicon Valley: Senators urge Trump to bar Huawei products from electric electric grid \textbar{} Ex-security officials Trump emergency declaration \textbar{} New malicious cyber tool found \textbar{} Facebook faces questions on treatment of moderators moderators \\[0.5em]
Photos: Photos: former US Bre crane Ministry of Richmond \\[0.5em]
RC-23 supporters appear during a protest at thestownard Street in Fairfax Station on Sunday, October 26. Hide Caption 6 of 39 Photos: Photos: former US Bre crane Ministry of Richmond \\[0.5em]
RC-23 Prime Minister Rick Sware speaks to journalists at thestownard Street in Fairfax Station on Sunday, July 26. Hide Caption 7 of 39 Photos: Photos: former US Bre crane Ministry of Richmond \\[0.5em]
RC-23 supporters during protest besidestownard Street. \\[0.5em]
Photos: Photos: former US Bre crane Ministry of Richmond \\[0.5em]
RC-23 supporters Prime Minister Rick Sware speaks to journalists at thestownard Street in Fairfax Station on Sunday, July 26. Hide Caption 9 of 39 Photos: Photos: former US Bre crane Ministry of Richmond \\[0.5em]
RC-23 at Richmond International Airport Sunday, October 25. Hide Caption 10 of 39 Photos: Photos: former US Bre crane Ministry of Richmond \\[0.5em]
RC-23 International Airport Sunday, October 25. Hide Caption 11 of 39 Photos: Photos: former US Bre crane Ministry of Richmond \\[0.5em]
RC-23. Hide Caption 12 of 39 Photos: Photos: former US Bre crane Ministry of Richmond \\[0.5em]
RC-23 Prime Minister Rick Sware (51) speaks to journalists at thestownard Street in Fairfax Station on Sunday, October 26. Hide Caption 13 of 39 Photos: Photos: former US Bre crane Ministry of Richmond \\[0.5em]
RC-23 supporters during a protest besidestownard Street in Fairfax Station on Sunday, October 26. Hide Caption 14 of 39 Photos: former US Bre crane Ministry of Richmond \\[0.5em]
RC-23 supporters protest besidestownard Street in Fairfax Station on Sunday, October 26. Hide Caption 15 of 39 Photos: Photos: Photos: former US Bre crane Ministry of Richmond \\[0.5em]
RC-23 besidestownard Street in Fairfax Station on Sunday, October 26. Hide Caption 16 of 39 Photos: Photos: former US Bre crane Ministry of Richmond \\[0.5em]
RC-23 International Airport Sunday, October 25 \textbar{} AP AFP\textless{}\textbar{}endoftext\textbar{}\textgreater{}Ready to fight back? Sign up for Take Action Now and get three actions in your inbox every week. You will receive occasional promotional offers for programs that support The Nation's journalism. You can read our Privacy Policy here. Sign up for Take Action Now and get three actions in your inbox every week. \\[0.5em]
Thank you for signing up. For more from The Nation, check out our latest issue \\[0.5em]
Subscribe now for as little as \$2 a month! \\[0.5em]
Support Progressive Journalism The Nation is reader supported: Chip in \$10 or more to help us continue to write about the issues that matter. The Nation is reader supported: Chip in \$10 or more to help us continue to write about the issues that matter. \\[0.5em]
Fight Back! Sign up for Take Action Now and we'll send you three meaningful actions you can take each week. You will receive occasional promotional offers for programs that support The Nation's journalism. You can read our Privacy Policy here. Sign up for Take Action Now and we'll send you three meaningful actions you can take each week. \\[0.5em]
Thank you for signing up. For more from The Nation, check out our latest issue \\[0.5em]
Travel With The Nation Be the first to hear about Nation Travels destinations, and explore the world with kindred spirits. Be the first to hear about Nation Travels destinations, and explore the world with kindred spirits. \\[0.5em]
Sign up for our Wine Club today. Did you know you can support The Nation by drinking wine \\
\midrule
"Dear Reader, As you can imagine, more people are reading The Jerusalem Post than ever before. Nevertheless, traditional business models are no longer sustainable and high-quality publications, like ours, are being forced to look for new ways to keep going. Unlike many other news organizations, we have not put up a little news." \\[0.5em]
SCROLL TO CONTINUE WITH CONTENT Help Keep Common Dreams Alive Our progressive news model only survives if those informed and inspired by this work support our efforts \\[0.5em]
Advertisements\textless{}\textbar{}endoftext\textbar{}\textgreater{}President Hillary Clinton Hillary Diane Rodham ClintonREAD: Cohen testimony alleges Trump knew Stone talked with WikiLeaks about DNC emails County GOP in Minnesota shares image comparing Sanders to Hitler Holder: 'Time to make the Electoral College a vestige of the past' MORE and Bernie Sanders Bernard (Bernie) SandersPush to end U.S. support for Saudi war hits Senate setback Sanders: 'I fully expect' fair treatment by DNC in 2020 after 'not quite even handed' 2016 primary Sanders: 'Damn right' I'll make the large pay 'fair share of share of taxes' Former MORE (I-Vt.) has also regularly expressed concern about his inability to stay in Syria to cut the deficit transition. \\[0.5em]
At Monday of The Hill's Morning, Sen. Dianne FeinsteinHillicon Valley: Senators urge Trump to bar Huawei products from electric grid \textbar{} Ex-security officials condemn condemn Trump emergency declaration \textbar{} New malicious cyber tool found Facebook faces faces questions treatment of moderators Key senators say administration should ban Huawei from tech in US electric grid Cots unflakably in effect after Politico reports Five reported MORE (D-N.Y.). and Rep. Charles SchumerJoe (Mitch) SchumerJoe TrumpWhite Senate GOP on emergency declaration Houseads with Senate GOP on emergency declaration Senate Report against Trump The Hill's 12:30 Report: Sanders set to shake up 2020 2020 MORE (R-Wis.) said in an interview Thursday. "If you just start coming down and doing it, it's not a very small part of the political process." \\[0.5em]
The Democrats have been occupational to defend Congress as "right" and some Republicans view them as the Republican nominee, according to a new Poll poll. \\[0.5em]
ADVERTISEMENT \\[0.5em]
Democrats, including Senate Majority Leader Harry Reid Addison James (Jack) Trump The Hill's Morning Report: Sanders set to shake up fight against Trump The Hill's Morning Report: Sanders set to shake up fight over against Trump The Hill's 12:30 Report: Sanders set to shake up 2020 2020 MORE (R-Ky.) said Tuesday in an email Thursday. It's hard to confirm why the statements are accurate. \\[0.5em]
ADVERTISEMENT \\[0.5em]
Sen. Mitch McConnell Davis Paul PaulBrexit and exit: A transatlantic comparison Five Five takeaways from McCabe's allegations against Trump The Hill's's 12:30 Report: Sanders set to shake up ahead of 2020 MORE (R-Ky.) Majority Leader Mitch McConnell James (Bernie) RyanBrexit and exit: A transatlantic comparison pen Five takeaways from McCabe's Trump The Hill's 12:30 Report: Sanders set to shake up 2020 2020 MORE (R-Ky.) Sen. Paul Paul Paul PaulBrexit and exit: A transatlantic takeaways from McCabe's allegations against Trump The Hill's 12:30 Report: Sanders set to shake up 2020 2020 2020 MORE (R-Ky.) said after the start of the economic year, a major trade deal has been negotiated. \\[0.5em]
The Foreign Ministry confirmed Thursday off their announcement from an engaged source telling the state whether the government will end up using the international force to shut out the veto of the world's trade talks. \\[0.5em]
The British newspaper has said that after the start of the economic year, a major trade deal has been negotiated. \\[0.5em]
It is off their announcement from an engaged source telling the state whether the government will end up using the international force to shut out the veto of the world's trade talks. \\[0.5em]
The Foreign Ministry said in a statement Thursday that the UK will integrate a deal into a temporary dispute and could discuss partners in the UK. \\[0.5em]
We'll tell you what's true. You can form your own view. From 15p EUR 0.18 \$0.18 \$0.27 a day, more exclusives, analysis and extras. \\[0.5em]
The European Commission will call for a treaty with the European Union to help resist both the trade deals and the Trans-Pacific Partnership (IPA) agreement with UK leaders. \\[0.5em]
France was "delakbing" at the exact conclusion of an ongoing conflict with the senior countries in 2013. \\[0.5em]
The Foreign Ministry said at a briefing that the deal will consider the conclusion of an agreement which forces the UK to resist its deals between the two countries. \\[0.5em]
Speaking earlier Thursday, the Foreign Ministry said it would not discuss if these \\
\end{longtable}

\begin{longtable}{>{\raggedright\arraybackslash}p{\dimexpr\textwidth-2\tabcolsep\relax}}
\caption{4 randomly selected generated samples from the OWT checkpoint. Sampling configuration: NFE=256, $S_{\mathrm{churn}}=45.9$, $\gamma=0.18$, $S_{\mathrm{noise}}=1.003$, entropy window $[0,1]$, $\sigma_{\mathrm{term}}=0.0794$, ATI $\eta=0$, $(\text{GenPPL},\text{entropy})=(34.83,5.32).$} \label{tab:owt-samples-nfe256-ch45-9-q0-1} \\
\toprule
\textbf{Generated Sample} \\
\midrule
\endfirsthead
\caption[]{4 randomly selected generated samples from the OWT checkpoint. Sampling configuration: NFE=256, $S_{\mathrm{churn}}=45.9$, $\gamma=0.18$, $S_{\mathrm{noise}}=1.003$, entropy window $[0,1]$, $\sigma_{\mathrm{term}}=0.0794$, ATI $\eta=0$. (Continued)} \\
\toprule
\textbf{Generated Sample} \\
\midrule
\endhead
\midrule
\multicolumn{1}{r}{\textit{Continued on next page}} \\
\endfoot
\bottomrule
\endlastfoot
would be added to legal authority. \\[0.5em]
"In practice, it should be the meaningful relief, any of [the Special Capitalism Government] before prodion would be changed," Shahur said. \\[0.5em]
He continued that the PM refused to run the PM but granted plans to prodion tell his Congress. \\[0.5em]
"The questioning of prodion tell his Congress that I do not know how to make entitlements and are subject to those required to file borrowed," Shahur said. \\[0.5em]
"These changes would hardly involve holding the Special Capitalism Government borrowed from rules that are enacted by Parliament," she said. \\[0.5em]
Maur said changes cannot be debated by KOIS in Samamik Bihar, Andures Nadu and Karnataka. \\[0.5em]
Firstpost is now on WhatsApp. For the latest analysis, commentary and news updates, sign up for our WhatsApp services. Just go to Firstpost.com/Whatsapp and hit the Subscribe button.\textless{}\textbar{}endoftext\textbar{}\textgreater{}The Denver Browns will be eating home for NFL coaches and football teams early Friday. A NFL official at the White House said the Browns will be able to grant NFL coaches to sign in the NFL and suggest teams will be once a week carrying ammunition including ammunition, weapons and guns. \\[0.5em]
NEWSLETTERS Get the Gn Wgs newsletter delivered to your inbox We're sorry, but something went wrong Get the pulse of Arizona -- Local news, in-depth state state writers and what. ( means, Ch to your inbox. Please try again soon, or contact Customer Service at 1-888-424-022527. Delivery: Varies Invalid email address Thank you! You're almost signed up for Wgs Keep an eye out for an email to confirm your newsletter registration. More newsletters \\[0.5em]
And Friday, the Denver Browns said it will not extend new arrangements for the coaches to sign in the NFL while playing the best ends backs. \\[0.5em]
Joe Stanal may be reached at...al@globe.com . Follow him @jalachamal, or on Facebook on Facebook.\textless{}\textbar{}endoftext\textbar{}\textgreater{}Carlos Democratic nominee Donald Trump, get on a GOP presidential challenger, drew to criticism over support for the bid on Wednesday to take re-election in April as his Republican chairman and the presumptive Democratic board of directors. \\[0.5em]
The New York GOP nomination is scheduled next week to meet Republican nominee's 3rd side on the Saturday Night Live tour and watch the GOP's first-ever sitcom TV movie and will reportedly be portrayed as fastest for "goldors" for presumptive elections on the Sunday Night Live starting on Wednesday and Tuesday. \\[0.5em]
We'll tell you what's true. You can form your own view. From 15p EUR 0.18 \$0.18 \$0.27 a day, more exclusives, analysis and: 'I fully expect' fair treatment by DNC in 2020 after 'not quite even handed' 2016 primary Sanders: 'Damn right' I'll make the large corporations pay 'fair share of taxes' MORE (I-Vt.), the candidate for the presumptive Democratic nomination and presumptive Republican presidential nominee Donald Trump, was a brother of former Democratic nominee-elect Donald Trump Donald John TrumpREAD: Cohen testimony alleges Trump knew Stone talked with WikiLeaks about DNC emails Trump urges North Korea to denuclearize ahead of summit Venezuela's Maduro says he fears 'bad' people around Trump MORE, former Republican presidential nominee Donald Trump Donald John TrumpREAD: Cohen testimony alleges Trump knew Stone talked with WikiLeaks about DNC emails Trump urges North Korea to denuclearize ahead of summit Venezuela's Maduro says he fears 'bad' people around Trump MORE (I-Vt.), who is organizing Obama's bid for the presumptive Democratic National Committee. \\[0.5em]
ADVERTISEMENT \\[0.5em]
The GOP nomination is scheduled to take re-election in April when he signaled Republican presidential nominee Hillary Clinton Hillary Diane Rodham ClintonSanders: 'I fully expect' fair treatment by DNC in 2020 after 'not quite even handed' 2016 primary Sanders: 'Damn right' I'll make the large corporations pay 'fair share of taxes' Former Sanders campaign spokesman: Clinton staff are 'biggest a--holes in American politics' MORE and Vermont Sen. Bernie Sanders Bernard Bernard (Bernie) Sanders to end U.S. support for Saudi war hits Senate setback Sanders: 'I fully expect' fair treatment by DNC in 2020 after 'not quite even handed' 2016 primary primary Sanders: 'Damn' I'll make the large corporations pay 'fair share of taxes' Former MORE (I-Vt.) on Wednesday. \\[0.5em]
ADVERTISEMENT \\[0.5em]
The New York GOP nomination was originally on the Sunday Night Live starting Friday and Tuesday. \\[0.5em]
We'll tell you what's true. You can form your own view. \\[0.5em]
At The Independent, no one tells us what to write. That's why, in an era of political lies and Brexit bias, more readers are turning to an independent source. Subscribe from just 15p a day for extra exclusives, events and ebooks -- all with no ads. \\[0.5em]
Subscribe now \\
\midrule
camping, wearing galay pabbs." \\[0.5em]
Senior speaker James Simque, a spokesman for the Catholic evangelical Church of Jesus Christ of Latter-day Saints, characterized the "performance" as a tragedy. \\[0.5em]
"It's a situation. It's the tragedy that is necessary," he said. \\[0.5em]
"I'm trying to put them back on that process." \\[0.5em]
Dunque said Lanett was never immediately involved. An incident happened at 5:45 p.m. after shooting at firefighters during a protest at a community fire tower in central Portland. Lanett was responsible for the massacre but suffered numerous mental health issues with no alleged recourse or reason for anyone alert. One avalanche involved four arrests and one in a spinning residence near Archeoth Street and Cager St Englando in a spinning residence near Archeoth Street. \\[0.5em]
The tsunami at 200 corner of Archeoth Street where hospital erupted at about 4:55:30 a.m., Sabden Bell, a nurse, was shot near the 540st Temple of Redlake St. \\[0.5em]
On a deadly scene at 6:05 a.m., local car owner Charlie Taffi said the guns were fired at him and the car quickly slipped inside. \\[0.5em]
Sherden Taffi said all of the men grabbed a pool in the street but spotted Bell in the more while she shot out a gun when she saw a weapon. \\[0.5em]
"I didn't know what to do," he said. "When I was cooking I felt like I was marching in a few times but I couldn't sleep too. I couldn't do a gun. I think made her exit for good." \\[0.5em]
Masteri said Lanett was covered with a "wushling eye pistol." \\[0.5em]
"there's a assembled meal. It's a crime at stake as such," she said. \\[0.5em]
Her session party will be held at the Archeoth Polating College of Dader-Aallinian University near Hue of the Catholic Catholic Church to date Hospital. \\[0.5em]
Anyone with information is on board situation could contact Portland State Police to call her at 307-282-5232. Email her at jlurt@freep.net \\[0.5em]
Follow her on Twitter: @barmlurt\textless{}\textbar{}endoftext\textbar{}\textgreater{}Looking for news you can trust? \\[0.5em]
Subscribe to our free newsletters. \\[0.5em]
The former front-runner reported that Donald Trump will ask former top strategist Michael Flynn to leave the investigation of the possible ties between the Republican nominee Donald Trump during a plan to keep up the FBI office in the 2016 election, according to Kansas Democratic National Convention chairman Tim Gillansson. \\[0.5em]
Trump previously confirmed a new plan to cultivate throughout the Republican-led investigation. \\[0.5em]
"The tool responsible for the investigation is not a second motive," Gillansson told me on Sunday night. \\[0.5em]
The announcement came just a few one day before Trump urged FBI Director James Comey and FBI director Michael Flynn to move forward, unless mandated. \\[0.5em]
"There are few reasons for them to help this investigation," Trump said Wednesday leading through an exclusive interview with the DOJ. \\[0.5em]
Kumann said she also criticized Trump for investigating allegations, saying that the issue "is most improbable." \\[0.5em]
Donald Trump has appointed frequent U.S. Sen. Ted Cruz, who has recently been elected as a top policy adviser to presumptive Republican presidential nominee Donald Trump, and former New Jersey Sen. Lindsey Graham (R-Ill.), Secret Service chair Nathan Brave, who resigned from the Obama administration at the time. \\[0.5em]
Obama has also urged Senate Republicans to extend a plan to ask Comey to figure out the possible ties between the Republican nominee Hillary Clinton. \\[0.5em]
Following a news conference on Tuesday, Gillansson answered. \\[0.5em]
"I encourage you to continue meeting with FBI Director Michael Comey not to get a thorough collection process because the process would not have been negotiated," Brave said in a statement led by Wall Street Journal journalist Brad Mullke. \\[0.5em]
"Deputy is thatEverybody else understand that this very comprehensive investigation sparked the controversy."\textless{}\textbar{}endoftext\textbar{}\textgreater{}Two Republican representatives of Colin Powell announced the notice of pending charges against U.S. Chief of Staff Secretary Clinton, that the preference to lead for foreign policy twice -- the head of the administration Department, was that they actually played a role in determining dealings between the president and private "skame organizations of one country, he added." \\[0.5em]
President Powell sought to update this position from his senior national security adviser. \\[0.5em]
On Thursday, former former House member Fred Issa said he was responding to his testimony during a meeting before the Senate Appropriations Committee to indicate the public acting as an unidentified FBI officer would presumably cause confusion through the establishment of a federal investigation. \\[0.5em]
President Powell sought to update this position from his senior national security adviser. \\[0.5em]
On Thursday, former House member Fred Issa said he was responding to his testimony during a meeting before the Senate Appropriations Committee to indicate the public acting as an unidentified FBI officer would presumably cause confusion through the establishment of a federal investigation.... \\
\midrule
/7)\textless{}\textbar{}endoftext\textbar{}\textgreater{}An Palestinian group that legitimizes Palestinians in the Gaza Strip and the northeastern West Bank slams Mahmoud Mohammed Salah in the West Bank in the east of West Bank. \\[0.5em]
The shooting took place Wednesday, Aug. 12, killing 38 civilians, Palestinian Authority Hamas said, for the latest since the U.S.-Israel coup started. \\[0.5em]
Israel is presenting a rebuke of public protests against Israel and the Palestinian side of the Gaza Strip in two areas, the Foreign Ministry announced Friday. \\[0.5em]
The Guardian revealed that pro-Israel group unveiled a more concerted campaign to attack Palestinians in the West Bank, which is the territory of the northeastern Gaza Strip and northern Gaza. \\[0.5em]
Shortly after the attack was picked up by an Israeli army commander, who was shot from the left leg and suffering from four Sanctings. \\[0.5em]
Israel has reported that Jerusalem and one of the major settlements hit a First West Bank area of Bryuboo that killed seven people, the Foreign Ministry said. The deadly shooting near the city of Jerusalem was another rebuke of public protests against Palestinian people in Gaza. \\[0.5em]
The Foreign Ministry said the Gaza area was occupied by Israel and its current churches, causing devastating harm and mass atrocities amongst Israeli people in Gaza. \\[0.5em]
UN officials said that Israel on Saturday demanded a crackdown against Palestinian President Mahmoud Mohammed Salah. \\[0.5em]
A First West Bank shooting did not address reports. No arrests were needed due to crimes, and Mohammed Salah did not respond to request for comment. \\[0.5em]
A government source said the shooting has been secretly captured by Israeli forces. \\[0.5em]
Israel spokesman Alasser Al Walhha say both victims pose a confrontation between Palestinians and Palestinian President Mahmoud Mohammed Salah. \\[0.5em]
A UN source did not immediately respond to requests for comment. \\[0.5em]
Keep updated: Sign up to our newsletter Email * Please enter a valid email address Sign up Please wait... Thank you for signing up. We've got more newsletters we think you'll find interesting. Click here Oops. Something went wrong. Please try again later. Try again Thank you, The email address you have provided is already registered. Close\textless{}\textbar{}endoftext\textbar{}\textgreater{}San Antonio City Idaho Marquivul's Adutio Sajul Sajala \\[0.5em]
Alter's anch Adutio Sajul Sajala is an international center for health and environmental impacts, according to a report compiled by the San Francisco Register. "San Antonio City Idaho has developed Marquivul's anch Adutio Sajul Sajala," the department said. \\[0.5em]
The risk study presented by the CDC to limit the risk for disease in humans remains "decision" in the past decade and calculates the worst weight memory loss, the department said. \\[0.5em]
Yillivul's anch Adutio Sajul Sajala affects patients, according to the report by The San Francisco Register. \\[0.5em]
The study found an "sasured stall" attached in the risk prevention system to manage the incidence of opioid disease called Marquivul's anch Adutio Sajul Sajala. \\[0.5em]
"Psychletsri Medical opioid disease because of mammalian brain cancer may actually contain its effects," the department said in a statement. \\[0.5em]
The decline in levels of opioid disease has emerged from 2010 to 2007. According to a report, the CDC published its own data from early 1994 to 2009. \\[0.5em]
More than 2.5 million injured opioid victims were refugees, according to the CDC from the U.S. Centers for Disease Control and Prevention. \\[0.5em]
The report compared the number of people with brain disorders related to a study 2008 through 2009 at the National journal Medical Letters that "drug infections and brain disorders are the same as dementia, a small majority of people do not qualify for the infection of the opioid disease," the CDC said. \\[0.5em]
Nearly 60,000 refugees are infected with technology and brain therapy and brain respiratory therapy are rare, the CDC said. \\[0.5em]
In 2010, the number of injured opioid victims visited Idaho, and more than 14,500 refugees have been infected in recent years. \\[0.5em]
Last week, more than 1.3 million opioid victims with brain respiratory therapy have been diagnosed asrome opioid drugs. Some opioid drugs have been naturally associated with infectious diseases including Parkinson's disease or Parkinson's disease, the CDC said. Another myriad virus diseases, such as Alzheimer's disease or Parkinson's lung disease, is now often associated with clinical patients with eye deprivation and other health symptoms. \\[0.5em]
About 8,500 refugees are infected, the CDC said. \\[0.5em]
More than 2.5 million opioid victims were infected in regions Creed Idaho, a report concluded. \\[0.5em]
"San Antonio City Idaho has been contracted to quantify the magnitude of the virus," the CDC said. \\[0.5em]
"The CDC website is also part of the study and offers a dialogue to patients with the Hawann Watson County Sheriff's Office," the report said. \\[0.5em]
Copyright 2014 by Nexstar Broadcasting.com. All rights reserved. This material may not be published, broadcast, rewritten or redistributed \\
\midrule
support for same-sex marriage at a presidential debate in San Diego, thereby citing broad laws eliminating marriage.\textless{}\textbar{}endoftext\textbar{}\textgreater{}ES News Newsletter Enter your email address Please enter an email address Email address is invalid Fill out this field Email address is invalid You already have an account. Please log in or register with your social account \\[0.5em]
Los Angeles Police charged David Duuolo, 37 at 4:50 p.m., who allegedly came from his flat after he was charged on charges, including the pair on a chase at 8580 Bruce Ave., the alleged driver. \\[0.5em]
Mituolo, 37, was returning his 67-year-old son from his apartment in San Antonio, New York at 2:45 a.m.. Photo: Ian Mizzing) \\[0.5em]
San Francisco Police Department detectives said Wednesday they contacted any investigation for one incident. Phil County Officer James Mitchell contacted Waron Sheriff Sgt. Ian Mizzing. The investigation was over, 41 hours before crash, police said. \\[0.5em]
Investigators said they noticed an unarmed man hitting anything from his flat at about 5 a.m. and accidentally injuring his back patrol vehicle, police have said. \\[0.5em]
City police said they might see Duuolo be into a flat, breaking his hi into a flat and killing him on her head, said Waron Phil County Attorney Randy Vyjit. \\[0.5em]
Jeremy Duuolo and his friends were caught near a recovered passenger about 2:50 p.m west of 633 Bruce Avenue Monday. \\[0.5em]
About days after the crash an passenger flight crashed from a road at about 8:30 a.m. Northern in response to the therefore include sustained muscle damage, police said. \\[0.5em]
Officials and the Federal Safety Administration have not provided any details. Vyjit said a police believe that the plane was recovered because Waron officials had been briefed in order to determine whether the plane was recovered. \\[0.5em]
"It's reasonable to assume that this occurred," Vyjit told Business Insidermedia.com. \\[0.5em]
"They believe that the plane was recovered. . . . They do believe that the plane was recovered and that the shootings were made before the crash." \\[0.5em]
The detectives have responded to multiple arrests for the first time. A witness who has not reviewed the incident on Smateau Street, said the student tried to defend him when it attacked. \\[0.5em]
"I don't think I don't don't want my brother to go home anymore," Vyjit said. \\[0.5em]
The student allegedly killed her 61-year-old son. In the front car, despite the guns, the student ran at his secret price, then stopped him for one of the bombs, noting. \\[0.5em]
"Ah, I don't like the student just didn't want to turn me on the ground, but his son's student did not want to imprison him," Vyjit said. \\[0.5em]
Jeremy Duuolo and his friends, whose wife, Martha, went to San Antonio where he never needed to as I as they started cooking. Tulkabce occurred between Smateau Street and Jordll's Park in Parkfield, Mo., Maddery Street and 10th Street in San Antonio, Wash. Vyjit told Business Insidermedia.com that the police department is closed investigation. \\[0.5em]
"I believe it is damaging. We have limited evidence because we are closed for investigation," Vyjit said. "I think that there was power at all." \\[0.5em]
Daters and Waron officials are engaged together in this shooting. \\[0.5em]
"I don't want to realize that he is a person who is typically maximised and presreilled. I don't want to realize that he is a bad person who is left him in sheer condition," Atlanta Police have confirmed Monday. \\[0.5em]
Police are investigating some of the most dangerous people in Atlanta on occasion back because of contemplating and heightened sex violence for 3 weeks. \\[0.5em]
(Related More)\textless{}\textbar{}endoftext\textbar{}\textgreater{}You can eat your own gtadummyis in a very popular dish: Chocolate and idiosyncres are actually unique food. They have plenty of candy and some coconut packs, some spinach or with a slightly promulgating pipe bowl that rocked East Wlerheakin. You won't eat like the traditional gtadummyis (tberries for cherry, packs, and wood fuckies). They show a popular dish for under an hour. Useful bread can be sliced and topped with warm vanilla, such as illustrating your manure. \\[0.5em]
The premium gtadummyis is becoming also a popular dish because, after home as one of the most productive food restaurants on the list. There are Form that goes long ago to regret that; they need a lot of ample bread. \\[0.5em]
Jookien, which is called East Wlerheakin, is to take to the hospital gtadummyis is good, but won't include idiosyncres uncomfortable with wine powder and a series of juicy bread sauce. This is an odd texture and six coconut drinks and the bowl works really well using premium quality \\
\end{longtable}

\subsection{OpenWebText: CoBit-M}
\label{app:medium_samples}

\begin{quote}\small
\textbf{Leading excerpt of an uncurated generation.} This is the beginning of the first of four randomly selected CoBit-M generations, truncated at the marked point. The checkpoint has 462M parameters; sampling uses 256 NFEs, \(S_{\mathrm{churn}}=53.55\), target \(\gamma=0.21\), \(S_{\mathrm{noise}}=1.003\), entropy window \([0,1]\), and \(\eta=0\). The operating point has GenPPL 19.48 and entropy 5.40. Repetition and malformed wording are preserved.

complex problem."[22] Most like this as babies stayingverse is what regular hand geeks. Ultimately, at the game range the system is very crowded and "very dangerous."[22] Given both, he describes the R\$ factor caused by the roundup rejecting the graph and concludes that "the D+3 is not used to reason the dramagame the game."[22][23]

Reception [ edit ]

An essay in the journal by Richard Lip at the Institute of Neurological Psychiatry summarizes that the cognitive ability of women and males, zai and later hand geeks that are generally satisfied about the cognitive ability of men and contraceptives, have similar functionality.[24]\textless{}\textbar{}endoftext\textbar{}\textgreater{}Ready to fight back? Sign up for Take Action Now and get three actions in your inbox every week. You will receive occasional promotional offers for programs that support The Nation's journalism. You can read our Privacy Policy here. Sign up for Take Action Now and get three actions in your inbox every week.

Thank you for signing up. For more from The Nation, check out our latest issue

Subscribe now for as little as \$2 a month!

\textit{[Excerpt ends; generation continues.]}
\end{quote}

\section{Broader impacts}
\label{app:broader_impacts}

Compact bitwise input and output representations may make vocabulary boundaries
less memory intensive, while parallel generation may support more flexible text
completion. Our measurements concern specific hardware, model configurations,
and profiling workloads; they do not establish lower total training cost,
deployment energy use, or environmental impact.

Like other learned text generators, \methodname{} may reproduce biases in its
training data, generate inaccurate or harmful content, and be misused to produce
misinformation. Success on mathematical benchmarks does not establish
reliability, fairness, or safety in other settings, and the models studied here
have not been evaluated for deployment. Any deployment or model release would
require evaluation of these risks and documentation of appropriate safeguards;
the restricted execution used to score generated programs addresses only the
evaluation procedure.

\end{document}